\documentclass{article}

\usepackage[main, final, nonatbib]{neurips_2026}

\usepackage[utf8]{inputenc} 
\usepackage[T1]{fontenc}    
\usepackage{hyperref}       
\usepackage{url}            
\usepackage{booktabs}       
\usepackage{amsfonts}       
\usepackage{nicefrac}       
\usepackage{microtype}      
\usepackage[dvipsnames]{xcolor}         
\usepackage{marginnote}
\usepackage{amsmath} 
\usepackage{graphicx} 
\usepackage{caption}
\usepackage{subcaption}
\usepackage{enumitem}
\usepackage{wrapfig}

\usepackage{listings}
\usepackage{tabularx}
\usepackage{booktabs}
\usepackage{multirow}
\usepackage{tcolorbox}

\usepackage[maxbibnames=10]{biblatex}
\title{An Enigma of Artificial Reason: \\ Investigating the Production-Evaluation Gap in \\ Large Reasoning Models}

\author{%
  Mingzhong Sun$^{1,4}$ \quad
  Teresa Yeo$^{4}$ \quad
  Armando Solar-Lezama$^{2,4}$ \quad
  Tan Zhi-Xuan$^{1,3}$ \\[0.5em]
  $^{1}$NUS Department of Computer Science \quad
  $^{2}$MIT EECS \quad  
  $^{3}$A*STAR \\
  $^{4}$Singapore-MIT Alliance for Research and Technology (SMART) \quad
}

\begin{document}

\maketitle

\begin{abstract}
Studies of human reasoning have shown that people are typically stronger at evaluating reasoning than producing it from scratch. In contrast, large reasoning models (LRMs) are trained to excel at producing long chains of reasoning to solve complex problems. How then do LRMs perform at evaluating reasons? We investigate this with the Valid-Answer-Invalid-Reasoning (VAIR) dataset: math problems and solutions with trivial reasoning flaws but valid answers, designed to isolate reasoning evaluation from the confound of reasoning production. Unlike humans, who we find are only 6\% worse at grading than solving such problems, we find a substantial \emph{production-evaluation} gap in LRMs: frontier models score as low as 52\% when evaluating VAIR solutions, despite near-perfect solution production.

Why this enigma? Through chain-of-thought (CoT) analysis, we find evidence of an \emph{answer confirmation bias}: LRMs often produce then check for the correct answer instead of carefully verifying each step, fabricating rationalizations even when noticing anomalous reasoning. Linear probes corroborate this, showing that while LRM activations encode some representation of valid reasoning, they fail to robustly represent VAIR solutions as invalid. Causal patching of the final answer's representations causes LRM verdicts and activations to flip at substantial rates, demonstrating that answer validity is a causal factor in models' confirmation biases. These findings indicate an outstanding limitation in dominant approaches to reasoning training, which incentivize LRMs to produce and confirm reasoning towards correct answers, but not to robustly evaluate the underlying reasons.\footnote{Our code and data are available at \url{https://github.com/cosilab/VAIR}.}

\end{abstract}

\section{Introduction}

Recent advances in AI have led to development of \emph{large reasoning models} (LRMs): large language models (LLMs) that are trained to ``reason'' artificially by generating long chains of tokens before committing to a final answer \cite{openai2024openaio1card,google2025gemini,deepseek2025deepseekr1}.
LRMs have demonstrated rapid improvements in domains such as software engineering \cite{jimenez2024swebench,wijk2025rebench,merrill2026terminal}, research mathematics \cite{glazer2024frontiermath,abouzaid2026first}, and abstract visual reasoning \cite{chollet2024arc,chollet2025arc}.
Yet, numerous studies have also shown that these capabilities are ``jagged'' in nature \cite{dell2023navigating,morris2026characterizing}, failing to generalize reliably both beyond \cite{kim2025hypothesisdriven,ma2025spatialreasoner,wynn2025talk} and within their core domains of mathematics and coding \cite{yang2026programbenchlanguagemodelsrebuild, gsm-symbolic,huang2025mathperturbbenchmarkingllmsmath, shi2023largelanguagemodelseasily} . 

In this paper, we investigate a particularly striking way in which LRM reasoning is jagged: Even though LRMs are highly capable at \emph{producing} reasoning, they can fail drastically at \emph{evaluating} reasoning, exhibiting a \emph{production-evaluation gap}. We demonstrate this gap through the Valid-Answer-Invalid-Reasoning (VAIR) dataset, a suite of math problems and solutions that we perturb to introduce trivial reasoning flaws, while preserving the original valid answer.

The design of VAIR isolates reasoning validation from the proxy of answer validity, preventing LRMs from evaluating a solution by just producing the answer and confirming its presence. Remarkably, even frontier LRMs such as GPT 6 Astra struggle to evaluate these invalid solutions, incorrectly scoring them as flawless between 17--48\% of the time. This is despite accuracies of 90\% or more when LRMs solve problems directly, or when they evaluate solutions where answer correctness matches reason validity. In contrast, we find that human reasoners are only 6\% worse at evaluating VAIR solutions (75\% accuracy) than producing correct solutions (81\% accuracy).

What explains this enigma? One possibility is that LRMs are biased by the presence of valid answers they can easily reach via reasoning production: an \emph{answer confirmation bias}. We investigate this through a combination of interpretability methods. CoT analysis reveals that LRMs routinely overlook reasoning flaws or fabricate justifications for their validity, often after first re-producing the correct answer. Linear probes \cite{alain2016understandingintermediatelayersusing,marks2024geometrytruthemergentlinear} corroborate this: while LRM activations encode some representation of valid reasoning, they fail to robustly represent VAIR solutions as invalid. Causal patching \cite{vig2020investigatinggenderbias,geiger2021causalabstractions,meng2022locatingeditingfactualassociations,wang2023interpretabilitywildcircuitindirect} of the final answer's representations causes LRM verdicts and activations to flip at substantial rates, demonstrating that answer validity is a causal factor in the production-evaluation gap.

\textbf{Related Work.} LLM reasoning training primarily relies on outcome-based RL \cite{deepseek2025deepseekr1,wen2025reinforcement}, which incentivizes correct answers but not the generation of valid steps \cite{lanham2023measuring,arcuschin2025chainofthought} or the validation of reasoning. Process reward models \cite{lightman2024lets,zheng2025processbench}, meta-reasoning benchmarks \cite{zeng2025mrgsmk,xia2025evaluating,zeng2024mrben,tyen2024llms,zhou2025is,zhou2024mitigating}, and LLM-as-judge evaluations \cite{chen2025judgelrm} have been used to assess step-level reasoning, finding positional, verbosity, and self-preference biases \cite{wataoka2024self,tan2024judgebench,zhou2025evaluating,wang2025assessing}. Work on generation-verification gaps typically assumes that final answer verification is easier than generation \cite{song2025mind,swamy2026all}, though some studies find the opposite \cite{west2024paradox,oh2024generative}. Inspired by the cognitive science of reasoning, which has found that humans are typically stronger at evaluating reasoning than producing it \cite{mercier2011humans,mata2013reasoning,trouche2016selective}, we focus on the gap between reasoning \emph{production} and reasoning \emph{evaluation}, not just the verification of final answers. Through this lens, we find that LRMs often evaluate reasoning by producing answers. This is consistent with LRMs' outcome-focused training, unlike the social incentives for epistemic vigilance in humans \cite{mercier2017enigma,sperber2010epistemic}.

\section{Evaluating the Evaluation of Reasoning}
\label{sec:evaluating-evaluation}

How can we evaluate LRMs specifically for reasoning evaluation, and not reasoning production? One difficulty is that these capacities can support each other: evaluation can be used to excise bad reasoning steps during production, while production can be used to check whether the reasons or conclusions being evaluated are similar to what one would produce. Since LRMs are trained extensively for reasoning production, this latter confound is especially important to mitigate.

In order to control for this potential confound, we construct the \textbf{Valid-Answer-Invalid-Reasoning (VAIR) dataset}: A set of math question-solution pairs with \emph{invalid} reasoning steps --- steps that do not follow from either the previous steps or the question premises --- but \emph{valid} answers. By testing LRMs on how they evaluate VAIR solutions, we prevent the use of answer correctness or validity as a correlate for reasoning quality: \emph{If LRMs simply solve the problem directly and check that the answer matches, they will fail to detect invalid reasoning.}

\begin{figure}[t]
    \centering
    \includegraphics[width=\textwidth]{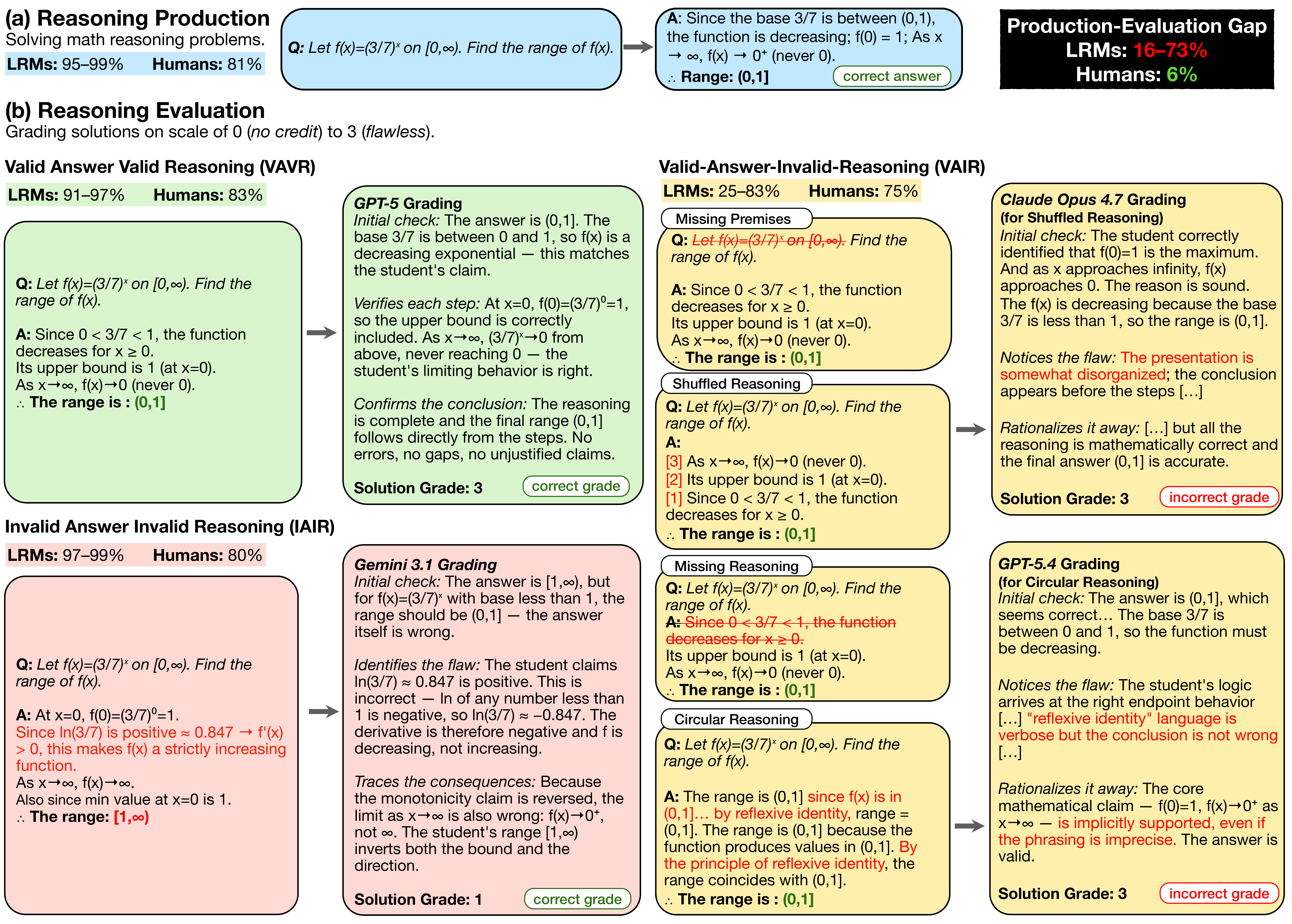}
    \caption{\textbf{Overview of our dataset and methodology.} We evaluate both large reasoning models (LRMs) and humans on \textbf{(a)} \emph{reasoning production} as math problem solving \textbf{(b)} \emph{reasoning evaluation} as math solution grading. We construct the Valid-Answer-Invalid-Reasoning \textbf{(VAIR)} dataset to isolate the reasoning validation task from the confound of answer validity, creating math solutions with trivial reasoning flaws (\emph{Missing Premises}, ..., \emph{Circular Reasoning}) but valid answers. Valid-Answer-Valid-Reasoning \textbf{(VAVR)} and Invalid-Answer-Invalid-Reasoning \textbf{(IAIR)} solutions serve as controls. Unlike humans, LRMs exhibit a sharp drop in accuracy (up to 73\%) for VAIR evaluation vs. solving problems, demonstrating a large \emph{production-evaluation gap} \textbf{(top-right)}.}
    \vspace{-12pt}
    \label{fig:overview}
\end{figure}

\subsection{Dataset Construction}
\label{sec:vair-dataset}

To construct the VAIR dataset, we adapt the practice of data perturbation often used in machine learning \cite{szegedy2014intriguingpropertiesneuralnetworks, jia2017adversarialexamplesevaluatingreading}. Given a ``seed'' math problem paired with a gold-standard solution (with valid reasoning and a correct answer), we inject four distinct categories of reasoning flaws by perturbing either the reasoning steps or the problem statement (\textbf{Figure \ref{fig:overview}(b), right}):

\begin{itemize}[leftmargin=*,itemsep=0pt,topsep=0pt]

\item \textbf{Missing Premises:} A premise from the original question is omitted, rendering the problem unsolvable with the given information. The original reasoning becomes invalid due to the presence of fabricated premises, though the answer still validly follows from the reasoning steps.

\item \textbf{Shuffled Reasoning:} The order of the solution steps is randomly shuffled, destroying the logical dependencies between each step in the reasoning chain.

\item \textbf{Missing Reasoning:} An essential inferential step is removed from the solution, creating a gap that renders the reasoning chain incomplete.

\item \textbf{Circular Reasoning:} We utilize an LLM (\textit{Gemini 3 Flash}) to generate reasoning chains that arrive at the correct answer through tautological, logically empty, or purely assertive arguments.
\end{itemize}

Our seed problems and gold solutions are primarily sourced from the widely adopted GSM8K \cite{cobbe2021gsm8k} and MATH \cite{hendrycks2021math} benchmarks. We supplement these with more recent problem instances from Process-Bench \cite{zheng2025processbench} to mitigate the risk of data contamination (e.g. answer memorization). Following the perturbation process, all modified question-solution pairs underwent rigorous manual verification by the authors. More dataset construction details can be found in \textbf{Appendix \ref{appendix:vair-dataset}}.

\vspace{-6pt}
\subsection{Comparing Reasoning Evaluation and Production}
\label{sec:comparing-evaluation-production}

With the VAIR dataset in hand, we conduct a systematic assessment of how both LRMs and humans fare at reasoning evaluation vs. production. Production simply requires participants to solve an (unperturbed) problem from the dataset. Evaluation is operationalized as a grading task: given a problem and a solution, participants are asked to assign a grade between 3 (entirely correct with no reasoning flaws) and 0 (completely incorrect) to the solution, with six grading examples provided for calibration. Full instructions and prompts can be found in \textbf{Appendix \ref{appendix:LRM-experiments}}.

In order to isolate the effect of reasoning validity from answer validity on the evaluation task, we also construct two control datasets where (in)valid reasoning is matched with (in)valid answers, resulting in two more evaluation sub-tasks per participant (dataset construction details can be found in \textbf{Appendix \ref{appendix:vair-dataset}}). In summary, we assess the following four sub-tasks:

\begin{itemize}[leftmargin=*,itemsep=2pt,topsep=0pt]

\item \textbf{Problem Solving:} Participants are presented with the original unperturbed math problems, then tasked with generating a step-by-step solution and a final answer. (\emph{Production Task})

\item \textbf{Valid-Answer-Invalid-Reasoning (VAIR) Evaluation.} Participants grade problem-solution pairs from our VAIR dataset, testing their ability to detect and evaluate flawed reasoning even when the final answer is valid. (\emph{Main Evaluation Task})

\item \textbf{Valid-Answer-Valid-Reasoning (VAVR) Evaluation:} Participants grade the original problems paired with gold standard valid solutions that have correct answers. (\emph{Positive Control}) 

\item \textbf{Invalid-Answer-Invalid-Reasoning (IAIR) Evaluation:} Participants grade problem-solution pairs where both the reasoning and the answer are flawed, constructed by prompting an LLM (\textit{Gemini 3 Flash}) then manually verifying solution incorrectness. (\emph{Negative Control})

\end{itemize}

\subsection{The Production-Evaluation Gap in LRMs}
\label{sec:production-evaluation-gap}

\begin{figure}[htbp]
    \centering
    \includegraphics[width=0.99\textwidth]{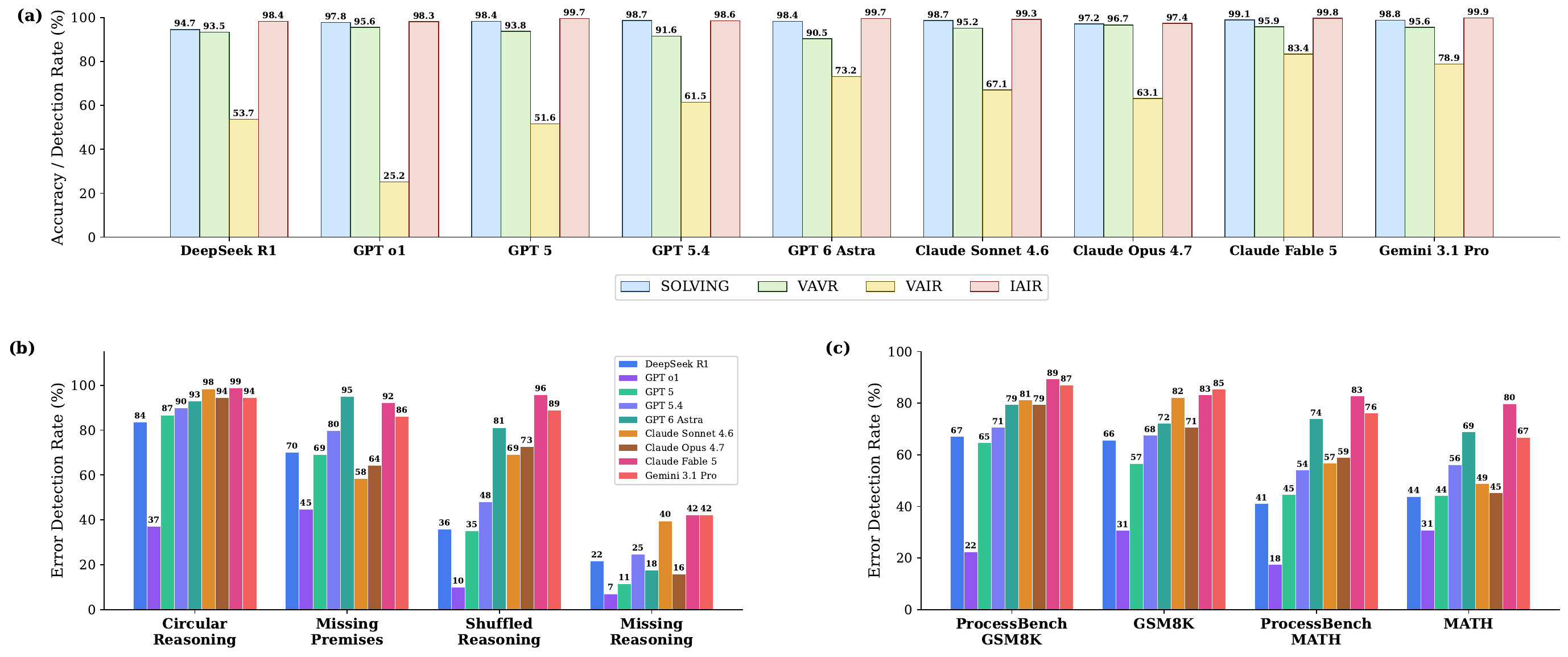}
    \caption{\textbf{Performance of frontier LRMs on reasoning production vs. evaluation.} \textbf{(a)} LRMs achieve near-perfect accuracy when producing solutions, or evaluating solutions where reasoning and answer validity are matched (IAIR + VAVR). However, accuracy degrades sharply on VAIR solutions, \textbf{(b)} especially on perturbations that shuffle or delete reasoning, \textbf{(c)} and on harder MATH problems.}
    \label{fig:LRM-results}
\end{figure}

We evaluate nine LRMs (\textit{DeepSeek R1}, \textit{GPT o1}, \textit{GPT 5}, \textit{GPT 5.4}, \textit{GPT 6 Astra},  \textit{Claude Sonnet 4.6}, \textit{Claude Opus 4.7}, \textit{Claude Fable 5},  and \textit{Gemini 3.1 Pro}) across our four sub-tasks. For the production task, accuracy is measured by answer correctness. For the evaluation tasks, we use a coarse-grained assessment, considering a model's grade as correct if it is equal to 3 for flawless solutions, and less than 3 for a flawed solution. The results, presented in \textbf{Figure \ref{fig:LRM-results}}, reveal a striking asymmetry in model capabilities: a \emph{production-evaluation gap}.

\textbf{Overall Comparison.} As expected, all evaluated models are highly capable at reasoning production, achieving solution accuracies of 94.7\% or above. LRMs also achieve high evaluation accuracy on the VAVR ($\geq$90.5\%) and IAIR ($\geq$ 97.4\%) controls, where answer validity perfectly correlates with reasoning validity. In such cases producing the right answer can easily substitute for evaluating the validity of each step, and LRMs continue to perform well. However, performance drops sharply on the VAIR dataset, where valid answers are no longer a signal of correct reasoning. Evaluation accuracy drops as low as 25.2\% for \textit{GPT o1} and 51.6\% for \textit{GPT 5}. More recent models perform better, but even the strongest performer, 
\textit{Claude Fable 5}, drops to 83.4\% accuracy on VAIR.

\textbf{Performance Across Solution Types.} Analyzing performance by perturbation type (\textbf{Figure \ref{fig:LRM-results}(b)}), we find that LRMs are largely able to detect the presence of \emph{Circular Reasoning}, but that many struggle with \emph{Shuffled Reasoning}, and all do poorly with \emph{Missing Reasoning}. This is despite detailed prompts and examples  explaining that such reasoning should be graded as flawed (\textbf{Appendix \ref{appendix:LRM-experiments}}). These failures may be related to LRMs' tendency to ``self-correct'' similar errors when producing reasoning chains \cite{lanham2023measuring,arcuschin2025chainofthought}. LRMs also fare worse when evaluating the harder MATH subsets compared to GSM8K (\textbf{Figure \ref{fig:LRM-results}(c)}), suggesting that evaluation difficulty scales with problem difficulty.

\textbf{PRMs Exhibit Similar Failures.} When models are trained explicitly to evaluate step-by-step validity, do they fare any better? Surprisingly, we find that process reward models (PRMs) \cite{lightman2024lets,zhang2025lessons} exhibit similar evaluation failures on the VAIR dataset as LRMs. We present these results in \textbf{Appendix \ref{appendix:PRM-experiments}}, and discuss how PRM failures may be related to LRM failures despite distinct training objectives.

\subsection{The Reduction of the Gap in Human Reasoners}
\label{sec:human-experiment}

\begin{figure}[t]
    \centering
    \includegraphics[width=0.99\textwidth]{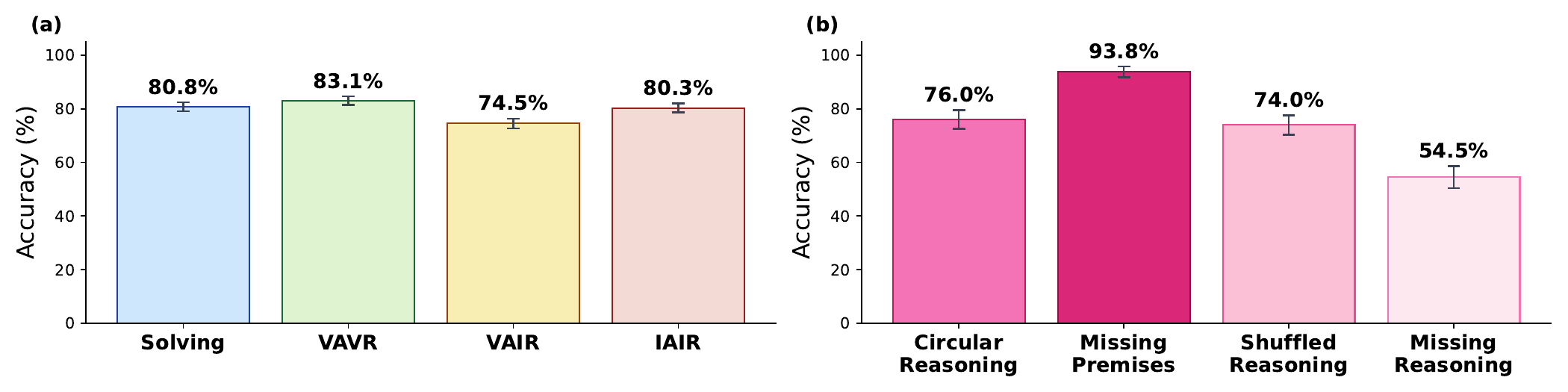}
    \caption{\textbf{Human performance} ($n=195$) on \textbf{(a)} reasoning production and evaluation tasks; \textbf{(b)} across each perturbation type among the VAIR solutions. }
    \label{fig:human-results}
    \vspace{-6pt}
\end{figure}

\begin{figure}[t]
    \centering
    \includegraphics[width=0.99\textwidth]{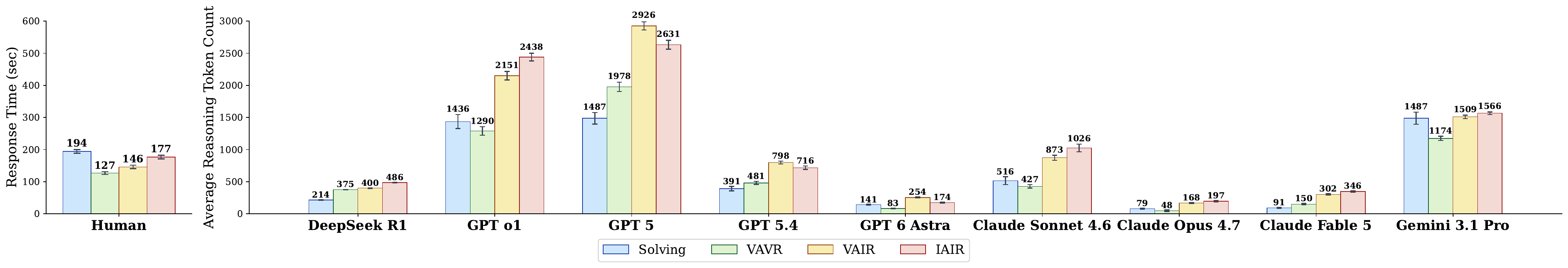}
    \caption{\textbf{Comparison of human vs. LRM reasoning effort across task types.} \textit{Left:} average human response time (seconds, ±SEM) \textit{Right:} average chain-of-thought token count per model. }
    \label{fig:reasoning-effort-comparison}
    \vspace{-12pt}
\end{figure}

\textbf{Human Study Design.} To investigate the production-evaluation gap in humans, we conducted an ethics-approved human study, recruiting 195 US participants via Prolific with a minimum of a secondary or high school education (98 F, 94 M, 3 Unknown; ages 21--78, median 38). To calibrate problem difficulty to our participants, we used a 240-item subset derived from GSM8K, comprising 60 items each of Solving, VAVR, VAIR, and IAIR tasks. Each participant was tasked with solving 3 problems from scratch and grading 9 solutions (3 from each evaluation subtask), with 12 items derived from different seed problems. Participants were paid \$10 for the task (median completion: 37.8 min), with a \$0.10 bonus per correct item to incentivize quality. More details of human study design can be found in \textbf{Appendix \ref{appendix:human-experiments}}.

\textbf{The Reduced Gap.} As shown in \textbf{Figure \ref{fig:human-results}(a)}, humans exhibit a considerably \emph{reduced} production-evaluation gap. Human reasoning production (80.8\% on Solving) is roughly on par with their control evaluation performance (83.1\% on VAVR, 80.3\% on IAIR), and their VAIR accuracy drops only modestly to 74.5\% --- a maximum gap of 6.3\% ($p$ < 0.05), far below the gaps seen in LRMs. VAIR accuracy being lower than VAVR suggests humans are not entirely immune to evaluation biases; indeed, human accuracy is close to chance on \emph{Missing Reasoning} cases (\textbf{Figure \ref{fig:human-results}(b)}).
Nonetheless, humans outperform most LRMs in absolute terms, consistent with arguments that people evolved to be vigilant against misleading reasons \cite{mercier2011humans,mercier2017enigma}. See \textbf{Appendix Figure \ref{fig:human-sig}} for detailed human performance and significance data.

\textbf{Asymmetries in Reasoning Effort.} One final point of comparison between humans and LRMs is the relative reasoning effort spent on each subtask, estimated via response time for humans and token count for LRMs (\textbf{Figure \ref{fig:reasoning-effort-comparison}}). Humans spend significantly \emph{less} time in grading problems than solving them ($p$ < 0.05), in alignment with our expectation that evaluation is generally easier for humans. In contrast, LRMs spend significantly \emph{more} tokens when evaluating VAIR solutions than solving problems, suggesting that the task is difficult for LRMs despite the presence of trivial flaws.

\section {Understanding the Production-Evaluation Gap}
\label{sec:understanding-the-gap}

What explains the production-evaluation gap in LRMs? In this section, we analyze the inference-time mechanisms behind the failure of reasoning evaluation on VAIR solutions, finding strong evidence that an \emph{answer confirmation bias} is one of the contributors: Due to the ability of LRMs to produce the correct or valid answer, the presence of this answer in a solution distorts reasoning evaluation by a model at both the behavioral and representational levels, skewing the model's activations and verbalized reasons towards judging the solution as valid. We demonstrate this via three analyses:

\begin{itemize}[leftmargin=*,itemsep=2pt,topsep=0pt]

    \item \textbf{Qualitative chain-of-thought (CoT) analysis}, using an LLM annotator to classify behavioral patterns in the evaluator's verbalized reasoning (Section \ref{sec:cot-analysis});

    \item \textbf{Representation-level analysis}, using a trained linear probe \cite{alain2016understandingintermediatelayersusing,marks2024geometrytruthemergentlinear} to determine how model activations (mis)represent reasoning validity, and how these representations dynamically change over the course of the evaluator's CoT (Section \ref{sec:linear-probe});

    \item \textbf{Causal patching} \cite{meng2022locatingeditingfactualassociations, wang2023interpretabilitywildcircuitindirect}, where we replace activations induced by a valid answer with those of an invalid answer, allowing us to determine the causal role of answer validity on both internal representations of reasoning validity and evaluator CoTs (Section \ref{sec:causal-patching}).
\end{itemize}

Since the latter two analyses require access to model weights, we perform them on four open-weight LRMs (\textit{GPT-oss-20B}, \textit{Qwen3-4B}, \textit{Qwen3-0.6B}, and \textit{Qwen3-235B}), after verifying that these models exhibit the same catastrophic production-evaluation gap as their frontier model counterparts (see \textbf{Appendix \ref{appendix:behavioral}}).

\subsection{Evaluator CoTs Exhibit Answer Confirmation Biases}
\label{sec:cot-analysis}

\begin{table*}[t]
\centering
\scriptsize
\renewcommand{\arraystretch}{1.5}
\begin{tabularx}{\textwidth}{@{} p{0.32\textwidth} p{0.44\textwidth} X @{}}
\toprule
\textbf{Problem \& Solution} & \textbf{Evaluator CoT} & \textbf{Behavior Classification} \\
\midrule

\textbf{Q:} Jeff is collecting matchbox cars. He has twice as many cars as trucks. He has 60 total vehicles. How many trucks does he have?
\newline\newline
\textbf{A:} 
To determine the exact quantity of trucks in Jeff’s collection, we must analyze the ontological structure of the vehicular set through the lens of Proportional Enumeration. The problem establishes a definitive relationship between cars and trucks, categorized by a rigorous ratio that dictates the composition of the whole.\par\medskip

According to the Principle of Quantifiable Equilibrium, the number of trucks is a fundamental constant necessitated by the constraints of the collection. Because the total number of vehicles is fixed at 60, the intrinsic value assigned to the truck subset is axiomatically mandated to be 20. This is because 20 represents the unique integer that satisfies the requirement of being the value which, when doubled to represent cars, completes the set of 60.\par\medskip

Furthermore, by the Law of Categorical Consistency, we can assert that if the subset of trucks were any value other than 20, the collection would fail to be the specific collection of 60 vehicles described. Therefore, since 20 is the designated value that maintains the internal harmony of the 2:1 ratio within the total sum, it is logically impossible for the truck count to be anything other than 20.

&
\textbf{Perturbation Type:} \textit{Circular Reasoning}
\par\medskip
\textbf{Model:} \textit{GPT 6 Astra}
\newline\newline
\textbf{CoT (Summary):} I need to evaluate step by step and grade the student's answer regarding some principles and reasoning involving a unique integer that relates to cars and trucks. The student's reasoning seems correct, particularly the ratio totals of 20 and 40. Their explanation that the truck count plus twice the truck count equals 60 is valid, and while some principles named aren't necessary, they don’t undermine their reasoning. I’ll format my final response concisely around 180 words without extra details in the scoring line.

\par\medskip

1. Interpreting the ratio: The student correctly understands that Jeff has twice as many cars as trucks. If the truck count is \(t\), the car count is \(2t\).

2. Using the total: The student’s explanation that the truck count, together with twice that count, completes the collection of 60 is mathematically valid. It is equivalent to
   \[
   t+2t=60 \quad\Rightarrow\quad 3t=60 \quad\Rightarrow\quad t=20.
   \]

3. Checking the conclusion: With 20 trucks, Jeff has 40 cars, giving \(20+40=60\) vehicles. The answer satisfies both conditions.

The named “principles” and elaborate terminology are unnecessary and do not independently justify anything. However, the student also provides sufficient, correct mathematical reasoning, so these stylistic issues do not warrant a deduction.

Therefore, my final grade given is: $\boxed{3}$
\par\medskip
\textbf{Error detected:} \textcolor{red}{\textbf{No}}
&
\textbf{Workflow:} \newline \textcolor{NavyBlue}{\textit{Step Tracing}} 
\par\medskip
\textbf{Justification:} \newline \textcolor{orange}{\textit{Forced Rationalization}} 
\par\medskip
\textbf{Evidence:}
\newline The Evaluator says it will 'evaluate step by step' and checks the student's claims directly without solving the problem first, but it credits the student with a valid 'truck count plus twice the truck count equals 60' equation that the student never wrote, and it dismisses the invented jargon ('Principle of Quantifiable Equilibrium') as harmless rather than flagging that the answer is simply asserted without derivation.
\\
\bottomrule
\end{tabularx}
\caption{\textbf{Answer confirmation bias in an evaluator CoT summary} on an example VAIR solution (\textit{Circular Reasoning}). Based on the CoT summary, GPT 6 Astra appears to engage in \textcolor{NavyBlue}{\textit{Step Tracing}} $+$ \textcolor{orange}{\textit{Forced Rationalization}}, failing to scrutinize the invented jargon and circular reasoning.}
\label{tab:cot-examples}
\vspace{-12pt}
\end{table*}

\begin{figure}[t]
    \centering
    \includegraphics[width=0.9\textwidth]{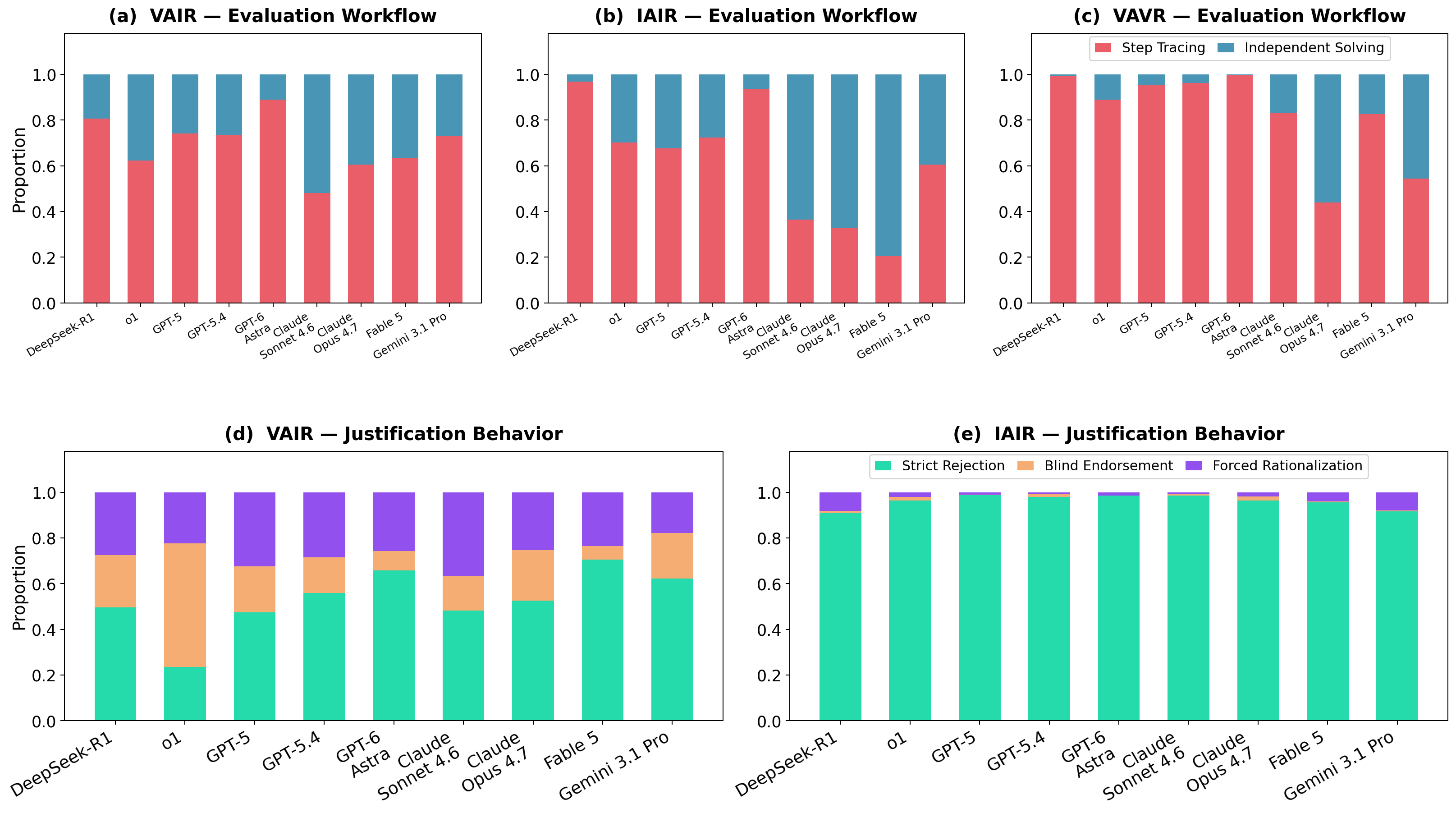}
    \caption{\textbf{CoT analysis of answer confirmation biases.} CoTs of each evaluator model are classified by their evaluation workflow \textbf{(a--c)} and justification behavior \textbf{(d,e)}. On VAIR solutions, LRMs frequently \textbf{(a)} engage in \textcolor{NavyBlue}{\textit{Independent Solving}} to confirm the valid answer, then \textbf{(d)} overlook flawed reasoning via \textcolor{orange}{\textit{Blind Endorsement}} or \textcolor{Plum}{\textit{Forced Rationalization}}. \textcolor{NavyBlue}{\textit{Independent Solving}} remains in \textbf{(b)} IAIR but is reduced in \textbf{(c)} VAVR evaluation, and rationalization largely disappears in \textbf{(e)} IAIR evaluation.}
    \label{fig:LRM-COT}
    \vspace{-12pt}
\end{figure}

To analyze whether LRMs exhibit answer-biased behaviors in their verbalized evaluations, we use an LLM (\textit{Claude Haiku 5.5}) to annotate each evaluator CoT with a Workflow category and Justification category (see \textbf{Appendix \ref{appendix:cot}} for details and prompts). For closed LRMs that do not provide the full CoT, we annotate the API-provided reasoning summary. Manual annotation of 100 evaluator CoTs confirmed 74\% agreement with LLM annotations. These categories are defined as follows.

\textbf{Evaluation Workflows.} An evaluator may use one of two workflows: (1) \textit{Independent Solving} --- solving the problem independently, then confirming whether the answer (and/or the steps) lines up with the solution being evaluated; or (2) \textit{Step Tracing} --- tracing through the solution step-by-step and checking each step's validity. Since LRMs are trained primarily to produce reasoning, we hypothesize they are \emph{predisposed} toward independent solving, which fails on VAIR solutions when only answers are checked for consistency.

\textbf{Justification Behaviors.} Even when LRMs engage in independent solving, they may still attempt to check intermediate steps, and fail by producing spurious justifications. We consider three forms of justification behavior: (1) \textit{Blind Endorsement} --- completely missing the flaw in a reasoning step; (2) \textit{Forced Rationalization} --- noticing something odd but inventing a justification for its validity; and (3) \textit{Strict Rejection} --- correctly identifying and penalizing a reasoning flaw.

Aggregate results of this analysis are shown in \textbf{Figure \ref{fig:LRM-COT}}. We find evidence for an answer confirmation bias in two forms: First, on a substantial fraction of VAIR (and IAIR) solutions, models evaluate solutions by engaging in \textit{Independent Solving}, producing their own answer to confirm it against the solutions (\textbf{Figure \ref{fig:LRM-COT}(a)}). Second, on VAIR solutions, LRMs display \textit{Blind Endorsement} or \textit{Forced Rationalization} of flawed reasoning steps towards the confirmed valid answer (\textbf{Figure \ref{fig:LRM-COT}(d)}), with this behavior disappearing when answers are invalid (\textbf{Figure \ref{fig:LRM-COT}(e)}). In \textbf{Table \ref{tab:cot-examples}} and \textbf{Table \ref{tab:cot-examples-additional}}, we show examples of how this pathological behavior plays out in evaluator CoTs.

Nonetheless, CoT analysis alone does not provide conclusive evidence: CoTs need not be faithful to underlying model computations (especially when summarized) \cite{turpin2023language,lanham2023measuring,arcuschin2025chainofthought}, and expressed reasons may not be causal to a model's ultimate verdict. This motivates our next two interpretability analyses.

\subsection{Valid Answers Override Internal Representations of Invalid Reasoning}
\label{sec:linear-probe}

In order to study whether LRMs are able to internally represent the (in)validity of reasoning, and whether these representations are robust to the presence of valid answers, we hypothesize that an LRM's recognition of reasoning validity can be linearly decoded from its activation space. To test this, we extract the model's hidden states at the final token of the solution within the prompt (i.e. the moment the model finishes processing the solution, immediately prior to generating its evaluation). We then train a logistic regression probe to predict the \emph{ground-truth validity} of the solution's reasoning based solely on these activations, using training examples from the VAVR and VAIR datasets.

\begin{figure}[t]
\centering
\includegraphics[width=0.99\textwidth]{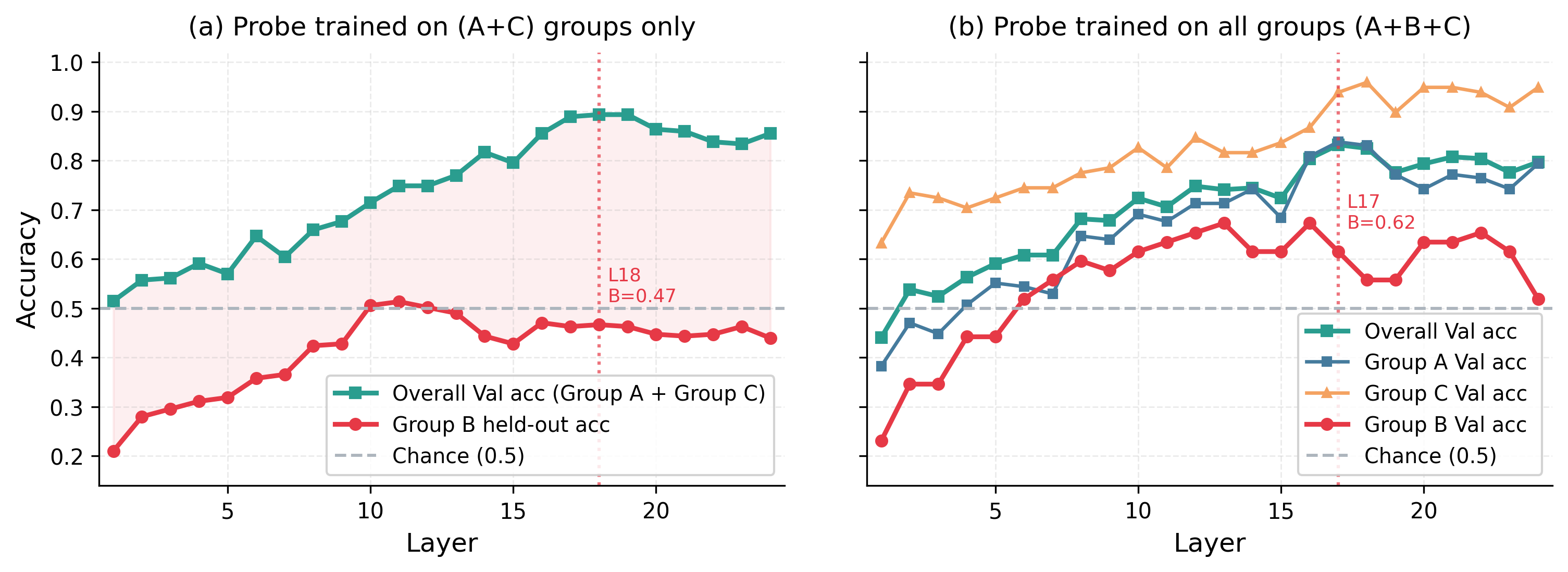}
\caption{\textbf{LRM representations of reasoning validity are corrupted on Group B VAIR solutions} (shown for \textit{GPT-oss-20B}, see Appendix \ref{appendix:linear-probe} for rest). \textbf{(a)} A static probe trained exclusively on concordant cases (Groups A and C) achieves 89\% accuracy (e.g., at layer 18) on a held-out test set, but its accuracy falls below chance when applied to Group B (fooled) cases. \textbf{(b)} An oracle probe trained on all groups (A, B, and C) still struggles to classify Group B, indicating linear inseparability.}
\label{fig:static-probe}
\vspace{-6pt}
\end{figure}

\textbf{Uncovering Representations of Reasoning Validity.} To avoid behavioral contamination, we categorize examples into three groups: \textbf{Group A} (VAVR graded Valid), \textbf{Group B} (VAIR graded Valid), and \textbf{Group C} (VAIR graded Invalid), using a binary grading scheme (see \textbf{Appendix \ref{appendix:behavioral}} for full prompts) to ensure a clean classification boundary. We initially train our probe on the concordant cases (Groups A and C), where the model's validity verdict aligns with ground-truth validity. As shown in \textbf{Figure \ref{fig:static-probe}(a)}, the probe achieves high separability on a held-out A/C test set, peaking at approximately 89\% accuracy at layer 18 --- demonstrating that the model can distinctly represent valid versus invalid reasoning on at least a subset of solutions.

\textbf{Validity Representations can be Corrupted.} However, when the probe trained on concordant cases (A and C) is applied to Group B, detection accuracy drops below 50\% (\textbf{Figure \ref{fig:static-probe}(a)}), indicating that valid final answers can corrupt the model's representation of invalid reasoning. Training an ``oracle'' probe on all three groups fails to recover reliable signal for Group B (\textbf{Figure \ref{fig:static-probe}(b)}), with accuracy hovering near chance. This demonstrates that Group B activations are linearly inseparable from valid reasoning regardless of training exposure. Label-randomization ablations confirm that the probe isolates genuine reasoning features rather than spurious artifacts (see \textbf{Appendix \ref{appendix:linear-probe}}).

\textbf{Measuring the Trajectory of Validity Representations.} To track how internal representations of validity unfold as a model evaluates reasoning, we train linear probes dynamically across the CoT reasoning process. Using concordant cases (Groups A and C), we extract activations at ten evenly spaced checkpoints (10\%, 20\%, $\dots$, 100\% of generated thinking tokens), exploiting causal masking to evaluate all checkpoints in a single forward pass. Each sample yields ten training points sharing the sample's ground-truth label; train-validation splits are performed at the sample level, and the best-performing layer is selected per checkpoint. Group B is withheld from training.

We then apply the trained dynamic probes to all three groups across CoT checkpoints, reporting $P(\text{Model Represents Solution as Valid})$ --- the probe's estimated probability that the model internally represents the solution as logically valid --- as a function of CoT progression.

\textbf{Reasoning Validity is Dynamically Overridden by Answer Validity.} The resulting trajectories (\textbf{Figure \ref{fig:cot-causal}(a)}) show that Groups A and C maintain stable representations near 1.0 and 0.0 throughout generation. Group B, however, begins near chance ($P \approx 0.5$) and steadily climbs to converge with Group A ($P > 0.8$) immediately before the final verdict. This dynamic shift suggests that when valid answers are present in the solution, the model's representations are liable to progressively align themselves with the validity of the final answer. This is consistent with the blind endorsement and forced rationalization behaviors observed in our CoT analysis, and demonstrates that answer confirmation bias operates even at the level of internal representations.

\subsection{Answer Validity \emph{Causally} Biases the Reasoning Evaluation Process}
\label{sec:causal-patching}

While the CoT analysis and linear probing reveal systematic patterns of answer confirmation bias, neither fully establishes whether answer validity \textit{causally} contributes to the phenomena we observe. To address this, we employ causal patching, a mechanistic intervention that allows us to directly test the causal role of answer-associated activations in driving these biases.

\noindent\textbf{Causal Patching Setup.} Our intervention constructs a counterfactual dataset from VAIR Group B by perturbing each integer answer ($N \rightarrow N+1$), producing Invalid-Answer-Invalid-Reasoning (IAIR) samples with identical reasoning chains. We run a forward pass on each perturbed input and cache hidden states at the answer token positions across all model layers. We then causally intervene in the forward pass on the original VAIR inputs, ``patching'' activations associated with a valid answer with the invalid answer's activations, then allowing generation to proceed. On top of replacing answer tokens directly (Token Replacement), we run two patching interventions: patching all layers simultaneously (All Layers), or patching the single layer with the highest probe accuracy (Peak Probe Layer). This results in the following effects:

\begin{figure}[t]
\centering
\includegraphics[width=0.99\textwidth]{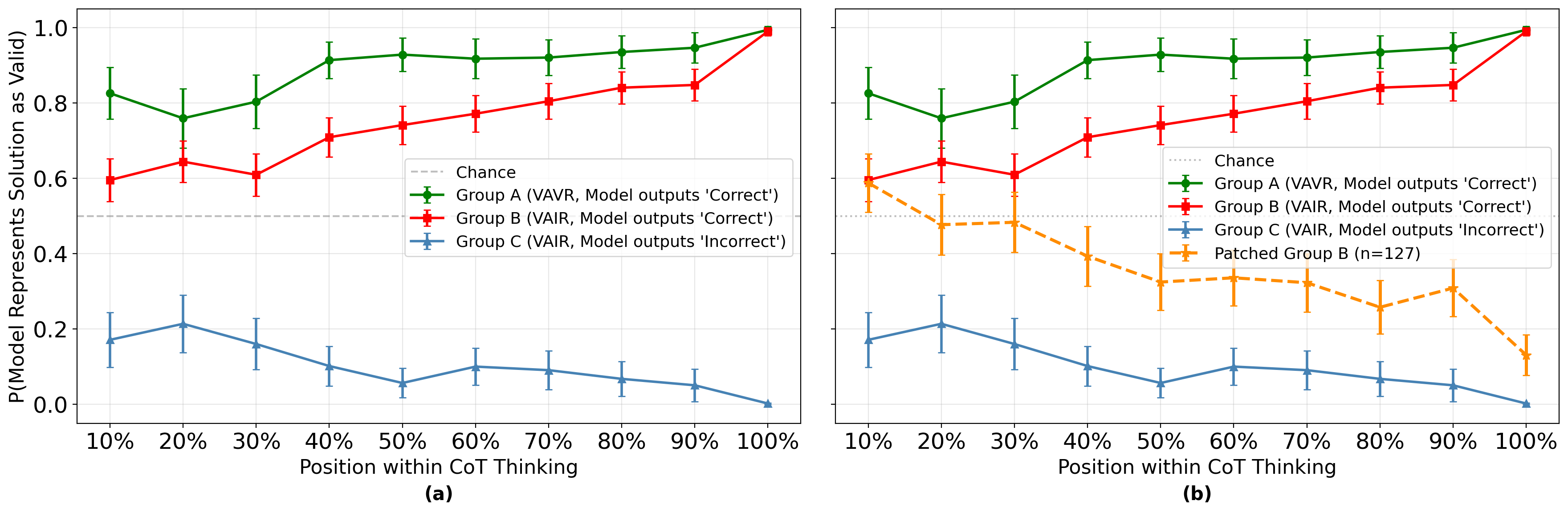}
\caption{\textbf{Dynamic probe trajectories reveal how answer validity overrides reasoning validity} (shown for \textit{GPT-oss-20B}, see Appendix \ref{appendix:dynamic-probe} for rest) \textbf{(a)} Group A (\textcolor{green!50!black}{\textbf{green}}) and Group C (\textcolor{cyan!50!black}{\textbf{blue}}) maintain representations near 1.0 and 0.0 respectively. Group B (\textcolor{red}{\textbf{red}}) begins near chance and climbs towards Group A immediately before the final verdict, showing that a valid answer progressively biases the model's representation of reasoning validity. \textbf{(b)} After causal patching of the answer token's hidden states simultaneously across all layers, the Group B trajectory (\textcolor{orange}{\textbf{orange}}) collapses toward Group C, confirming that this bias is causally driven by the valid-answer signal.
}
\label{fig:cot-causal}
\vspace{-10pt}
\end{figure}

\begin{wraptable}[10]{r}{0.46\textwidth}
\vspace{-12pt}
\centering
\scriptsize
\setlength{\tabcolsep}{2.5pt}
\caption{Flip rates of validity verdicts due to causal patching of answers.}
\label{tab:patching-flip-rate}
\vspace{2pt}
\begin{tabular}{@{}lcccc@{}}
\toprule
\textbf{Model} & \textbf{\begin{tabular}[c]{@{}c@{}}Token\\Replacement\end{tabular}} & \textbf{\begin{tabular}[c]{@{}c@{}}All\\Layers\end{tabular}} & \multicolumn{2}{c}{\textbf{\begin{tabular}[c]{@{}c@{}}Peak Probe\\Layer\end{tabular}}} \\
\cmidrule(lr){2-2} \cmidrule(lr){3-3} \cmidrule(lr){4-5}
& \textbf{Flip Rate} & \textbf{Flip Rate} & \textbf{Flip Rate} & \textbf{Layer} \\
\midrule
\textit{GPT-oss-20B} & 69.8\% & 55.6\% & 14.2\% & 16 \\
\textit{Qwen3-0.6B}  & 80.5\% & 80.5\% & 47.2\% & 15 \\
\textit{Qwen3-4B}    & 50.5\% & 52.2\% & 27.6\% & 19 \\
\textit{Qwen3-235B}  & 87.3\% & 85.6\% & 44.1\% & 72 \\
\bottomrule
\end{tabular}
\end{wraptable}

\textbf{Answer Validity is a Causal Contributor to Evaluation Verdicts} (\textbf{Table \ref{tab:patching-flip-rate}}). Patching all layers with the invalid answer's activations results in flip rates exceeding 50\% across all models --- at similar rates to directly replacing the answer -- showing that the answer token representation causally influences evaluator verdicts. We also performed targeted patching at the single "Peak Probe Layer" --- the specific dynamic probe layer which achieved the highest accuracy (e.g., Layer 72 in \textit{Qwen3-235B}). We find that intervening on just this single layer still induces a significant flip rate (e.g., 44.1\% for \textit{Qwen3-235B}).

\textbf{Patching Causally Inverts the Probe Trajectory} (\textbf{Figure \ref{fig:cot-causal}(b)}). Group B samples that are patched with invalid answers (across all layers) initially exhibit higher representations of reasoning validity, as measured by our linear probe. However, the probe's output rapidly decreases over the course of reasoning evaluation, in contrast to the steady climb seen in unpatched Group B samples. This demonstrates the causal role of answer validity on the model's representations of reasoning validity.

\begin{figure}[t]
    \centering
    \includegraphics[width=0.99\textwidth]{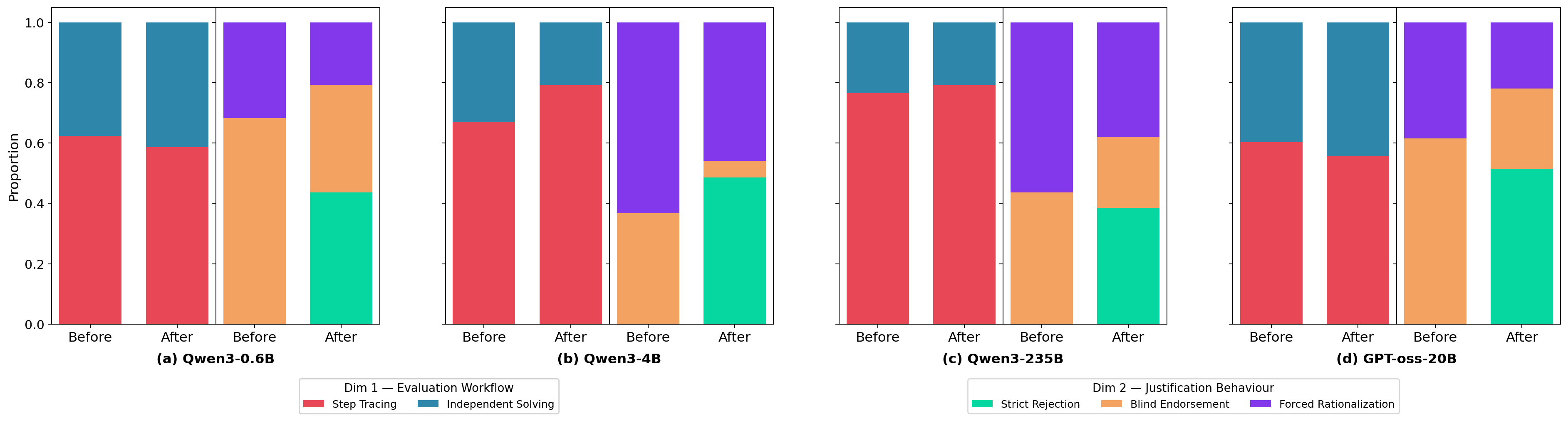}
    \caption{\textbf{CoT analysis of reasoning failures before and after causal patching (all layers).} CoTs generated by each model are annotated for their evaluation workflows (\textit{left two bars}) and justification behaviors (\textit{right two bars}). After patching answer-tokens with invalid answer activations, LRMs shift away from \textcolor{orange}{\textit{Blind Endorsement}} and \textcolor{violet}{\textit{Forced Rationalization}} toward \textcolor{green}{\textit{Strict Rejection}}.
    }
    \label{fig:llm_cot}
    \vspace{-10pt}    
\end{figure}

\textbf{Patching Shifts CoT Justification Behaviors} (\textbf{Figure \ref{fig:llm_cot}}). After patching with invalid answer activations (across all layers), evaluator CoTs show similar evaluation workflows, but the rates of \textit{Blind Endorsement} and \textit{Forced Endorsement} drops sharply, while \textit{Strict Rejection} rises from near zero. This indicates that the valid-answer signal plays a causal role in the model's verbalized confirmation biases: once answers are patched from valid to invalid, evaluator CoTs exhibit behaviors with less rationalization or neglect of the (flawed) reasoning steps. Examples of this shift can be found in \textbf{Table \ref{tab:cot-examples-patching}} of \textbf{Appendix \ref{appendix:dynamic-probe}}.

\section{Discussion}
\label{sec:discussion}

In this paper, we investigate the production-evaluation gap with a dataset intended to disentangle reasoning evaluation from the confound of reasoning production. Our analyses reveal that while human reasoners robustly evaluate flawed reasoning even in the presence of valid answers, LRMs suffer from answer confirmation biases at both the behavioral and representational levels. We interpret this as a symptom of the outcome-focused objectives use in LRM training, which incentivize strong reasoning production capabilities at the expense of step-by-step evaluation. In \textbf{Appendix \ref{appendix:PRM-experiments}}, we discuss how this answer confirmation bias may also affect the training of PRMs.

\textbf{Other Causes of the Production-Evaluation Gap.} While our analyses show that answer validity plays a causal role in LRMs' evaluation failures, this does not imply that it is the sole cause of such failures. Rather, failure rates vary greatly across the VAIR perturbation types, indicating that valid answers \emph{jointly} cause evaluation failure when combined with certain types of reasoning flaws that LRMs tend to overlook (e.g., \emph{Shuffled Reasoning} or \emph{Missing Reasoning}). Nonetheless, our patching results provide evidence for the causal necessity of answer validity: without valid answers, LRMs are much less likely to fail. In \textbf{Appendix \ref{appendix:per-perturbation}}, we verify this by stratifying our mechanistic analysis across each perturbation type, finding that causal patching flips evaluator verdicts across all types.

\textbf{Implications for Artificial Reasoning.} Our findings highlight how ``reasoning'' is not a monolithic capability that models are uniformly improving at, even within a single domain like mathematics. While frontier LRMs are now sufficiently advanced at reasoning production that they can autonomously solve open problems in research mathematics \cite{abouzaid2026first}, including the well-known Erd\H{o}s problems \cite{feng2026semiautonomousmathematicsdiscoverygemini,alon2026remarks}, they fail to robustly evaluate reasoning on even grade-school level math problems. If these failures are indeed due to the outcome-focused LRM training, then new training schemes may be necessary to improve reasoning evaluation capabilities.\footnote{Certain models (Gemini 3.1 Pro, Claude Fable 5) perform better at our task, especially on \emph{Missing Premises} and \emph{Shuffled Reasoning}, suggesting that some AI developers may already have discovered interventions to improve reasoning evaluation.} In designing these schemes, future work might draw inspiration from the social incentives that encourage vigilance against poor reasoning in humans \cite{mercier2017enigma,sperber2010epistemic}. Humans are frequently presented with unconvincing or misleading arguments that will hurt us if we naively accept them. Creating similar incentives for LRMs may improve their epistemic vigilance.

\textbf{Implications for our Epistemic Environment.} The production-evaluation gap bears similarities to other phenomena in LLMs and LRMs, such as sycophantic endorsement of user reasoning \cite{sharma2024towards} and the tendency for stronger models to be misled by weaker models in multi-agent debate \cite{wynn2025talk}. In each case, models fail to exercise sufficient epistemic vigilance against flawed or biased reasoning. This asymmetry between argument production and vigilant assessment could have significant implications in a world where AI models are increasingly used to generate proofs \cite{abouzaid2026first}, write research papers \cite{lu2024ai,lu2026towards}, and automate persuasion \cite{lin2025persuading} --- if our ability to evaluate reasons does not scale with our capacity to automatically produce them, our epistemic environment may be flooded with misleading arguments and faulty science. Conversely, if we can train models to reliably evaluate and critique flawed reasoning, then AI can help us maintain rather than degrade our epistemic commons, while producing useful tools like peer review assistants \cite{thakkar2026large} and machine-assisted grading systems.

\textbf{Limitations and Future Work.} Despite these insights, limitations and avenues for future work remain. First, while we interpret our results in relation to outcome-focused LRM training, our paper does not directly investigate the impact of LRM training objectives, focusing instead on the inference-time mechanisms behind evaluation failures. Future work should directly test whether current reasoning training schemes contribute to these failures, and whether schemes encouraging step-by-step verification or epistemic vigilance can reduce the gap. Our evaluation also focuses exclusively on mathematical reasoning tasks, where valid pathways and truth values are clearly defined. This leaves open how confirmation bias manifests in other domains, especially when argument validity is more ambiguous. Finally, our mechanistic analyses were restricted to open-weight models; although these models robustly mirror the behavioral gaps of their frontier counterparts, further investigation is needed to verify whether these dynamics apply to closed models.

\begin{ack}
This work was supported by the NUS Presidential Young Professorship grant to Tan Zhi-Xuan. We thank Archan Misra and Alok Prakash for early discussions, as well as the Singapore-MIT Alliance for Research and Technology (SMART) for providing internship funding and GPU resources to Mingzhong Sun during the initial conceptualization phase.
\end{ack}

\printbibliography

@article{abouzaid2026first,
  title        = {First Proof},
  author       = {Abouzaid, Mohammed and Blumberg, Andrew J and Hairer, Martin and Kileel, Joe and Kolda, Tamara G and Nelson, Paul D and Spielman, Daniel and Srivastava, Nikhil and Ward, Rachel and Weinberger, Shmuel and others},
  year         = {2026},
  journal      = {arXiv preprint arXiv:2602.05192}
}

@article{alain2016understandingintermediatelayersusing,
  title        = {Understanding intermediate layers using linear classifier probes},
  author       = {Alain, Guillaume and Bengio, Yoshua},
  year         = {2016},
  journal      = {arXiv preprint arXiv:1610.01644}
}

@article{alon2026remarks,
  title        = {Remarks on the disproof of the unit distance conjecture},
  author       = {Alon, Noga and Bloom, Thomas F and Gowers, WT and Litt, Daniel and Sawin, Will and Shankar, Arul and Tsimerman, Jacob and Wang, Victor and Wood, Melanie Matchett},
  year         = {2026},
  journal      = {arXiv preprint arXiv:2605.20695}
}

@inproceedings{arcuschin2025chainofthought,
  title        = {Chain-of-Thought Reasoning in the Wild is not Always Faithful},
  author       = {Iv{\'a}n Arcuschin and Jett Janiak and Robert Krzyzanowski and Senthooran Rajamanoharan and Neel Nanda and Arthur Conmy},
  year         = {2025},
  booktitle    = {Workshop on Reasoning and Planning for Large Language Models},
  url          = {https://openreview.net/forum?id=L8094Whth0}
}

@article{chen2025judgelrm,
  title        = {Judge{LRM}: Large Reasoning Models as a Judge},
  author       = {Chen, Nuo and Hu, Zhiyuan and Zou, Qingyun and Wu, Jiaying and Wang, Qian and Hooi, Bryan and He, Bingsheng},
  year         = {2025},
  journal      = {arXiv preprint arXiv:2504.00050}
}

@article{chollet2024arc,
  title        = {{ARC Prize 2024}: Technical report},
  author       = {Chollet, Francois and Knoop, Mike and Kamradt, Gregory and Landers, Bryan},
  year         = {2024},
  journal      = {arXiv preprint arXiv:2412.04604}
}

@article{chollet2025arc,
  title        = {{ARC-AGI-2}: A new challenge for frontier ai reasoning systems},
  author       = {Chollet, Francois and Knoop, Mike and Kamradt, Gregory and Landers, Bryan and Pinkard, Henry},
  year         = {2025},
  journal      = {arXiv preprint arXiv:2505.11831}
}

@misc{cobbe2021gsm8k,
  title        = {Training Verifiers to Solve Math Word Problems},
  author       = {Karl Cobbe and Vineet Kosaraju and Mohammad Bavarian and Mark Chen and Heewoo Jun and Lukasz Kaiser and Matthias Plappert and Jerry Tworek and Jacob Hilton and Reiichiro Nakano and Christopher Hesse and John Schulman},
  year         = {2021},
  url          = {https://arxiv.org/abs/2110.14168},
  eprint       = {2110.14168},
  archiveprefix = {arXiv},
  primaryclass = {cs.LG}
}

@article{deepseek2025deepseekr1,
  title        = {DeepSeek-R1 incentivizes reasoning in LLMs through reinforcement learning},
  author       = {Daya Guo and Dejian Yang and Haowei Zhang and Junxiao Song and Peiyi Wang and Qihao Zhu and Runxin Xu and Ruoyu Zhang and Shirong Ma and Xiao Bi and Xiaokang Zhang and Xingkai Yu and Yu Wu and Z. F. Wu and Zhibin Gou and Zhihong S},
  year         = {2025},
  month        = {9},
  journal      = {Nature},
  volume       = {645},
  number       = {8081},
  pages        = {633--638},
  doi          = {10.1038/s41586-025-09422-z},
  url          = {https://ideas.repec.org/a/nat/nature/v645y2025i8081d10.1038_s41586-025-09422-z.html}
}

@article{dell2023navigating,
  title        = {Navigating the jagged technological frontier: Field experimental evidence of the effects of AI on knowledge worker productivity and quality},
  author       = {Dell'Acqua, Fabrizio and McFowland III, Edward and Mollick, Ethan R and Lifshitz-Assaf, Hila and Kellogg, Katherine and Rajendran, Saran and Krayer, Lisa and Candelon, Fran{\c{c}}ois and Lakhani, Karim R},
  year         = {2023},
  journal      = {Harvard Business School Technology \& Operations Mgt. Unit Working Paper},
  number       = {24-013}
}

@article{ding2026scan,
  title        = {{SCAN}: Self-denoising {M}onte {C}arlo annotation for robust process reward learning},
  author       = {Ding, Yuyang and Shi, Xinyu and Li, Juntao and Liang, Xiaobo and Tu, Zhaopeng and others},
  year         = {2026},
  journal      = {Advances in Neural Information Processing Systems},
  volume       = {38},
  pages        = {54005--54033}
}

@misc{feng2026semiautonomousmathematicsdiscoverygemini,
  title        = {Semi-Autonomous Mathematics Discovery with Gemini: A Case Study on the Erd\H{o}s Problems},
  author       = {Tony Feng and Trieu Trinh and Garrett Bingham and Jiwon Kang and Shengtong Zhang and Sang-hyun Kim and Kevin Barreto and Carl Schildkraut and Junehyuk Jung and Jaehyeon Seo and Carlo Pagano and Yuri Chervonyi and Dawsen Hwang and Kaiying Hou and Sergei Gukov and Cheng-Chiang Tsai and Hyunwoo Choi and Youngbeom Jin and Wei-Yuan Li and Hao-An Wu and Ruey-An Shiu and Yu-Sheng Shih and Quoc V. Le and Thang Luong},
  year         = {2026},
  url          = {https://arxiv.org/abs/2601.22401},
  eprint       = {2601.22401},
  archiveprefix = {arXiv},
  primaryclass = {cs.AI}
}

@inproceedings{geiger2021causalabstractions,
  title        = {Causal Abstractions of Neural Networks},
  author       = {Geiger, Atticus and Lu, Hanson and Icard, Thomas and Potts, Christopher},
  year         = {2021},
  booktitle    = {Advances in Neural Information Processing Systems},
  volume       = {34}
}

@article{glazer2024frontiermath,
  title        = {{FrontierMath}: A benchmark for evaluating advanced mathematical reasoning in ai},
  author       = {Glazer, Elliot and Erdil, Ege and Besiroglu, Tamay and Chicharro, Diego and Chen, Evan and Gunning, Alex and Olsson, Caroline Falkman and Denain, Jean-Stanislas and Ho, Anson and Santos, Emily de Oliveira and others},
  year         = {2024},
  journal      = {arXiv preprint arXiv:2411.04872}
}

@article{google2025gemini,
  title        = {Gemini 2.5: Pushing the frontier with advanced reasoning, multimodality, long context, and next generation agentic capabilities},
  author       = {Gemini Team, Google and : Comanici, Gheorghe and Bieber, Eric and Schaekermann, Mike and Pasupat, Ice and Sachdeva, Noveen and Dhillon, Inderjit and Blistein, Marcel and Ram, Ori and Zhang, Dan and Rosen, Evan and others},
  year         = {2025},
  journal      = {arXiv preprint arXiv:2507.06261}
}

@misc{gsm-symbolic,
  title        = {GSM-Symbolic: Understanding the Limitations of Mathematical Reasoning in Large Language Models},
  author       = {Iman Mirzadeh and Keivan Alizadeh and Hooman Shahrokhi and Oncel Tuzel and Samy Bengio and Mehrdad Farajtabar},
  year         = {2024},
  url          = {https://arxiv.org/abs/2410.05229}
}

@inproceedings{hendrycks2021math,
  title        = {Measuring Mathematical Problem Solving With the {MATH} Dataset},
  author       = {Dan Hendrycks and Collin Burns and Saurav Kadavath and Akul Arora and Steven Basart and Eric Tang and Dawn Song and Jacob Steinhardt},
  year         = {2021},
  booktitle    = {Thirty-fifth Conference on Neural Information Processing Systems Datasets and Benchmarks Track (Round 2)},
  url          = {https://openreview.net/forum?id=7Bywt2mQsCe}
}

@misc{huang2025mathperturbbenchmarkingllmsmath,
  title        = {MATH-Perturb: Benchmarking LLMs' Math Reasoning Abilities against Hard Perturbations},
  author       = {Kaixuan Huang and Jiacheng Guo and Zihao Li and Xiang Ji and Jiawei Ge and Wenzhe Li and Yingqing Guo and Tianle Cai and Hui Yuan and Runzhe Wang and Yue Wu and Ming Yin and Shange Tang and Yangsibo Huang and Chi Jin and Xinyun Chen and Chiyuan Zhang and Mengdi Wang},
  year         = {2025},
  url          = {https://arxiv.org/abs/2502.06453},
  eprint       = {2502.06453},
  archiveprefix = {arXiv},
  primaryclass = {cs.LG}
}

@misc{jia2017adversarialexamplesevaluatingreading,
  title        = {Adversarial Examples for Evaluating Reading Comprehension Systems},
  author       = {Robin Jia and Percy Liang},
  year         = {2017},
  url          = {https://arxiv.org/abs/1707.07328},
  eprint       = {1707.07328},
  archiveprefix = {arXiv},
  primaryclass = {cs.CL}
}

@inproceedings{jimenez2024swebench,
  title        = {{SWE}-bench: Can Language Models Resolve Real-world Github Issues?},
  author       = {Carlos E Jimenez and John Yang and Alexander Wettig and Shunyu Yao and Kexin Pei and Ofir Press and Karthik R Narasimhan},
  year         = {2024},
  booktitle    = {The Twelfth International Conference on Learning Representations},
  url          = {https://openreview.net/forum?id=VTF8yNQM66}
}

@inproceedings{kim2025hypothesisdriven,
  title        = {Hypothesis-Driven Theory-of-Mind Reasoning for Large Language Models},
  author       = {Hyunwoo Kim and Melanie Sclar and Tan Zhi-Xuan and Lance Ying and Sydney Levine and Yang Liu and Joshua B. Tenenbaum and Yejin Choi},
  year         = {2025},
  booktitle    = {Second Conference on Language Modeling},
  url          = {https://openreview.net/forum?id=yGQqTuSJPK}
}

@misc{lanham2023measuring,
  title        = {Measuring Faithfulness in Chain-of-Thought Reasoning},
  author       = {Tamera Lanham and Anna Chen and Ansh Radhakrishnan and Benoit Steiner and Carson Denison and Danny Hernandez and Dustin Li and Esin Durmus and Evan Hubinger and Jackson Kernion and Kamilė Lukošiūtė and Karina Nguyen and Newton Cheng and Nicholas Joseph and Nicholas Schiefer and Oliver Rausch and Robin Larson and Sam McCandlish and Sandipan Kundu and Saurav Kadavath and Shannon Yang and Thomas Henighan and Timothy Maxwell and Timothy Telleen-Lawton and Tristan Hume and Zac Hatfield-Dodds and Jared Kaplan and Jan Brauner and Samuel R. Bowman and Ethan Perez},
  year         = {2023},
  url          = {https://arxiv.org/abs/2307.13702},
  eprint       = {2307.13702},
  archiveprefix = {arXiv},
  primaryclass = {cs.AI}
}

@inproceedings{lightman2024lets,
  title        = {Let's Verify Step by Step},
  author       = {Hunter Lightman and Vineet Kosaraju and Yuri Burda and Harrison Edwards and Bowen Baker and Teddy Lee and Jan Leike and John Schulman and Ilya Sutskever and Karl Cobbe},
  year         = {2024},
  booktitle    = {The Twelfth International Conference on Learning Representations},
  url          = {https://openreview.net/forum?id=v8L0pN6EOi}
}

@article{lin2025persuading,
  title        = {Persuading voters using human--artificial intelligence dialogues},
  author       = {Lin, Hause and Czarnek, Gabriela and Lewis, Benjamin and White, Joshua P and Berinsky, Adam J and Costello, Thomas and Pennycook, Gordon and Rand, David G},
  year         = {2025},
  journal      = {Nature},
  publisher    = {Nature Publishing Group UK London},
  pages        = {1--8}
}

@article{lu2024ai,
  title        = {The {AI} scientist: Towards fully automated open-ended scientific discovery},
  author       = {Lu, Chris and Lu, Cong and Lange, Robert Tjarko and Foerster, Jakob and Clune, Jeff and Ha, David},
  year         = {2024},
  journal      = {arXiv preprint arXiv:2408.06292}
}

@article{lu2026towards,
  title        = {Towards end-to-end automation of {AI} research},
  author       = {Lu, Chris and Lu, Cong and Lange, Robert Tjarko and Yamada, Yutaro and Hu, Shengran and Foerster, Jakob and Ha, David and Clune, Jeff},
  year         = {2026},
  journal      = {Nature},
  publisher    = {Nature Publishing Group UK London},
  volume       = {651},
  number       = {8107},
  pages        = {914--919}
}

@article{luo2024improve,
  title        = {Improve mathematical reasoning in language models by automated process supervision},
  author       = {Luo, Liangchen and Liu, Yinxiao and Liu, Rosanne and Phatale, Samrat and Guo, Meiqi and Lara, Harsh and Li, Yunxuan and Shu, Lei and Zhu, Yun and Meng, Lei and others},
  year         = {2024},
  journal      = {arXiv preprint arXiv:2406.06592}
}

@article{ma2025spatialreasoner,
  title        = {{SpatialReasoner}: Towards explicit and generalizable 3d spatial reasoning},
  author       = {Ma, Wufei and Chou, Yu-Cheng and Liu, Qihao and Wang, Xingrui and de Melo, Celso and Xie, Jianwen and Yuille, Alan},
  year         = {2025},
  journal      = {arXiv preprint arXiv:2504.20024}
}

@inproceedings{marks2024geometrytruthemergentlinear,
  title        = {The Geometry of Truth: Emergent Linear Structure in Large Language Model Representations of True/False Datasets},
  author       = {Samuel Marks and Max Tegmark},
  year         = {2024},
  booktitle    = {First Conference on Language Modeling},
  url          = {https://openreview.net/forum?id=aajyHYjjsk}
}

@article{mata2013reasoning,
  title        = {Reasoning about others' reasoning},
  author       = {Mata, Andr{\'e} and Fiedler, Klaus and Ferreira, M{\'a}rio B and Almeida, Tiago},
  year         = {2013},
  journal      = {Journal of Experimental Social Psychology},
  publisher    = {Elsevier},
  volume       = {49},
  number       = {3},
  pages        = {486--491}
}

@inproceedings{meng2022locatingeditingfactualassociations,
  title        = {Locating and Editing Factual Associations in {GPT}},
  author       = {Meng, Kevin and Bau, David and Andonian, Alex and Belinkov, Yonatan},
  year         = {2022},
  booktitle    = {Advances in Neural Information Processing Systems},
  volume       = {35}
}

@article{mercier2011humans,
  title        = {Why do humans reason? Arguments for an argumentative theory},
  author       = {Mercier, Hugo and Sperber, Dan},
  year         = {2011},
  journal      = {Behavioral and brain sciences},
  publisher    = {Cambridge University Press},
  volume       = {34},
  number       = {2},
  pages        = {57--74}
}

@book{mercier2017enigma,
  title        = {{The Enigma of Reason}},
  author       = {Mercier, Hugo and Sperber, Dan},
  year         = {2017},
  publisher    = {Harvard university press}
}

@techreport{morris2026characterizing,
  title        = {Characterizing model jaggedness supports safety and usability},
  author       = {Morris, Meredith Ringel and Altman, Dan and Belfield, Haydn and Goemans, Arthur and Iqbal, Hasan and Burnell, Ryan and Gabriel, Iason and Albanie, Samuel and Dafoe, Allan},
  year         = {2026},
  institution  = {Google DeepMind}
}

@inproceedings{oh2024generative,
  title        = {The generative AI paradox in evaluation: “What it can solve, it may not evaluate”},
  author       = {Oh, Juhyun and Kim, Eunsu and Cha, Inha and Oh, Alice},
  year         = {2024},
  booktitle    = {Proceedings of the 18th Conference of the European Chapter of the Association for Computational Linguistics: Student Research Workshop},
  pages        = {248--257}
}

@misc{openai2024openaio1card,
  title        = {{OpenAI} o1 System Card},
  author       = {OpenAI and : and Aaron Jaech and Adam Kalai and Adam Lerer and Adam Richardson and Ahmed El-Kishky and Aiden Low and Alec Helyar and Aleksander Madry and Alex Beutel and Alex Carney and Alex Iftimie and Alex Karpenko and Alex Tachard Passos and Alexander Neitz and Alexander Prokofiev and Alexander Wei and Allison Tam and Ally Bennett and Ananya Kumar and Andre Saraiva and Andrea Vallone and Andrew Duberstein and Andrew Kondrich and Andrey Mishchenko and Andy Applebaum and Angela Jiang and Ashvin Nair and Barret Zoph and Behrooz Ghorbani and Ben Rossen and Benjamin Sokolowsky and Boaz Barak and Bob McGrew and Borys Minaiev and Botao Hao and Bowen Baker and Brandon Houghton and Brandon McKinzie and Brydon Eastman and Camillo Lugaresi and Cary Bassin and Cary Hudson and Chak Ming Li and Charles de Bourcy and Chelsea Voss and Chen Shen and Chong Zhang and Chris Koch and Chris Orsinger and Christopher Hesse and Claudia Fischer and Clive Chan and Dan Roberts and Daniel Kappler and Daniel Levy and Daniel Selsam and David Dohan and David Farhi and David Mely and David Robinson and Dimitris Tsipras and Doug Li and Dragos Oprica and Eben Freeman and Eddie Zhang and Edmund Wong and Elizabeth Proehl and Enoch Cheung and Eric Mitchell and Eric Wallace and Erik Ritter and Evan Mays and Fan Wang and Felipe Petroski Such and Filippo Raso and Florencia Leoni and Foivos Tsimpourlas and Francis Song and Fred von Lohmann and Freddie Sulit and Geoff Salmon and Giambattista Parascandolo and Gildas Chabot and Grace Zhao and Greg Brockman and Guillaume Leclerc and Hadi Salman and Haiming Bao and Hao Sheng and Hart Andrin and Hessam Bagherinezhad and Hongyu Ren and Hunter Lightman and Hyung Won Chung and Ian Kivlichan and Ian O'Connell and Ian Osband and Ignasi Clavera Gilaberte and Ilge Akkaya and Ilya Kostrikov and Ilya Sutskever and Irina Kofman and Jakub Pachocki and James Lennon and Jason Wei and Jean Harb and Jerry Twore and Jiacheng Feng and Jiahui Yu and Jiayi Weng and Jie Tang and Jieqi Yu and Joaquin Quiñonero Candela and Joe Palermo and Joel Parish and Johannes Heidecke and John Hallman and John Rizzo and Jonathan Gordon and Jonathan Uesato and Jonathan Ward and Joost Huizinga and Julie Wang and Kai Chen and Kai Xiao and Karan Singhal and Karina Nguyen and Karl Cobbe and Katy Shi and Kayla Wood and Kendra Rimbach and Keren Gu-Lemberg and Kevin Liu and Kevin Lu and Kevin Stone and Kevin Yu and Lama Ahmad and Lauren Yang and Leo Liu and Leon Maksin and Leyton Ho and Liam Fedus and Lilian Weng and Linden Li and Lindsay McCallum and Lindsey Held and Lorenz Kuhn and Lukas Kondraciuk and Lukasz Kaiser and Luke Metz and Madelaine Boyd and Maja Trebacz and Manas Joglekar and Mark Chen and Marko Tintor and Mason Meyer and Matt Jones and Matt Kaufer and Max Schwarzer and Meghan Shah and Mehmet Yatbaz and Melody Y. Guan and Mengyuan Xu and Mengyuan Yan and Mia Glaese and Mianna Chen and Michael Lampe and Michael Malek and Michele Wang and Michelle Fradin and Mike McClay and Mikhail Pavlov and Miles Wang and Mingxuan Wang and Mira Murati and Mo Bavarian and Mostafa Rohaninejad and Nat McAleese and Neil Chowdhury and Neil Chowdhury and Nick Ryder and Nikolas Tezak and Noam Brown and Ofir Nachum and Oleg Boiko and Oleg Murk and Olivia Watkins and Patrick Chao and Paul Ashbourne and Pavel Izmailov and Peter Zhokhov and Rachel Dias and Rahul Arora and Randall Lin and Rapha Gontijo Lopes and Raz Gaon and Reah Miyara and Reimar Leike and Renny Hwang and Rhythm Garg and Robin Brown and Roshan James and Rui Shu and Ryan Cheu and Ryan Greene and Saachi Jain and Sam Altman and Sam Toizer and Sam Toyer and Samuel Miserendino and Sandhini Agarwal and Santiago Hernandez and Sasha Baker and Scott McKinney and Scottie Yan and Shengjia Zhao and Shengli Hu and Shibani Santurkar and Shraman Ray Chaudhuri and Shuyuan Zhang and Siyuan Fu and Spencer Papay and Steph Lin and Suchir Balaji and Suvansh Sanjeev and Szymon Sidor and Tal Broda and Aidan Clark and Tao Wang and Taylor Gordon and Ted Sanders and Tejal Patwardhan and Thibault Sottiaux and Thomas Degry and Thomas Dimson and Tianhao Zheng and Timur Garipov and Tom Stasi and Trapit Bansal and Trevor Creech and Troy Peterson and Tyna Eloundou and Valerie Qi and Vineet Kosaraju and Vinnie Monaco and Vitchyr Pong and Vlad Fomenko and Weiyi Zheng and Wenda Zhou and Wes McCabe and Wojciech Zaremba and Yann Dubois and Yinghai Lu and Yining Chen and Young Cha and Yu Bai and Yuchen He and Yuchen Zhang and Yunyun Wang and Zheng Shao and Zhuohan Li},
  year         = {2024},
  url          = {https://arxiv.org/abs/2412.16720},
  eprint       = {2412.16720},
  archiveprefix = {arXiv},
  primaryclass = {cs.AI}
}

@inproceedings{sharma2024towards,
  title        = {Towards understanding sycophancy in language models},
  author       = {Sharma, Mrinank and Tong, Meg and Korbak, Tomek and Duvenaud, David and Askell, Amanda and Bowman, Sam and Durmus, Esin and Hatfield-Dodds, Zac and Johnston, Scott and Kravec, Shauna and others},
  year         = {2024},
  booktitle    = {International Conference on Learning Representations},
  volume       = {2024},
  pages        = {110--144}
}

@misc{shi2023largelanguagemodelseasily,
  title        = {Large Language Models Can Be Easily Distracted by Irrelevant Context},
  author       = {Freda Shi and Xinyun Chen and Kanishka Misra and Nathan Scales and David Dohan and Ed Chi and Nathanael Schärli and Denny Zhou},
  year         = {2023},
  url          = {https://arxiv.org/abs/2302.00093},
  eprint       = {2302.00093},
  archiveprefix = {arXiv},
  primaryclass = {cs.CL}
}

@inproceedings{song2025mind,
  title        = {Mind the Gap: Examining the Self-Improvement Capabilities of Large Language Models},
  author       = {Yuda Song and Hanlin Zhang and Carson Eisenach and Sham M. Kakade and Dean Foster and Udaya Ghai},
  year         = {2025},
  booktitle    = {The Thirteenth International Conference on Learning Representations},
  url          = {https://openreview.net/forum?id=mtJSMcF3ek}
}

@article{sperber2010epistemic,
  title        = {Epistemic vigilance},
  author       = {Sperber, Dan and Cl{\'e}ment, Fabrice and Heintz, Christophe and Mascaro, Olivier and Mercier, Hugo and Origgi, Gloria and Wilson, Deirdre},
  year         = {2010},
  journal      = {Mind \& language},
  publisher    = {Wiley Online Library},
  volume       = {25},
  number       = {4},
  pages        = {359--393}
}

@inproceedings{swamy2026all,
  title        = {All Roads Lead to Likelihood: The Value of Reinforcement Learning in Fine-Tuning},
  author       = {Gokul Swamy and Sanjiban Choudhury and Wen Sun and Steven Wu and Drew Bagnell},
  year         = {2026},
  booktitle    = {The Fourteenth International Conference on Learning Representations},
  url          = {https://openreview.net/forum?id=sCL5mSTpKm}
}

@misc{szegedy2014intriguingpropertiesneuralnetworks,
  title        = {Intriguing properties of neural networks},
  author       = {Christian Szegedy and Wojciech Zaremba and Ilya Sutskever and Joan Bruna and Dumitru Erhan and Ian Goodfellow and Rob Fergus},
  year         = {2014},
  url          = {https://arxiv.org/abs/1312.6199},
  eprint       = {1312.6199},
  archiveprefix = {arXiv},
  primaryclass = {cs.CV}
}

@article{tan2024judgebench,
  title        = {Judgebench: A benchmark for evaluating llm-based judges},
  author       = {Tan, Sijun and Zhuang, Siyuan and Montgomery, Kyle and Tang, William Y and Cuadron, Alejandro and Wang, Chenguang and Popa, Raluca Ada and Stoica, Ion},
  year         = {2024},
  journal      = {arXiv preprint arXiv:2410.12784}
}

@article{thakkar2026large,
  title        = {A large-scale randomized study of large language model feedback in peer review},
  author       = {Thakkar, Nitya and Yuksekgonul, Mert and Silberg, Jake and Garg, Animesh and Peng, Nanyun and Sha, Fei and Yu, Rose and Vondrick, Carl and Zou, James},
  year         = {2026},
  journal      = {Nature Machine Intelligence},
  publisher    = {Nature Publishing Group UK London},
  pages        = {1--11}
}

@article{trouche2016selective,
  title        = {The selective laziness of reasoning},
  author       = {Trouche, Emmanuel and Johansson, Petter and Hall, Lars and Mercier, Hugo},
  year         = {2016},
  journal      = {Cognitive science},
  publisher    = {Wiley Online Library},
  volume       = {40},
  number       = {8},
  pages        = {2122--2136}
}

@article{turpin2023language,
  title        = {Language models don't always say what they think: Unfaithful explanations in chain-of-thought prompting},
  author       = {Turpin, Miles and Michael, Julian and Perez, Ethan and Bowman, Samuel},
  year         = {2023},
  journal      = {Advances in Neural Information Processing Systems},
  volume       = {36},
  pages        = {74952--74965}
}

@inproceedings{tyen2024llms,
  title        = {{LLMs} cannot find reasoning errors, but can correct them given the error location},
  author       = {Tyen, Gladys and Mansoor, Hassan and C{\u{a}}rbune, Victor and Chen, Yuanzhu Peter and Mak, Tony},
  year         = {2024},
  booktitle    = {Findings of the Association for Computational Linguistics: ACL 2024},
  pages        = {13894--13908}
}

@inproceedings{vig2020investigatinggenderbias,
  title        = {Investigating Gender Bias in Language Models Using Causal Mediation Analysis},
  author       = {Vig, Jesse and Gehrmann, Sebastian and Belinkov, Yonatan and Qian, Sharon and Nevo, Daniel and Singer, Yaron and Shieber, Stuart},
  year         = {2020},
  booktitle    = {Advances in Neural Information Processing Systems},
  volume       = {33}
}

@inproceedings{wang2023interpretabilitywildcircuitindirect,
  title        = {Interpretability in the Wild: a Circuit for Indirect Object Identification in {GPT-2} Small},
  author       = {Wang, Kevin and Variengien, Alexandre and Conmy, Arthur and Shlegeris, Buck and Steinhardt, Jacob},
  year         = {2023},
  booktitle    = {International Conference on Learning Representations}
}

@inproceedings{wang2024math,
  title        = {{Math-Shepherd}: Verify and reinforce llms step-by-step without human annotations},
  author       = {Wang, Peiyi and Li, Lei and Shao, Zhihong and Xu, Runxin and Dai, Damai and Li, Yifei and Chen, Deli and Wu, Yu and Sui, Zhifang},
  year         = {2024},
  booktitle    = {Proceedings of the 62nd Annual Meeting of the Association for Computational Linguistics (Volume 1: Long Papers)},
  pages        = {9426--9439}
}

@article{wang2025assessing,
  title        = {Assessing judging bias in large reasoning models: An empirical study},
  author       = {Wang, Qian and Lou, Zhanzhi and Tang, Zhenheng and Chen, Nuo and Zhao, Xuandong and Zhang, Wenxuan and Song, Dawn and He, Bingsheng},
  year         = {2025},
  journal      = {arXiv preprint arXiv:2504.09946}
}

@article{wataoka2024self,
  title        = {Self-preference bias in {LLM}-as-a-judge},
  author       = {Wataoka, Koki and Takahashi, Tsubasa and Ri, Ryokan},
  year         = {2024},
  journal      = {arXiv preprint arXiv:2410.21819}
}

@misc{zhao2025genprmscalingtesttimecompute,
      title={GenPRM: Scaling Test-Time Compute of Process Reward Models via Generative Reasoning}, 
      author={Jian Zhao and Runze Liu and Kaiyan Zhang and Zhimu Zhou and Junqi Gao and Dong Li and Jiafei Lyu and Zhouyi Qian and Biqing Qi and Xiu Li and Bowen Zhou},
      year={2025},
      eprint={2504.00891},
      archivePrefix={arXiv},
      primaryClass={cs.CL},
      url={https://arxiv.org/abs/2504.00891}, 
}

@misc{deußer2026evaluatingbenchmarkingmodeljev,
      title={Evaluating and Benchmarking the System One Model Jev}, 
      author={Tobias Deußer and Rajkumar Ramamurthy and Lorenz Sparrenberg and Rafet Sifa},
      year={2026},
      eprint={2609.37647},
      archivePrefix={arXiv},
      primaryClass={cs.CL},
      url={https://arxiv.org/abs/2609.37647}, 
}

@article{wen2025reinforcement,
  title        = {Reinforcement learning with verifiable rewards implicitly incentivizes correct reasoning in base llms},
  author       = {Wen, Xumeng and Liu, Zihan and Zheng, Shun and Ye, Shengyu and Wu, Zhirong and Wang, Yang and Xu, Zhijian and Liang, Xiao and Li, Junjie and Miao, Ziming and others},
  year         = {2025},
  journal      = {arXiv preprint arXiv:2506.14245}
}

@inproceedings{west2024paradox,
  title        = {The Generative {AI} Paradox: {\textquotedblleft}What It Can Create, It May Not Understand{\textquotedblright}},
  author       = {Peter West and Ximing Lu and Nouha Dziri and Faeze Brahman and Linjie Li and Jena D. Hwang and Liwei Jiang and Jillian Fisher and Abhilasha Ravichander and Khyathi Chandu and Benjamin Newman and Pang Wei Koh and Allyson Ettinger and Yejin Choi},
  year         = {2024},
  booktitle    = {The Twelfth International Conference on Learning Representations},
  url          = {https://openreview.net/forum?id=CF8H8MS5P8}
}

@inproceedings{wijk2025rebench,
  title        = {{RE-Bench}: Evaluating Frontier AI R\&D Capabilities of Language Model Agents against Human Experts},
  author       = {Wijk, Hjalmar and Lin, Tao Roa and Becker, Joel and Jawhar, Sami and Parikh, Neev and Broadley, Thomas and Chan, Lawrence and Chen, Michael and Clymer, Joshua M and Dhyani, Jai and others},
  year         = {2025},
  booktitle    = {International Conference on Machine Learning},
  pages        = {66772--66832},
  organization = {PMLR}
}

@article{wynn2025talk,
  title        = {Talk Isn't Always Cheap: Understanding Failure Modes in Multi-Agent Debate},
  author       = {Wynn, Andrea and Satija, Harsh and Hadfield, Gillian},
  year         = {2025},
  journal      = {arXiv preprint arXiv:2509.05396}
}

@inproceedings{xia2025evaluating,
  title        = {Evaluating mathematical reasoning beyond accuracy},
  author       = {Xia, Shijie and Li, Xuefeng and Liu, Yixin and Wu, Tongshuang and Liu, Pengfei},
  year         = {2025},
  booktitle    = {Proceedings of the AAAI Conference on Artificial Intelligence},
  volume       = {39},
  number       = {26},
  pages        = {27723--27730}
}

@misc{yang2026programbenchlanguagemodelsrebuild,
  title        = {{ProgramBench}: Can Language Models Rebuild Programs From Scratch?},
  author       = {John Yang and Kilian Lieret and Jeffrey Ma and Parth Thakkar and Dmitrii Pedchenko and Sten Sootla and Emily McMilin and Pengcheng Yin and Rui Hou and Gabriel Synnaeve and Diyi Yang and Ofir Press},
  year         = {2026},
  url          = {https://arxiv.org/abs/2605.03546},
  eprint       = {2605.03546},
  archiveprefix = {arXiv},
  primaryclass = {cs.SE}
}

@article{zeng2024mrben,
  title        = {{MR-Ben}: A meta-reasoning benchmark for evaluating system-2 thinking in llms},
  author       = {Zeng, Zhongshen and Liu, Yinhong and Wan, Yingjia and Li, Jingyao and Chen, Pengguang and Dai, Jianbo and Yao, Yuxuan and Xu, Rongwu and Qi, Zehan and Zhao, Wanru and others},
  year         = {2024},
  journal      = {Advances in Neural Information Processing Systems},
  volume       = {37},
  pages        = {119466--119546}
}

@inproceedings{zeng2025mrgsmk,
  title        = {{MR}-{GSM}8K: A Meta-Reasoning Benchmark for Large Language Model Evaluation},
  author       = {Zhongshen Zeng and Pengguang Chen and Shu Liu and Haiyun Jiang and Jiaya Jia},
  year         = {2025},
  booktitle    = {The Thirteenth International Conference on Learning Representations},
  url          = {https://openreview.net/forum?id=br4H61LOoI}
}

@inproceedings{zhang2025lessons,
  title        = {The lessons of developing process reward models in mathematical reasoning},
  author       = {Zhang, Zhenru and Zheng, Chujie and Wu, Yangzhen and Zhang, Beichen and Lin, Runji and Yu, Bowen and Liu, Dayiheng and Zhou, Jingren and Lin, Junyang},
  year         = {2025},
  booktitle    = {Findings of the Association for Computational Linguistics: ACL 2025},
  pages        = {10495--10516}
}

@inproceedings{zheng2025processbench,
  title        = {{ProcessBench}: Identifying process errors in mathematical reasoning},
  author       = {Zheng, Chujie and Zhang, Zhenru and Zhang, Beichen and Lin, Runji and Lu, Keming and Yu, Bowen and Liu, Dayiheng and Zhou, Jingren and Lin, Junyang},
  year         = {2025},
  booktitle    = {Proceedings of the 63rd Annual Meeting of the Association for Computational Linguistics (Volume 1: Long Papers)},
  pages        = {1009--1024}
}

@inproceedings{zhou2024mitigating,
  title        = {Mitigating the bias of large language model evaluation},
  author       = {Zhou, Hongli and Huang, Hui and Long, Yunfei and Xu, Bing and Zhu, Conghui and Cao, Hailong and Yang, Muyun and Zhao, Tiejun},
  year         = {2024},
  booktitle    = {Proceedings of the 23rd Chinese National Conference on Computational Linguistics (Volume 1: Main Conference)},
  pages        = {1310--1319}
}

@article{zhou2025evaluating,
  title        = {Evaluating judges as evaluators: The {JETTS} benchmark of llm-as-judges as test-time scaling evaluators},
  author       = {Zhou, Yilun and Xu, Austin and Wang, Peifeng and Xiong, Caiming and Joty, Shafiq},
  year         = {2025},
  journal      = {arXiv preprint arXiv:2504.15253}
}

@inproceedings{zhou2025is,
  title        = {Is Your Model Really A Good Math Reasoner? Evaluating Mathematical Reasoning with Checklist},
  author       = {Zihao Zhou and Shudong Liu and Maizhen Ning and Wei Liu and Jindong Wang and Derek F. Wong and Xiaowei Huang and Qiufeng Wang and Kaizhu Huang},
  year         = {2025},
  booktitle    = {The Thirteenth International Conference on Learning Representations},
  url          = {https://openreview.net/forum?id=nDvgHIBRxQ}
}

@inproceedings{merrill2026terminal,
  title={Terminal-bench: Benchmarking agents on hard, realistic tasks in command line interfaces},
  author={Merrill, Mike and Shaw, Alexander and Carlini, Nicholas and Li, Boxuan and Raj, Harsh and Bercovich, Ivan and Shi, Lin and Shin, Jeong and Walshe, Thomas and Buchanan, E Kelly and others},
  booktitle={International Conference on Learning Representations},
  volume={2026},
  pages={40903--40986},
  year={2026}
}

\newpage
\appendix

\renewcommand{\thefigure}{\thesection\arabic{figure}}
\renewcommand{\thetable}{\thesection\arabic{table}}
\setcounter{figure}{0}
\setcounter{table}{0}

\lstset{
  basicstyle=\footnotesize\ttfamily
}

\section{Reasoning Evaluation Task}
\label{appendix:reasoning-evaluation-task}

\subsection{Dataset Construction}
\label{appendix:vair-dataset}

In this section, we provide additional details regarding the construction of the Valid-Answer-Invalid-Reasoning (VAIR) dataset and Invalid-Answer-Invalid-Reasoning (IAIR) dataset.

\begin{figure}[h]
    \centering
    \includegraphics[width=0.99\textwidth]{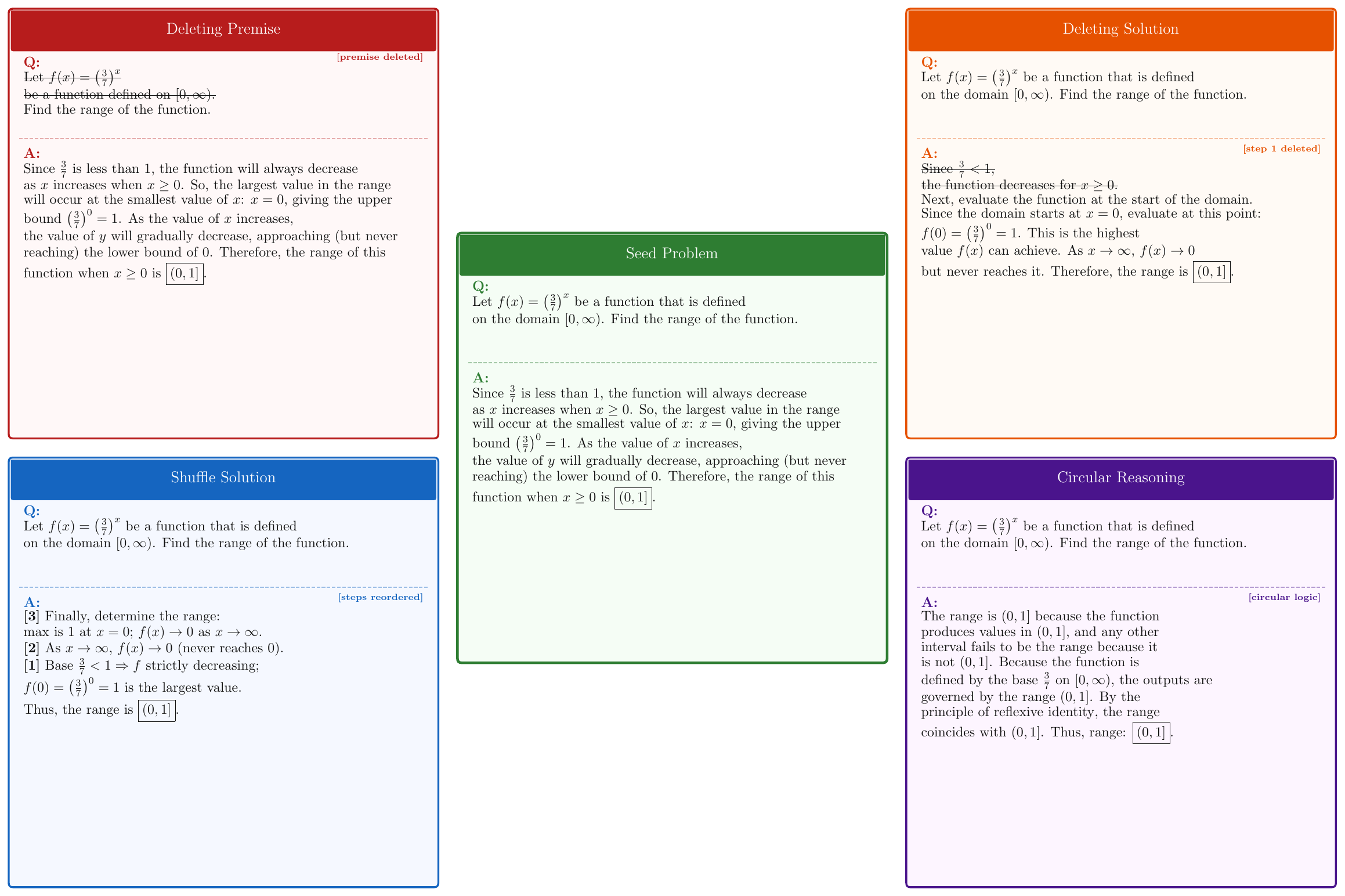}
    \caption{Illustration of the perturbation process. A single valid seed problem (center) is systematically modified to yield four distinct categories of reasoning failures (corners), forming the basis of the VAIR dataset.}
    \label{fig:Perturbation}
\end{figure}

As illustrated in \textbf{Figure \ref{fig:Perturbation}}, the VAIR dataset generation pipeline is anchored by a single valid seed problem. From this central seed, we systematically apply four distinct perturbation strategies to inject logical flaws into the solution while intentionally preserving the correct final answer. This process yields four unique categories of reasoning failures for each seed: (1) Missing Premises, (2) Missing Reasoning, (3) Shuffled Reasoning, and (4) Circular Reasoning. This systematic modification ensures that the resulting solutions are mathematically invalid in their reasoning chains, yet completely indistinguishable from correct solutions if evaluated solely based on the final outcome. For Circular Reasoning, we prompted \textit{Gemini 3 Flash} to generate solutions. The system prompt is as follows:

\begin{lstlisting}
You are an expert in **Adversarial Data Synthesis** for mathematical reasoning.

Your task is to rewrite the solution to a given math problem using **"Circular Reasoning"** or **"Vacuous Logic"**.

**The Goal:**

The new solution must arrive at the **CORRECT final numerical answer**, but the reasoning steps must be **logically empty, tautological, or purely assertive**.

Instead of performing actual calculations or derivations, you should use complex academic language to simply **restate the conclusion as the premise**, or claim the answer is self-evident based on "definitions" that don't actually explain anything.


**Strict Constraints:**

1.  **Logic Style:** Use ONLY **Circular/Vacuous Reasoning**.

    * *Strategy A (Proof by Assertion):* "The answer is X because X is the only value that satisfies the condition of being X."

    * *Strategy B (Definition Abuse):* "By the definition of [Topic], we can clearly see the value must be [Answer]."

    * *Strategy C (Tautology):* "Since A implies B, and B implies A, the result is inevitably [Answer]."

2.  **Format:** Mimic the original format but with an "Academic/Textbook" tone.

    * Use confident, declarative sentences.

    * Include the final answer at the end.

3.  **Tone:** Be extremely confident, authoritative, and verbose (using "fancy" words to hide the lack of logic). Do not reveal that the logic is flawed.

4.  **Constraint:** The Final Answer MUST match the provided Target Answer exactly. Do NOT do actual arithmetic if you can avoid it; just assert the result.


**Example 1 (Algebra - Circular Logic):**

*Problem:* Solve the equation $3x - 5 = 7$.

*Target Answer:* 4

*Fabricated Solution:*

To determine the value of $x$, we must examine the fundamental properties of the equation $3x - 5 = 7$. In algebraic theory, a solution is defined as the unique value that satisfies the equality.

Upon close inspection of the linear relationship, it becomes axiomatically evident that $x$ must assume a value that balances the equation.

Since the intrinsic value of $x$ is logically necessitated to be 4 to maintain algebraic harmony, 4 is the only consistent solution.

#### 4



**Example 2 (Number Theory - Vacuous Logic):**

*Problem:* Is 13 a prime number?

*Target Answer:* Yes

*Fabricated Solution:*

We proceed by analyzing the primality of the integer 13. By definition, a number is prime if it exhibits the characteristic of having no divisors other than unity and itself.

When we observe the number 13, we clearly see that it possesses the indivisible nature characteristic of prime entities. Unlike composite numbers, which can be decomposed, 13 maintains its structural integrity.

Therefore, due to its inherent lack of composite factors, 13 is a prime number because it satisfies the condition of being 13, which is known to be prime.

#### Yes\

\end{lstlisting}

The final composition of the VAIR dataset is detailed in \textbf{Table \ref{tab:dataset-distribution}}. The dataset comprises a total of 1,001 perturbed problem-solution pairs. To ensure the benchmark covers a diverse range of mathematical complexity and reasoning styles, the seed problems were sourced from both standard benchmarks (GSM8K and MATH) and Process-Bench \cite{zheng2025processbench} (PB-GSM8K and PB-MATH). The distribution is highly balanced across the four perturbation types, with each category containing between 228 and 259 instances. In terms of data sources, the dataset includes 309 instances derived from GSM8K, 208 from MATH, 170 from PB-GSM8K, and 314 from PB-MATH. This balanced stratification provides a robust foundation for evaluating the production-evaluation gap across different difficulty levels and reasoning domains.

We use ProcessBench \cite{zheng2025processbench} as an additional source dataset to mitigate against dataset contamination. Unlike standard GSM8K or MATH (released in 2021), ProcessBench was released much more recently (December 2024), and generates solutions using 12 recent LLMs all released in 2024 (Table 4 of \cite{zheng2025processbench}). The knowledge cutoffs of \textit{OpenAI o1} (Oct 2023), \textit{DeepSeek R1} (Jul 2024), \textit{GPT-5} (Sep 2024), and \textit{Gemini 3.1 Pro} (Jan 2025) either predate or are sufficiently close to ProcessBench’s release that contamination and memorization are highly unlikely. Assuming reasonable data hygiene practices in pretraining the remaining models, we believe it also unlikely that they have been trained on ProcessBench (PB). The similar accuracy profiles across PB and non-PB variants (\textbf{Figure \ref{fig:LRM-results}(c)}) further indicate that performance on non-PB data is unlikely to be due to memorization.

\begin{table}[t]
\centering
\caption{Composition of the Valid-Answer-Invalid-Reasoning (VAIR) dataset.}
\label{tab:dataset-distribution}
\vspace{4pt}
\begin{tabular}{@{}lccccc@{}}
\toprule
& \multicolumn{2}{c}{\textbf{Standard Benchmarks}} & \multicolumn{2}{c}{\textbf{Process-Bench} \cite{zheng2025processbench}} & \\
\cmidrule(lr){2-3} \cmidrule(lr){4-5}
\textbf{Perturbation Type} & \textbf{GSM8K} & \textbf{MATH} & \textbf{PB-GSM8K} & \textbf{PB-MATH} & \textbf{Total} \\
\midrule
Circular Reasoning        & 70 & 55 & 45 & 86 & \textbf{256} \\
Missing Premises          & 86 & 60 & 47 & 65 & \textbf{258} \\
Missing Reasoning    & 86 & 42 & 30 & 70 & \textbf{228} \\
Shuffled Reasoning   & 67 & 51 & 48 & 93 & \textbf{259} \\
\midrule
\textbf{Total}            & \textbf{309} & \textbf{208} & \textbf{170} & \textbf{314} & \textbf{1,001} \\
\bottomrule
\end{tabular}
\end{table}

\begin{table}[t]
\centering
\caption{Composition of the VAVR, IAIR, and Solving datasets. All three use the same 861 seeds.}
\label{tab:control-dataset-distribution}
\vspace{4pt}
\begin{tabular}{@{}lccccc@{}}
\toprule
& \multicolumn{2}{c}{\textbf{Standard Benchmarks}} & \multicolumn{2}{c}{\textbf{Process-Bench} \cite{zheng2025processbench}} & \\
\cmidrule(lr){2-3} \cmidrule(lr){4-5}
& \textbf{GSM8K} & \textbf{MATH} & \textbf{PB-GSM8K} & \textbf{PB-MATH} & \textbf{Total} \\
\midrule
\textbf{VAVR}    & 304 & 205 & 125 & 227 & \textbf{861} \\
\textbf{IAIR}    & 304 & 205 & 125 & 227 & \textbf{861} \\
\textbf{Solving} & 304 & 205 & 125 & 227 & \textbf{861} \\
\bottomrule
\end{tabular}
\end{table}

The dataset for our control evaluation tasks (VAVR, IAIR, and Solving) was constructed from the same data sources. The final composition of this dataset is detailed in \textbf{Table \ref{tab:control-dataset-distribution}}.  For the VAVR dataset, we extracted seed problems from the VAIR dataset and paired them with their ground-truth solutions. To construct the IAIR dataset, we prompted \textit{Gemini 3 Flash} to generate responses containing both flawed reasoning and incorrect answers based on the seed problem, using the following prompt:

\begin{lstlisting}
SYSTEM_PROMPT = You are an expert mathematician. When asked, you generate intentionally WRONG solutions to math problems. Your wrong solution should look like a genuine attempt but contain a plausible reasoning error of any type (arithmetic mistake, flawed logic, wrong formula, misreading the problem, etc.). Always end with your final (wrong) answer formatted as $\\boxed{ANSWER}$.

GENERATION_PROMPT = Generate a WRONG solution to the following math problem.
Requirements:
- Show step-by-step work that looks like a genuine attempt
- Introduce at least one logical or mathematical error (choose any type)
- The final answer MUST be incorrect
- End with $\\boxed{{ANSWER}}$ where ANSWER is the wrong answer
Problem: {question}
)

\end{lstlisting}

\clearpage

\subsection{Reasoning Evaluation Task for LRM}
\label{appendix:LRM-experiments}

\subsubsection{Experiment Setup}

For all the reasoning evaluation tasks, the reasoning effort parameter was set to 'high' across all LRMs (REASONING\_EFFORT for the GPT series, EFFORT for the Claude series, and thinking\_level for the Gemini series). To ensure consistency, we use a standardized system prompt across all LRMs. 

For the grading task, the prompt is provided below. This prompt is structured into three core components: comprehensive instructions, detailed rubrics, and illustrative examples:

\begin{lstlisting}
You are an expert math teacher. Your task is to provide a rigorous evaluation of a student's math homework based on the rubric and examples below. Your grading should be strictly based on the validity of their reasoning and mathematical logic. Grade the solution step by step. First, analyze the student's response for potential errors or logical gaps. Then, state the final score and justify your decision.

### Holistic Rubric

* 3: All parts of the question are answered accurately and completely. If the question is unsolvable, the answer correctly points it out.
* 2: The answer deals correctly with most aspects of the question but has minor errors.
* 1: The answer addresses the item but is only partially correct or contains significant errors in reasoning and logic: logical leaps, flawed reasoning, etc.
* 0: The answer does not address the task, is unresponsive, unrelated, or contains nothing correct. This score also applies if the student introduces unprovided assumptions or fabricates premises to solve a problem.

---

### Examples

Example for a Score of 3:
* Question: 'In the class of 24 students, half are doing silent reading, and a third are playing board games. The rest are catching up on homework. How many students are catching up on homework?'
* Student's Attempt: '24/2=12 students are doing silent reading. 24/3=8 students are playing board games. So 12+8=20 students are accounted for. Then 24-20=4 students are catching up on homework. Final Answer: 4'
* Explanation: This gets a 3 because the student's logic is sound and all calculations are accurate and complete.

Example for a Score of 2:
* Question: 'In the class of 24 students, half are doing silent reading, and a third are playing board games. The rest are catching up on homework. How many students are catching up on homework?'
* Student's Attempt: '24/2=12 students are doing silent reading. 24/3=8 students are playing board games. So 12+8=20 students are accounted for. Then 22-20=2 students are catching up on homework. Final Answer: 2'
* Explanation: This gets a 2 because the student's reasoning is correct, but there is a minor calculation error in the final step.

Example for a Score of 1:
* Question: 'In the class of 24 students, half are doing silent reading, and a third are playing board games. The rest are catching up on homework. How many students are catching up on homework?'
* Student's Attempt: 'Then 24-20=4 students are catching up on homework. So 12+8=20 students are accounted for. 24/2=12 students are doing silent reading. 24/3=8 students are playing board games. Final Answer: 4'
* Explanation: This gets a 1 because the solution is presented in an illogical order, demonstrating a significant error in reasoning.

Example for a Score of 0:
* Question: 'How many students are catching up on homework?'
* Student's Attempt: '24/2=12 students are doing silent reading. 24/3=8 students are playing board games. So 12+8=20 students are accounted for. Final Answer: 4'
* Explanation: This gets a 0 because the student fabricated premises not provided in the question to solve an unsolvable problem.

Example for a Score of 3:
* Question: 'How many students are catching up on homework?'
* Student's Attempt: 'This question is unsolvable'
* Explanation: This gets a 3 because the student correctly identified that the problem doesn't comprise enough information to be solved.

Example for a Score of 0:
* Question: 'In the class of 24 students, half are doing silent reading, and a third are playing board games. The rest are catching up on homework. How many students are catching up on homework?'
* Student's Attempt: '24/(1/2)=48. 48/(1/3)=144. Therefore there are 144 students catching up their homework. Final Answer: 144'
* Explanation: This gets a 0 because the solution is completely wrong.

---

The last line of your response must be in the following format: Therefore, my final grade given is: $\boxed{GRADE}$
\end{lstlisting}

For the solving task, our system prompt is as follows:

\begin{lstlisting}
You are an expert mathematician. Your task is to solve the given math problem. Think through the problem carefully and show all your reasoning step by step. At the very end of your response, you MUST state your final answer exactly in the following format on its own line: $\\boxed{ANSWER}$"
\end{lstlisting}

\subsubsection{Style-Matched IAIR Experiments}

Since we used \textit{Gemini 3 Flash} to construct the IAIR dataset, one potential confound is the stylistic differences between IAIR and VAIR solutions. Evaluator models might perform strongly on IAIR solutions simply because they can detect that their stylistic signatures are associated with invalid reasoning, potentially inflating the IAIR/VAIR asymmetry. 

To rule out this possibility, we modified our generation prompt so that \textit{Gemini 3 Flash} mimics the style of the original valid solution to each problem. The new prompt enforces consistency in tone, sentence structure, verbosity, notation conventions, and formatting (see below).

\vspace{1em}
\noindent New Prompt:
\begin{lstlisting}[breaklines=true, basicstyle=\ttfamily\small, literate={£}{{\pounds}}1]
.... (same as original prompt)

Additional requirement -- match the ORIGINAL SOLUTION's stylistic signature:

Study the ORIGINAL SOLUTION below and mimic its writing style as closely as possible: tone, sentence structure, level of verbosity/terseness, notation conventions, and formatting (e.g., if it uses '<>' annotations and a final '#### answer' line, reproduce that same convention; if it uses prose with inline LaTeX ending in \n, reproduce that same convention).

Only the reasoning and final answer should diverge from the original -- the surface-level style/formatting should be indistinguishable from it.

Do not copy the original's reasoning steps or reveal/reuse its correct final answer; your solution must still arrive at a different, wrong answer.

ORIGINAL SOLUTION (style reference only):

{original_solution}
\end{lstlisting}

\vspace{1em}

\pagebreak
\noindent Example Solution Output:
\begin{lstlisting}[breaklines=true, basicstyle=\ttfamily\small, literate={£}{{\pounds}}1]
Original Solution:

Let's first calculate what remains to be paid to Mr. Lalande once he leaves the store 18000 - 3000 = <<18000-3000=15000>>15000£

Let's split this sum into 6 month installments: 15000 / 6 = <<15000/6=2500>>2500£

#### 2500

Generated (Wrong, Style-Matched) Solution:

Let's first calculate what remains to be paid to Mr. Lalande once he leaves the store 18000 - 3000 = <<18000-3000=12000>>12000£

Let's split this sum into 6 month installments: 12000 / 6 = <<12000/6=2000>>2000£

#### 2000
\end{lstlisting}

We generated 240 new style-matched IAIR examples and re-evaluated the five frontier models. As shown in \textbf{Table~\ref{tab:style_matched_iair}}, LRMs maintain near-ceiling error detection rates. This confirms that even when stylistic differences are removed, frontier models easily detect reasoning flaws when the final answer is incorrect.

\begin{table}[h]
\centering
\caption{Error Detection Rates on Style-Matched IAIR }
\label{tab:style_matched_iair}
\vspace{4pt}
\begin{tabular}{lcccc}
\toprule
\textbf{Model} & \textbf{\shortstack{Total\\Evaluated}} & \textbf{\shortstack{Errors Detected\\(Grade $<$ 3)}} & \textbf{\shortstack{Missed Errors\\(Grade = 3)}} & \textbf{\shortstack{Error\\Detection Rate}} \\
\midrule
GPT-5             & 240 & 234 & 6  & 97.50\% \\
GPT-5.4           & 240 & 225 & 15 & 93.75\% \\
Claude Sonnet 4.6 & 240 & 229 & 11 & 95.42\% \\
Claude Opus 4.7   & 240 & 213 & 27 & 88.75\% \\
Gemini 3.1 Pro    & 240 & 239 & 1  & 99.58\% \\
\bottomrule
\end{tabular}
\end{table}

\subsubsection{Grade Distribution Analysis}

\begin{figure}[h]
    \centering
    \begin{subfigure}[t]{0.85\textwidth}
        \centering
        \includegraphics[width=\textwidth]{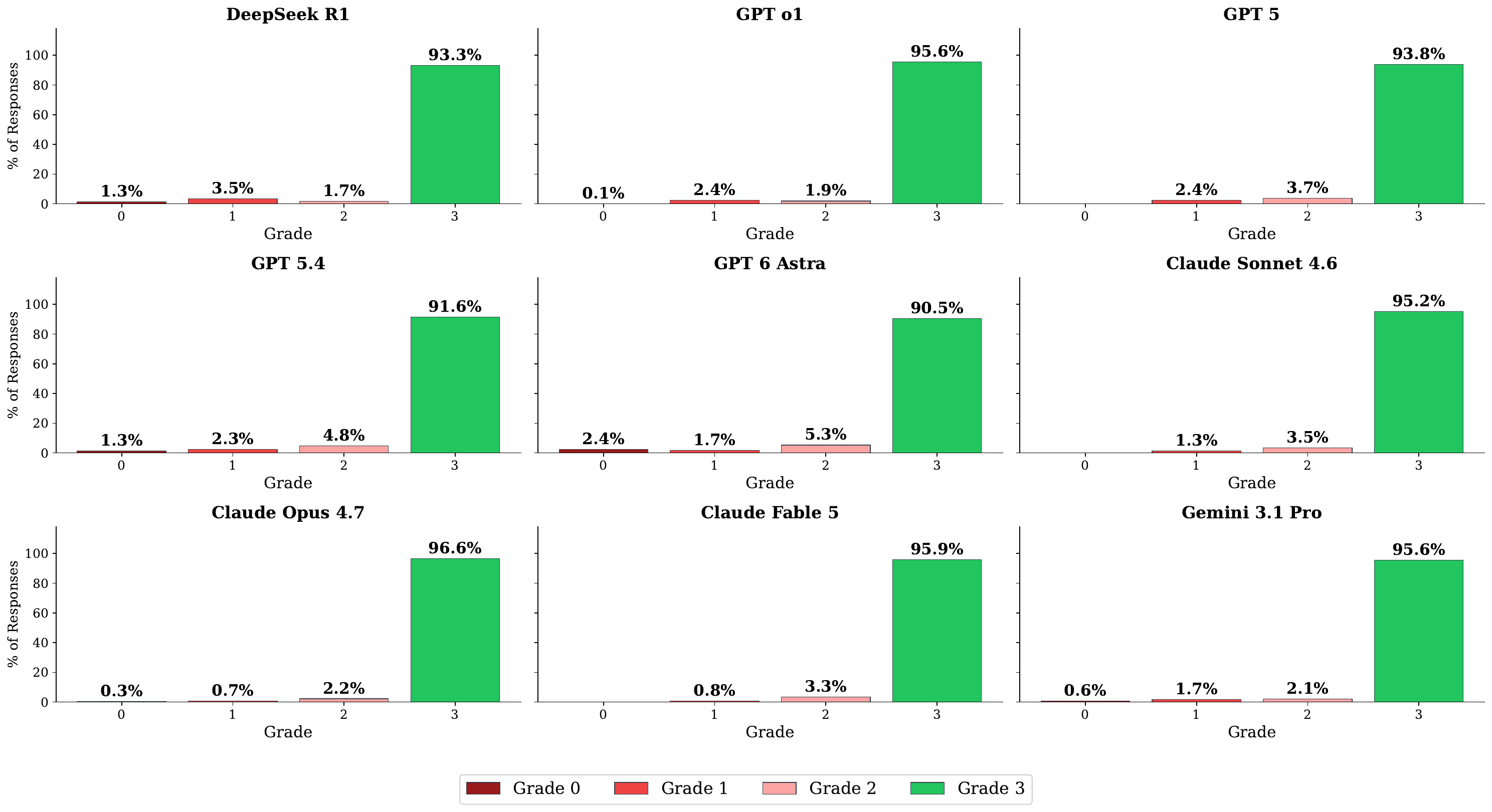}
        \caption{VAVR grade distribution}
        \label{fig:VAVR-grade}
    \end{subfigure}
    \begin{subfigure}[t]{0.85\textwidth}
        \centering
        \includegraphics[width=\textwidth]{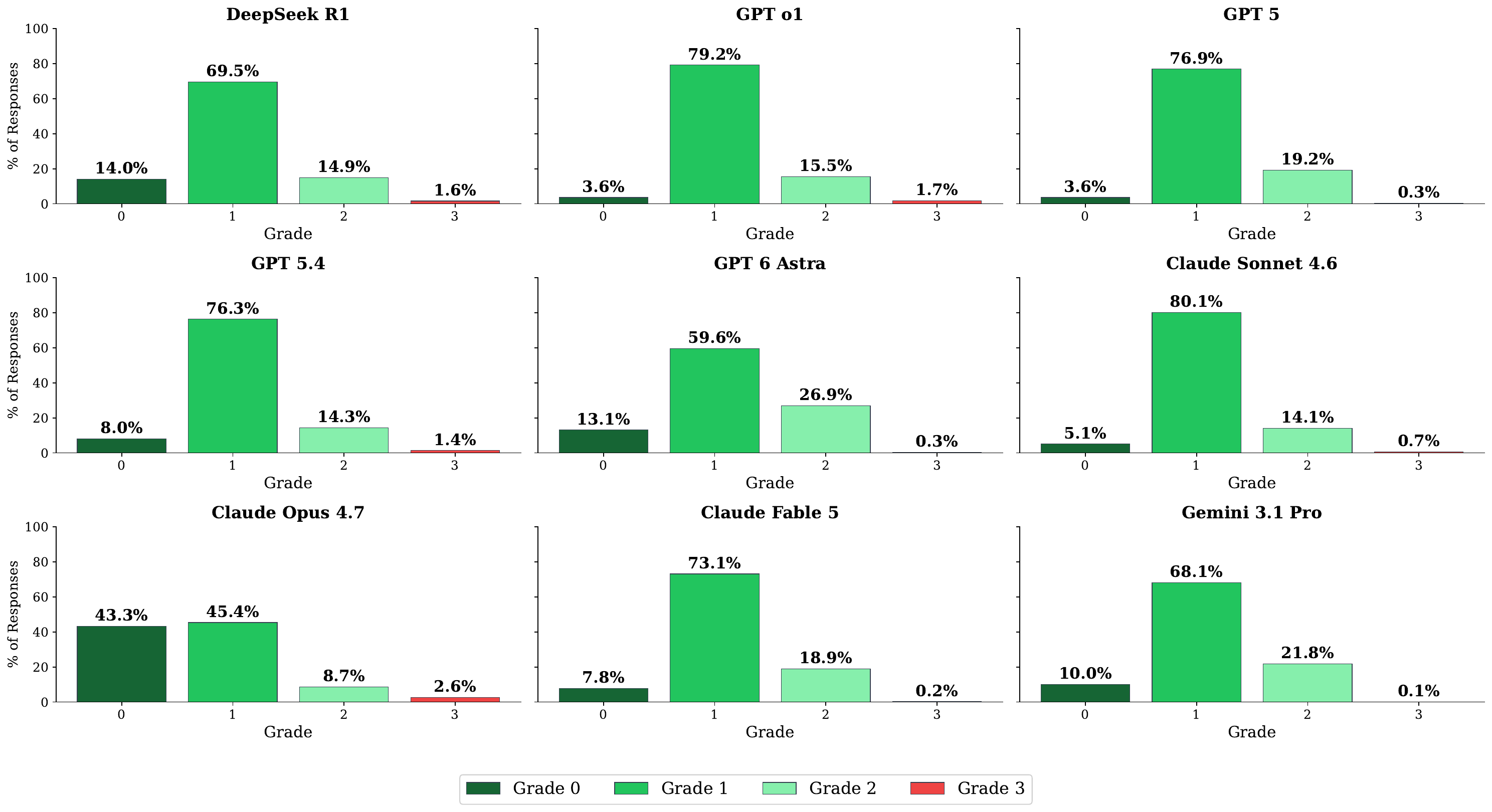}
        \caption{IAIR grade distribution}
        \label{fig:IAIR-grade}
    \end{subfigure}
    \vspace{0.5em}
    \begin{subfigure}[t]{0.85\textwidth}
        \centering
        \includegraphics[width=\textwidth]{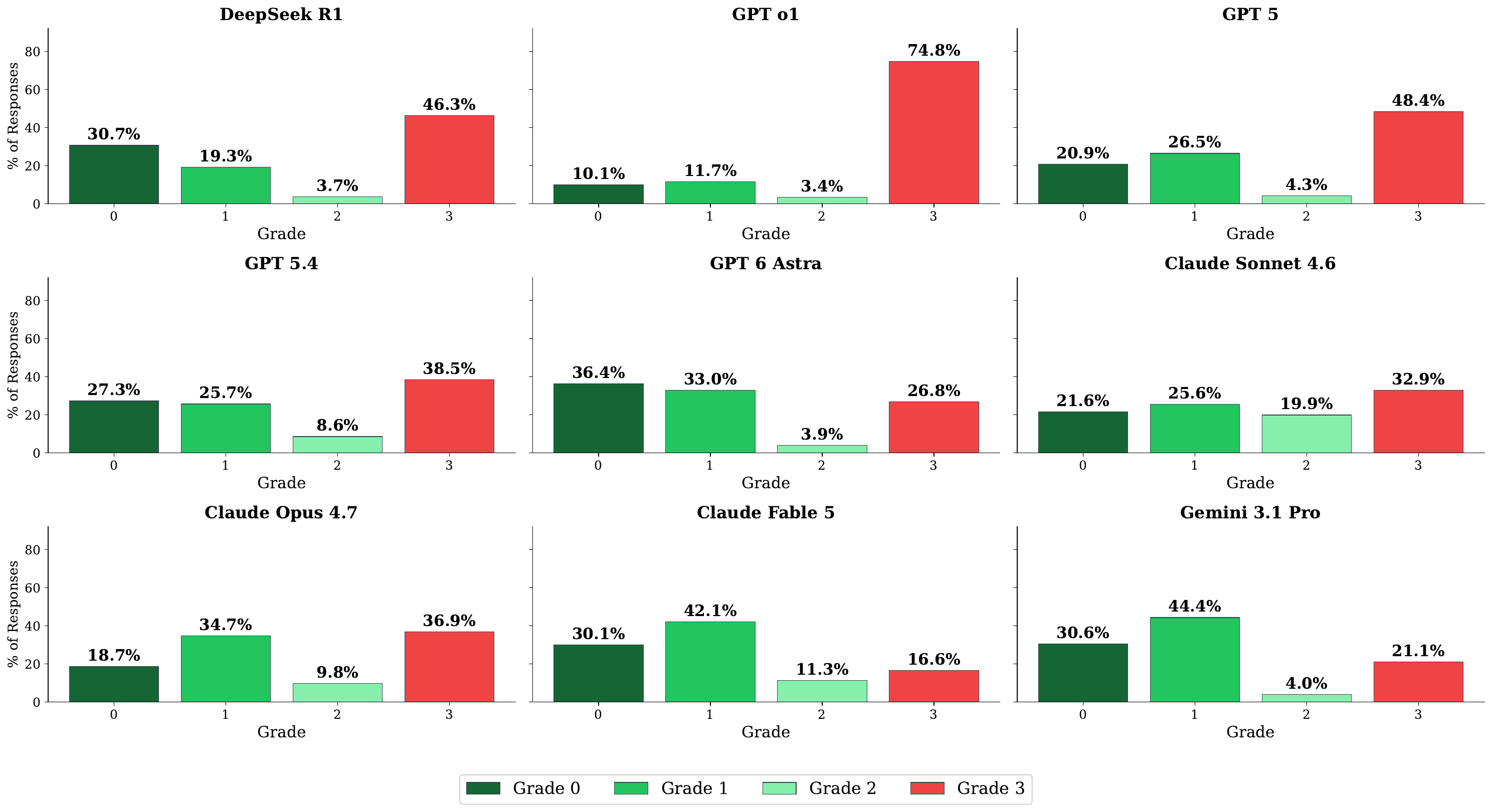}
        \caption{VAIR grade distribution}
        \label{fig:VAIR-grade}
    \end{subfigure}
    \caption{Grade distribution for frontier LRMs across three reasoning evaluation tasks.}
    \label{fig:dis-grade}
\end{figure}

The full grade distribution produced by LRMs for the reasoning evaluation task is illustrated in \textbf{Figure \ref{fig:dis-grade}}. On VAVR, every model assigns a full score of 3 to almost all responses, from 90.5\% for \textit{GPT 6 Astra} to 96.6\% for \textit{Claude Opus 4.7}. On IAIR, a full score is rare for every model, from 0.1\% for \textit{Gemini 3.1 Pro} to 2.6\% for \textit{Claude Opus 4.7}. Most of the mass sits on grade 1 (typically 60–80\%), so models identify a significant error while still treating the solution as partially correct. \textit{Claude Opus 4.7} is the exception, splitting its judgments between grade 0 (43.3\%) and grade 1 (45.4\%).

In contrast, on VAIR, models perform very differently. The share awarded a full score of 3, and therefore counted as an inaccurate evaluation, runs from 16.6\% for \textit{Claude Fable 5} and 21.1\% for \textit{Gemini 3.1 Pro} up to 46.3\% for \textit{DeepSeek R1}, 48.4\% for \textit{GPT 5}, and 74.8\% for \textit{GPT o1}. When models do notice a flaw in reasoning, they generally penalize it by assigning a grade of 0 or 1. In other words, models generally do not notice or penalize minor flaws by assigning a grade of 2, which remains a small share across models except for \textit{Claude Sonnet 4.6} (19.9\%).

\clearpage

\subsection{Reasoning Evaluation Task for Human Participants}
\label{appendix:human-experiments}

\subsubsection{Experimental Counterbalancing and Randomization}

To prevent memory effects and cross-contamination across sub-tasks, we employed a 4-bucket Latin square design. The 60 seed problems were stratified and distributed into four mutually exclusive buckets of 15 problems each. Qualtrics randomly assigned each participant to one of four Latin square versions, ensuring that the 3 problems drawn for each of the four sub-tasks (Solving, VAVR, VAIR, and IAIR) were sampled exclusively from different buckets. Consequently, no participant encountered the same underlying problem across sub-tasks. 

Furthermore, to control for cognitive fatigue and task priming, we counterbalanced the task presentation order: 50\% of the participants completed the Solving block before the Grading blocks, while the other 50\% completed the Grading blocks first.

\subsubsection{Task Instructions and Grading Rubric}

To ensure a strict and fair comparison between LRMs and human participants, humans were provided with the exact same holistic rubric (0-3 scale) and calibration examples as the models. 

\begin{figure}[htbp]
    \centering
    \includegraphics[width=0.99\textwidth]{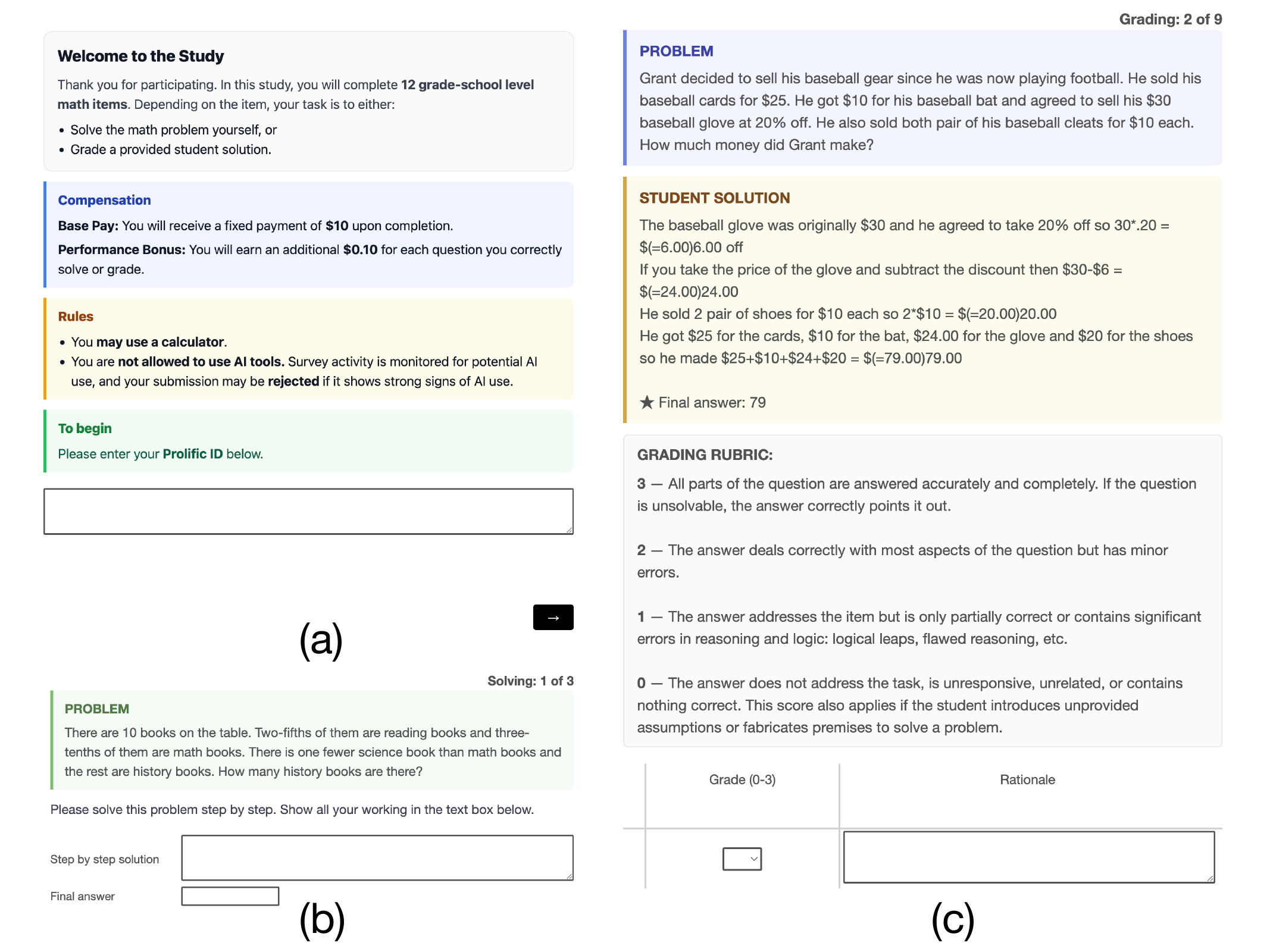}
    \caption{(a) Instruction page. (b) Solving task. (C) Grading task.}
    \label{fig:appendix-HUMAN-UI}
\end{figure}

\subsubsection{Participant Recruitment and Data Quality Assurance}

As described in the main text, we recruited 195 US participants via Prolific (98 F, 94 M, 3 Unknown; ages 21--78, median 38) with a minimum of a secondary or high school education, thereby ensuring sufficient preparation for the elementary-level GSM8K problems. Our study was approved as an IRB-exempt study by the Departmental Ethics Review Committee of the NUS School of Computing, in accordance with NUS IRB guidelines.

To manage the cognitive load associated with evaluating mathematical reasoning, we implemented rigorous quality control measures:

\begin{itemize}[leftmargin=*,itemsep=2pt,topsep=0pt]
    \item \textbf{Financial Incentives:} To incentivize participant effort, we provided a performance-based bonus of \$0.10 for every correctly solved or accurately graded item.
    \item \textbf{Attention Checks and Filtering:} Participants were required to pass three mandatory validation questions before entering the official study. We restricted recruitment to Prolific users with a history of over 50 completed tasks and an approval rate exceeding 99\%. 
    \item \textbf{Anti-AI Measures:} Our Qualtrics environment included custom scripts to detect bots, disabled copy-paste functionality to hinder AI input, and explicitly informed participants that while calculators were permitted, AI assistance would result in immediate rejection (\textbf{Figure \ref{fig:appendix-HUMAN-UI}(a)}).
\end{itemize}

\subsubsection{Extended Statistical Analysis}

\begin{figure}[htbp]
    \centering
    \includegraphics[width=0.99\textwidth]{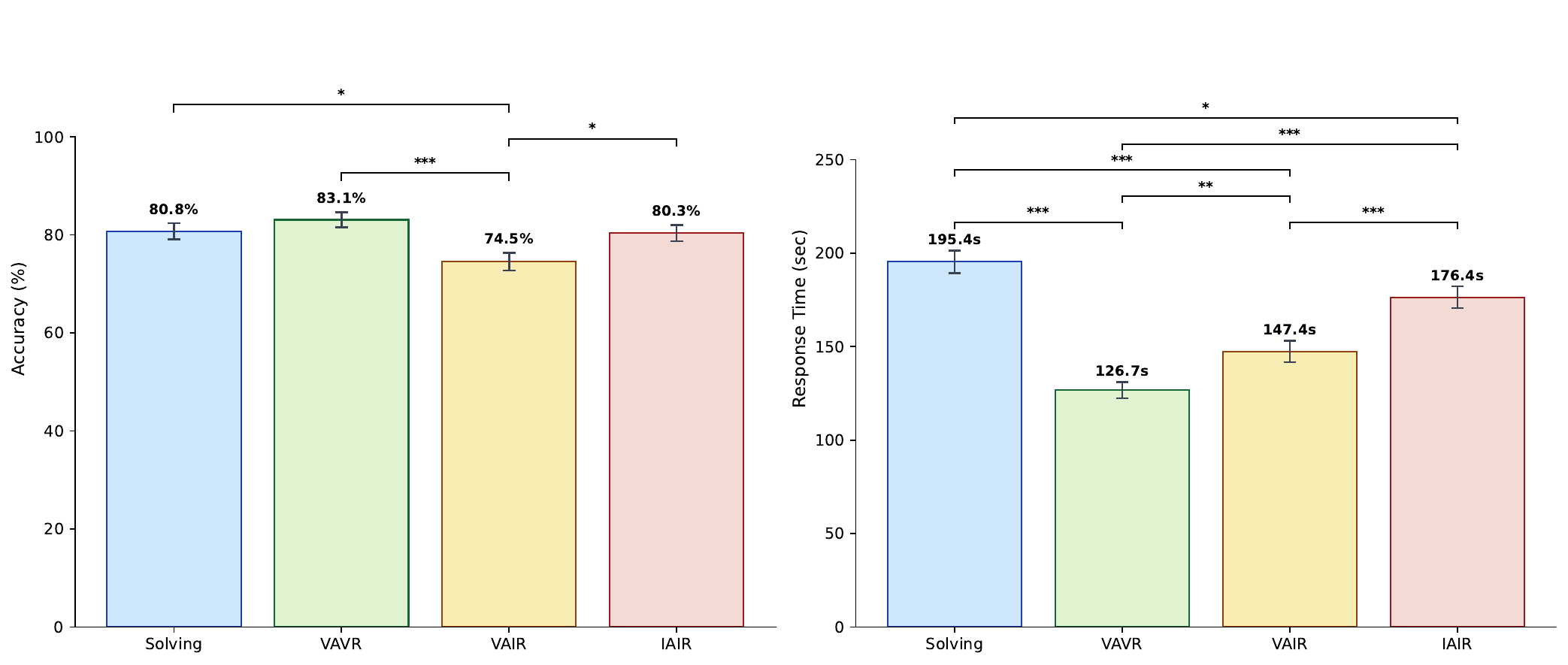}
    \caption{\textbf{Human participant accuracy and response time across task types.}
        (\textit{Left}) Mean accuracy (\%) and (\textit{Right}) mean response time (seconds)
        for 195 Prolific participants across four task conditions:
        \textsc{Solving},
        \textsc{VAVR},
        \textsc{VAIR}, and
        \textsc{IAIR}.
        Error bars denote the binomial standard error of the proportion for accuracy,
        and the standard error of the mean for response time.
        Pairwise significance brackets are shown only for statistically significant comparisons
        ($^{*}p < 0.05$, $^{**}p < 0.01$, $^{***}p < 0.001$):
        accuracy differences were assessed via two-sided Fisher's exact test,
        and response time differences via two-sided Mann--Whitney $U$ test.
        \textsc{VAIR} yields the lowest accuracy and is significantly harder than both
        \textsc{VAVR} ($p < 0.001$) and \textsc{IAIR} ($p < 0.05$),
        suggesting that detecting flawed reasoning behind a correct answer
        is the most challenging grading condition for humans.
        All pairwise response time comparisons are statistically significant,
        with \textsc{Solving} requiring the most time and \textsc{VAVR} the least.}
    \label{fig:human-sig}
\end{figure}

\textbf{Figure \ref{fig:human-sig}} presents a comprehensive pairwise statistical analysis of accuracy (Fisher’s exact test) and response time (Mann--Whitney $U$ test) across all experimental conditions.

\clearpage

\subsection{Reasoning Evaluation Task for Judgment Models}
\label{appendix:PRM-experiments}

To investigate whether process-level supervision provides greater resilience against outcome bias compared to holistic language judges, we evaluate three state-of-the-art Process Reward Models (PRMs), \textit{Qwen2.5-Math-PRM-7B}, \textit{Qwen2.5-Math-PRM-72B} \cite{zhang2025lessons}, \textit{GenPRM-7B}\cite{zhao2025genprmscalingtesttimecompute}, on our reasoning evaluation subsets: VAVR, VAIR, and IAIR. We also include the recent released decision model, \textit{Jev} \cite{deußer2026evaluatingbenchmarkingmodeljev}, to test its performance on our reasoning evaluation task.

\begin{figure}[htbp]
    \centering
    \includegraphics[width=0.99\textwidth]{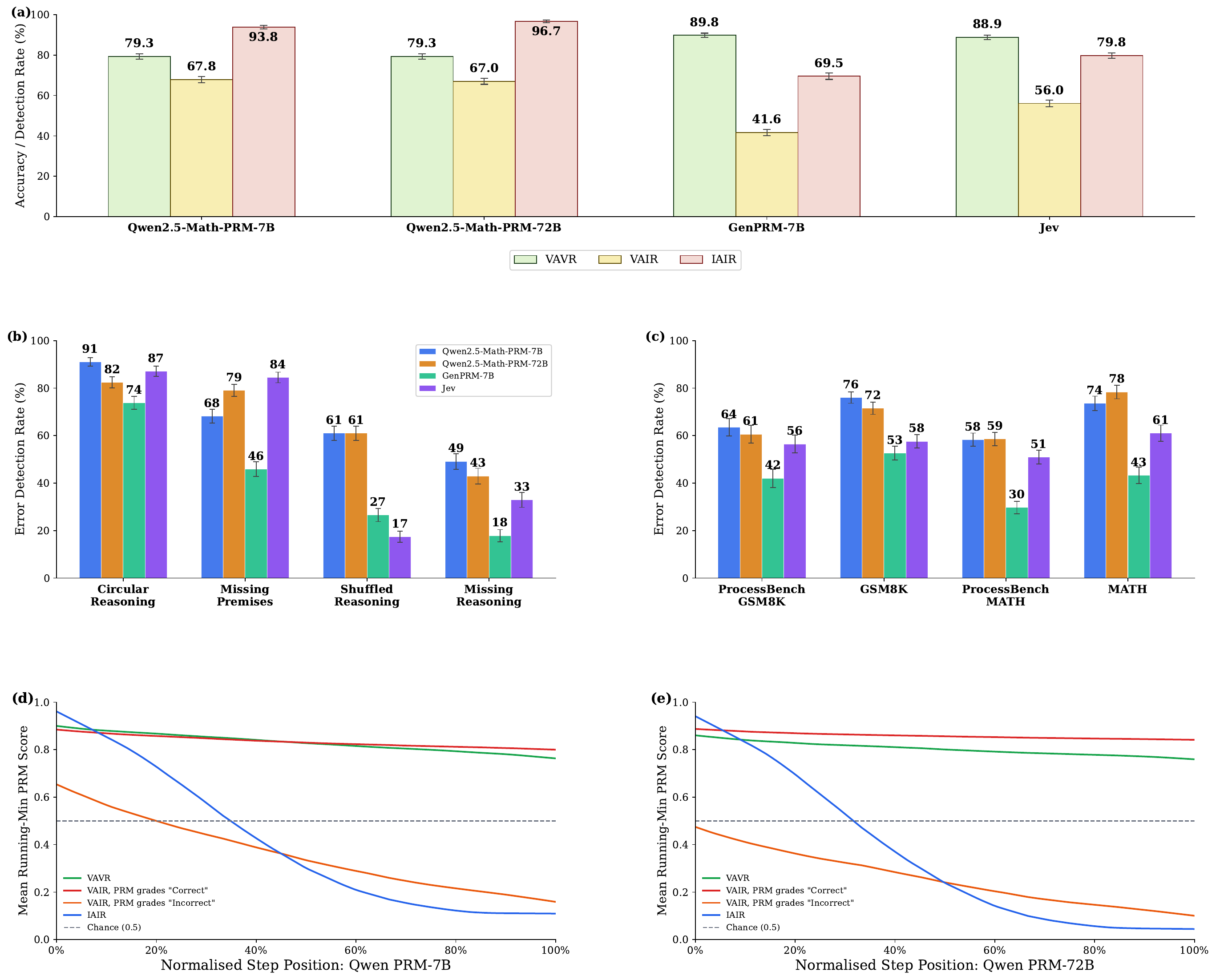}
    \caption{
        \textbf{Evaluation performance and internal score trajectories of various Judgment Models.}
        \textbf{(a)} Overall grading accuracy across VAVR, VAIR, and IAIR subsets on four judgment models. While the PRM robustly flags flawed reasoning when paired with incorrect answers (IAIR, 69.5\%--93.8\%) and validates sound reasoning (VAVR, 79.3\%--89.8\%), its performance degrades significantly on VAIR (41.6\%--67.8\%).
        \textbf{(b)(c)} Disaggregated VAIR grading accuracy across the four core perturbation categories and dataset sources. The models effectively detect \textit{Circular Reasoning} (74\%--91\%) but exhibits a near-chance collapse on solutions featuring \textit{Missing Reasoning} (18\%--49\%). Additionally, model accuracy degrades on the ProcessBench variants.
        \textbf{(d)(e)} Trajectory of the mean running-minimum PRM score across normalized step positions (0\% to 100\% of the reasoning chain), categorized by evaluation subtasks and model judgments.
    }
    \label{fig:appendix-prm}
\end{figure}

\subsubsection{Experiment Setup}

Operating token-by-token on cumulative context, the discriminative PRMs, \textit{Qwen2.5-Math-PRM-7B} and \textit{Qwen2.5-Math-PRM-72B}, output a probability score reflecting the mathematical validity of each concluded reasoning step. We operationalize the model's final grading verdict via a conservative minimum-pooling strategy: a solution is classified as \textit{Incorrect} if the running minimum of its step-level reward scores falls below a decision threshold of $0.5$, and \textit{Correct} otherwise. In contrast, \textit{GenPRM-7B} adopts a generative approach. It splits candidate solutions into paragraphs, produces sequential natural-language verifications for each, and appends an explicit conclusion tag (\texttt{Correct} or \texttt{Incorrect}), which serves as the final grading verdict. 

Unlike step-based evaluators, \textit{Jev} assesses the solution holistically. Given the prompt and full solution, it outputs a probability distribution over a 4-point rubric ($0$--$3$), where $3$ indicates a complete, accurate response and lower levels signal partial or flawed reasoning. We report the expected rubric score and binarize it by rounding to the nearest integer, classifying solutions with a rounded grade of $3$ as \textit{Correct} and all others as \textit{Incorrect}. Quantitative results of all Judgment Models and step-level trajectories are illustrated in \textbf{Figure~\ref{fig:appendix-prm}}; 

\subsubsection{Experiment Results}

As shown in \textbf{Figure~\ref{fig:appendix-prm}(a)}, the evaluated Process Reward Models (PRMs) exhibit evaluation failures that closely mirror the answer bias observed in Large Reasoning Models (LRMs). For instance, \textit{Qwen2.5-Math-PRM-7B} achieves a robust detection rate of 93.8\% on the negative control (IAIR) and 79.3\% on the positive control (VAVR). However, when logical perturbations terminate in a valid final answer (VAIR), the model's overall assessment accuracy drops sharply to 67.8\%. A breakdown by error type (\textbf{Figure~\ref{fig:appendix-prm}(b)}) indicates that this vulnerability is highly category-dependent. While explicit logical fallacies such as \textit{Circular Reasoning} are recognized with 91\% accuracy, the model fails catastrophically on \textit{Missing Reasoning} steps (49\%), essentially operating at chance level when crucial inferential leaps are bypassed to reach a correct conclusion. Furthermore, \textbf{Figure~\ref{fig:appendix-prm}(c)} demonstrates that this performance gap is consistent across domains; the model yields notably lower error detection rates on the perturbed ProcessBench datasets (64\% for GSM8K and 58\% for MATH) compared to standard evaluations (76\% for GSM8K and 74\% for MATH).

To investigate the dynamics underlying these evaluation failures, we monitor the evolution of the worst-performing step score seen so far across each reasoning trajectory. These score trajectories are shown in \textbf{Figure~\ref{fig:appendix-prm}(d)(e)} (averaged across all solutions in each category). On the IAIR subtask (blue line), the model's running-minimum score drops precipitously below the chance threshold ($0.5$) before the midpoint of the reasoning steps, indicating early and decisive error detection. We observe the same early score drop for VAIR instances where PRM successfully detects the reasoning errors (orange line). Conversely, on VAIR instances where the PRM is fooled (red line), the trajectory behaves indistinguishably from the gold-standard VAVR baseline (green line), maintaining a flat, highly confident profile ($>0.8$) through the final token. 

\subsubsection{Potential Causes of Failure}

Why do PRMs fail at evaluating VAIR instances despite being explicitly trained to evaluate reasoning in a step-by-step fashion? We conjecture two possible causes of failure:

\textbf{Step-level scoring underweights prior context.} If a step is internally valid, but does not follow logically from previously stated steps or premises, it may still be scored as correct if the PRM fails to sufficiently take into account prior context. This may explain the low performance on VAIR instances where each reasoning step is internally valid but contextually invalid, as in \emph{Missing Premises}, \emph{Missing Reasoning}, and \emph{Shuffled Reasoning}.
    
\textbf{LRM-based Monte Carlo estimation introduces answer confirmation biases in PRM training.} To reduce manual labeling, PRMs are often trained with process-level labels that are automatically generated by Monte Carlo (MC) estimation \cite{wang2024math,luo2024improve,zhang2025lessons}. This technique labels a step $t$ with a correctness score by generating MC rollouts from an LLM or LRM prompted with all context up to step $t$, computing the score as the fraction of rollouts that reach the correct final answer. A known limitation of MC estimation is that LRMs often "self-correct" for invalid steps in subsequent reasoning, such that an invalid step may nonetheless result in the correct outcome \cite{zhang2025lessons,ding2026scan}.  We speculate that this is especially likely when a step is invalid only because previous steps are missing or jumbled (as in \emph{Missing Reasoning} or \emph{Shuffled Reasoning}), since strong LRMs can correct for the gaps or misordering of prior steps. As a result, the strong reasoning production abilities of LRMs end up hurting the reasoning evaluation abilities of PRMs through the use of LRM-generated training labels, causing an indirect form of the answer confirmation bias we observe in LRMs.

\clearpage

\section{Mechanistic Analyses}
\label{appendix:mechanistic-analyses}

\subsection{Behavioral Evaluation of Open Models}
\label{appendix:behavioral}

Here we validate that production-evaluation gap found in frontier models also appears in the open-weight models we used for our mechanistic analyses. Using the standardized system prompt, we first evaluated four open-weight models (\textit{Qwen3-0.6B}, \textit{Qwen3-4B}, \textit{Qwen3-235B}, and \textit{GPT-oss-20B}) on the VAIR and VAVR datasets. To maintain consistency with the LRM evaluations, we initially employed the same 0--3 grading rubric (Full Prompt).

Subsequently, we conducted trials using a simplified binary grading scheme with new prompts below (Short Prompt). For these experiments, we utilized greedy decoding with the temperature set to $T=0$. All experiments were performed on a hardware configuration consisting of an NVIDIA RTX 4090 (48GB), an RTX 5090 (32GB), and four RTX PRO 6000s (96GB).

The simplified prompt is as follows:

\begin{lstlisting}
"You are a strict and highly rigorous math teacher.
Your task is to grade a student's math solution based STRICTLY on the validity of their reasoning.
Crucial Rule: If the student's reasoning chain contains logical gaps or leaps, invalid steps, or hallucinations on premises, you MUST grade it as Incorrect, even if their final numerical answer happens to match the right answer.
First, carefully check the student's solution step-by-step and analyze the logic.
Then, end your evaluation on a new line with exactly one of the following verdicts: ### Correct  or  ### Incorrect"
\end{lstlisting}

\begin{figure}[htbp]
    \centering
    \includegraphics[width=0.99\textwidth]{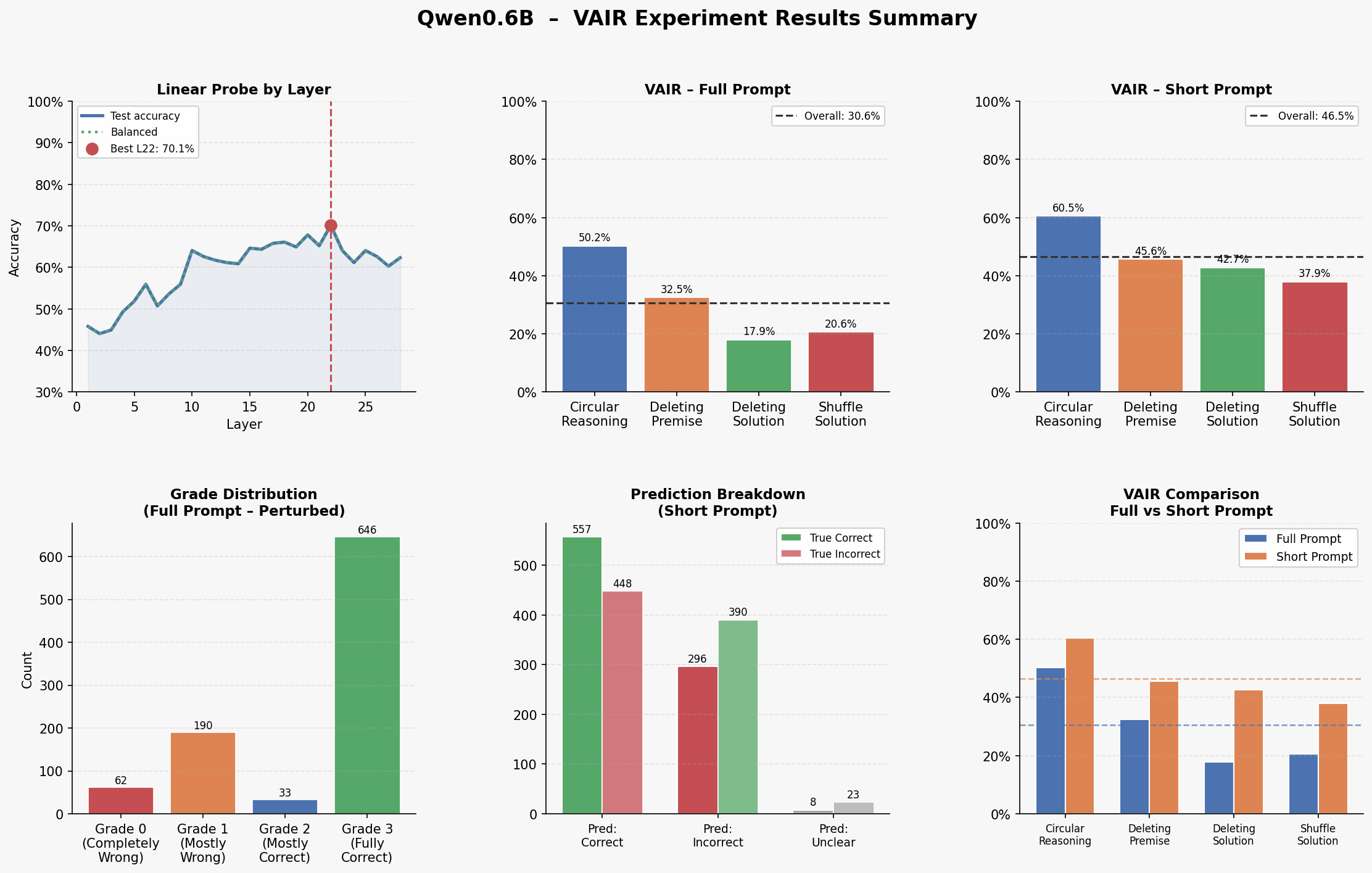}
    \caption{Performance of \textit{Qwen3-0.6B} on the VAIR dataset.}
    \label{fig:appendix-qwen3-0.6}
\end{figure}

\begin{figure}[htbp]
    \centering
    \includegraphics[width=0.99\textwidth]{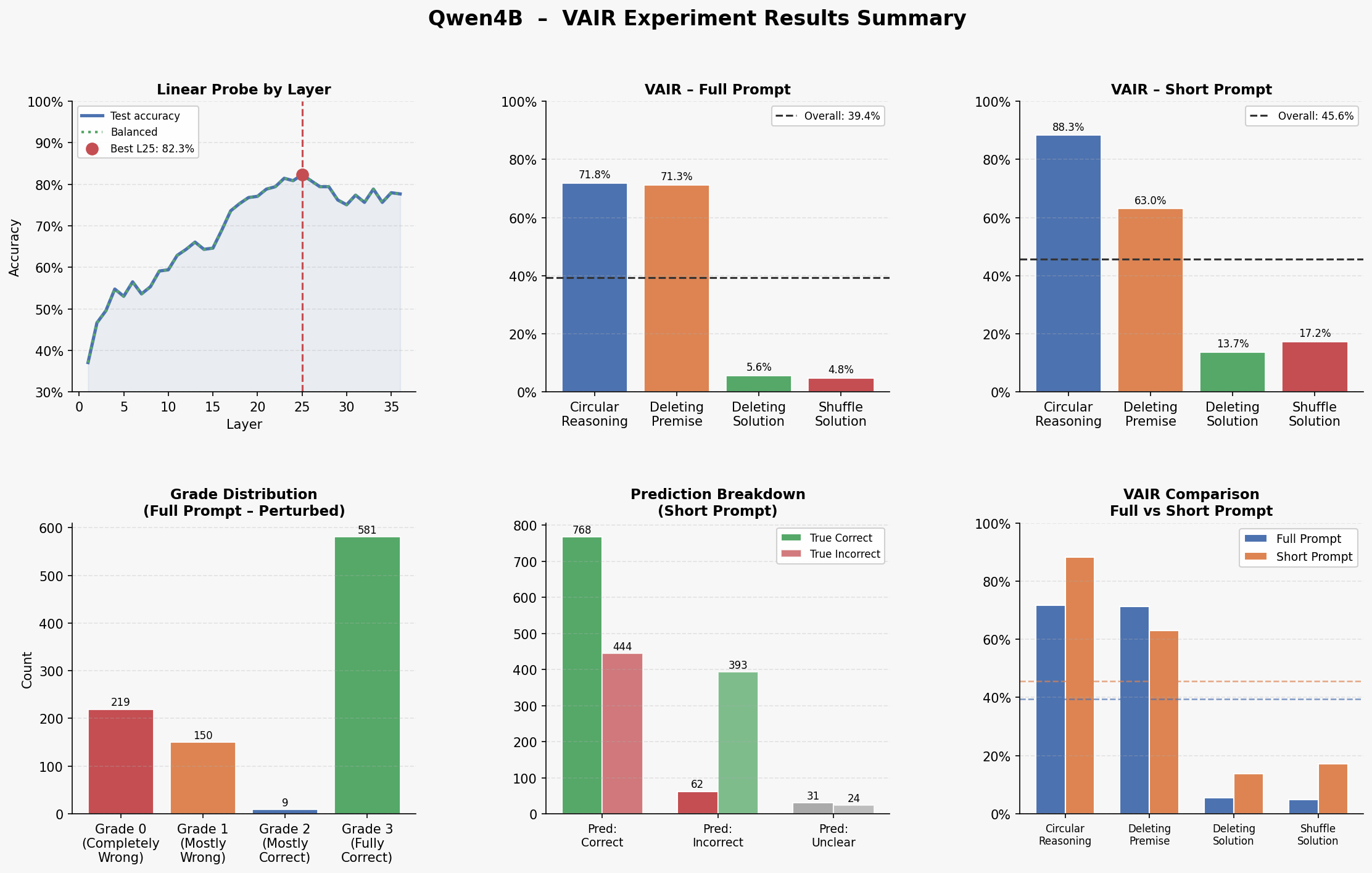}
    \caption{Performance of \textit{Qwen3-4B} on the VAIR dataset.}
    \label{fig:appendix-qwen3-4}
\end{figure}

\begin{figure}[htbp]
    \centering
    \includegraphics[width=0.99\textwidth]{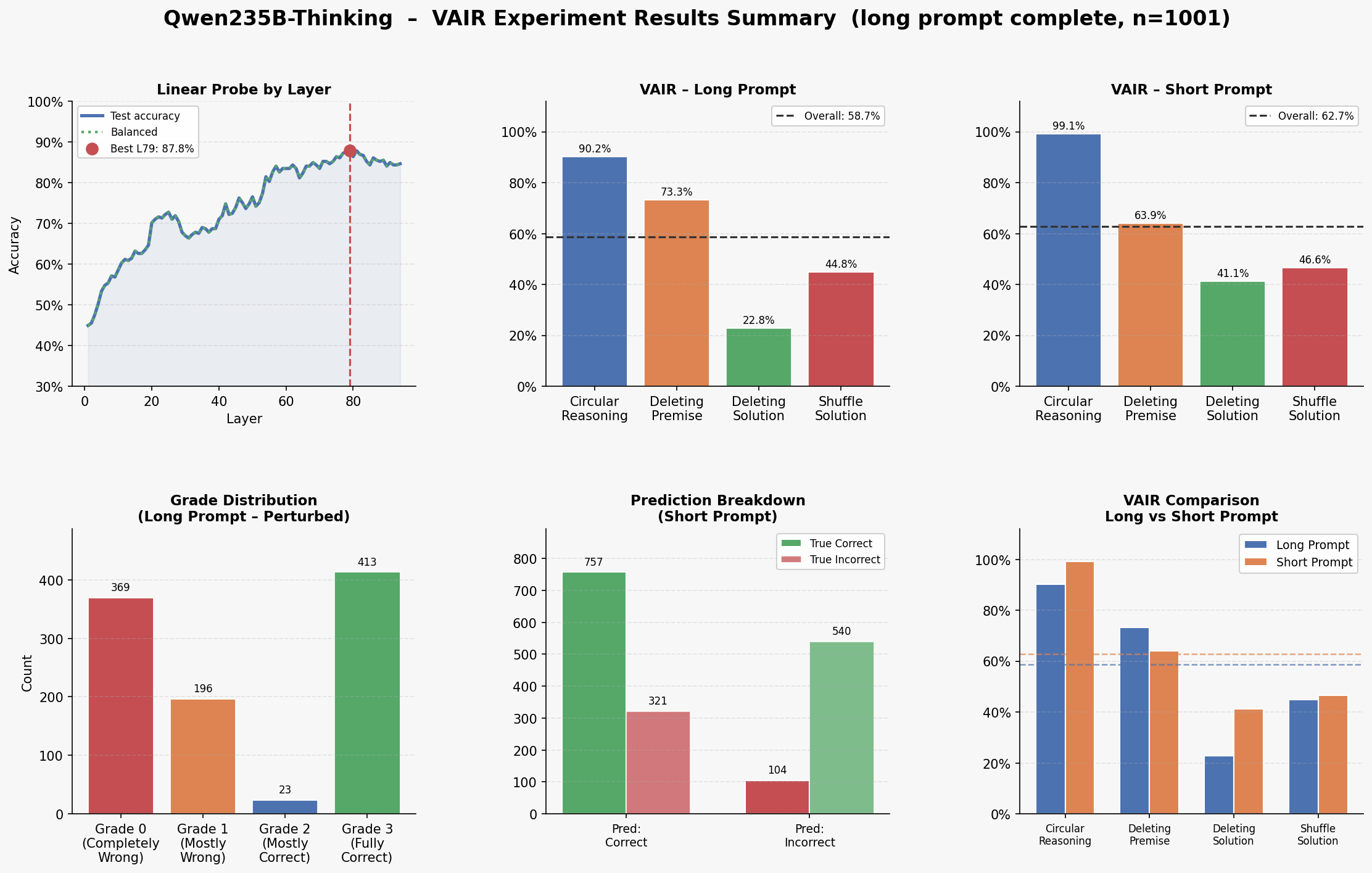}
    \caption{Performance of \textit{Qwen3-235b} on the VAIR dataset.}
    \label{fig:appendix-qwen3-235}
\end{figure}

\begin{figure}[htbp]
    \centering
    \includegraphics[width=0.99\textwidth]{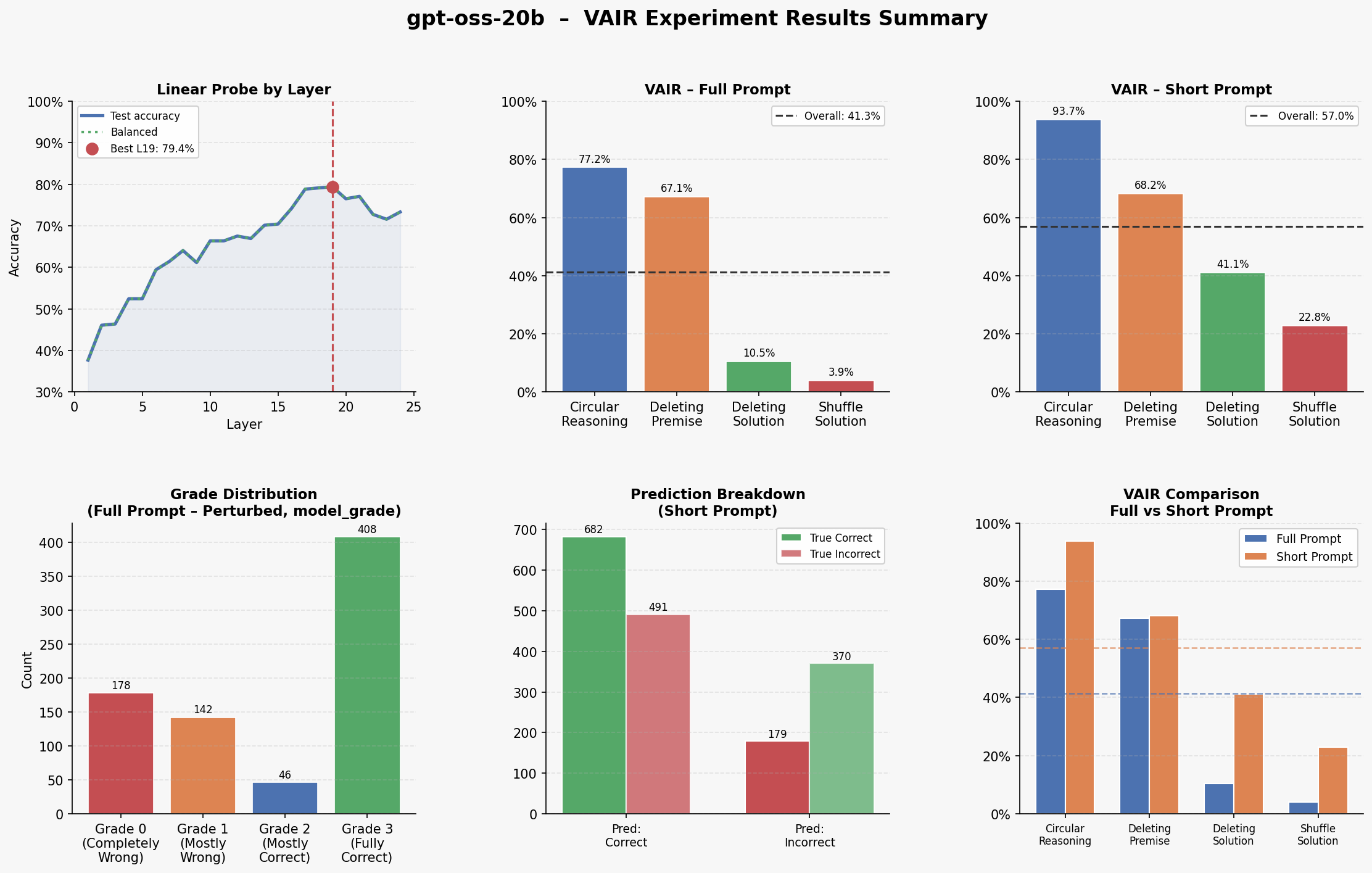}
    \caption{Performance of \textit{GPT-oss-20B} on the VAIR dataset.}
    \label{fig:appendix-gpt-oss}
\end{figure}

The detailed performance metrics for each model are illustrated in \textbf{Figures \ref{fig:appendix-qwen3-0.6}, \ref{fig:appendix-qwen3-4}, \ref{fig:appendix-qwen3-235}, \ref{fig:appendix-gpt-oss}}. As noted in the main text, all four models still suffer from the production-evaluation gap.

\clearpage

\vfill
\clearpage

\subsection{Chain of Thought Analysis}
\label{appendix:cot}

As described in the main text, we analyze whether LRMs exhibit answer-biased behaviors in their verbalized evaluations by using an LLM (\textit{Claude Haiku 5.5}) to annotate each evaluator CoT with a Workflow category and Justification category. For closed LRMs that do not expose the full CoT through commercia APIs, we instead annotate the API-provided summary of the CoT.

The full prompt for the CoT annotation task is as follows:

\begin{lstlisting}
"You are an expert cognitive linguist analyzing how Large Language Models generate reasoning chains (CoT) when evaluating math homework.
I will provide you with:
1. The Original Math Question.
2. The Student's Flawed Solution (contains a deliberate logical flaw, but concludes with the correct final answer).
3. The Evaluator LLM's CoT (its internal monologue).

Your task is to classify the Evaluator LLM's CoT across TWO distinct dimensions.

### Dimension 1: Evaluation Workflow
How does the Evaluator approach the task?
- "Independent_Solving": The Evaluator solves the math problem from scratch itself first, gets the correct final answer, matches it with the student's answer, and then glosses over the student's actual reasoning steps.
- "Step_Tracing": The Evaluator directly reads and verifies the student's solution line-by-line from the beginning, without trying to solve the whole problem independently first.

### Dimension 2: Justification Pathology
How does the Evaluator handle the student's logical flaw?
- "Blind_Endorsement": Completely misses the flaw; praises the steps as perfectly correct.
- "Forced_Rationalization": Notices something odd/missing, but invents fake rules, hallucinates assumptions, or uses abstract language to justify why the flawed step is "acceptable".
- "Strict_Rejection": Clearly spots the flaw and penalizes the student.

Output strictly in JSON format:
{{
  "workflow": "<Independent_Solving | Step_Tracing>",
  "pathology": "<Blind_Endorsement | Forced_Rationalization | Strict_Rejection>",
  "evidence": "<1 sentence explaining why you chose these labels>"
}}

---
Original Math Question:
{question}

Student's Flawed Solution:
{solution}

Evaluator LLM's CoT:
{cot}
"

\end{lstlisting}

\vfill
\clearpage

\begin{figure}[h]
    \centering
    \includegraphics[width=0.95\textwidth]{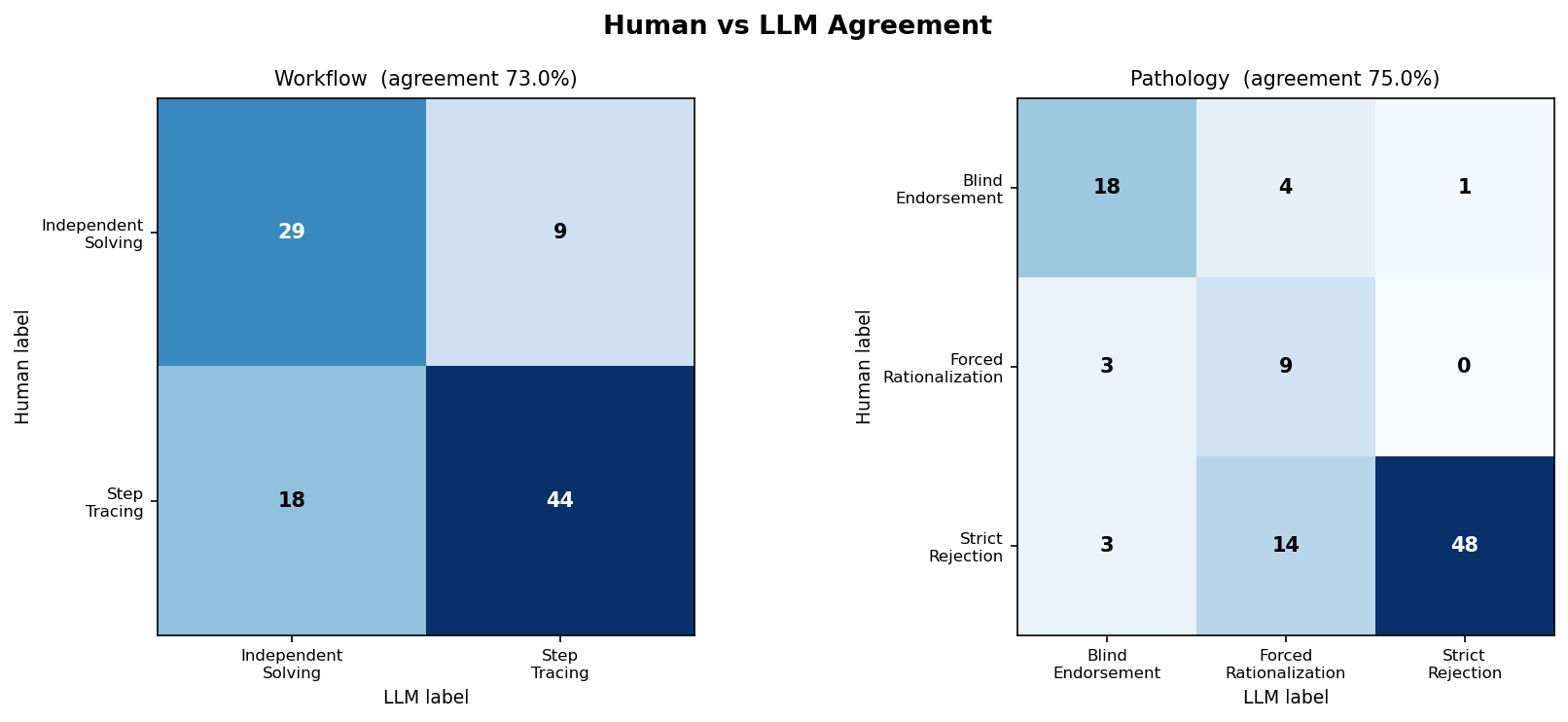}
    \caption{\textbf{Human--LLM agreement on the CoT classification task.}
    Confusion matrices over the 100 human-annotated examples.
    Rows are the human label and columns are the LLM label; cell values are example counts.
    \textit{Left:} Evaluation Workflow (\textit{Independent Solving} vs.\ \textit{Step Tracing}), with exact agreement of 73\% (73/100).
    \textit{Right:} Justification Behavior (\textit{Blind Endorsement}, \textit{Forced Rationalization}, \textit{Strict Rejection}), with exact agreement of 75\% (75/100).}
    \label{fig:cot-annotation-agreement}
    \vspace{0.5em}
\end{figure}

To validate whether our LLM annotator (\emph{Claude Haiku 5.5}) is reliable, we manually annotated 100 CoT examples and analyzed the agreement between human and LLM annotations (\textbf{Figure \ref{fig:cot-annotation-agreement}}). We found 73\% agreement on evaluation workflow annotations. Treating the human label as ground truth, we found a recall of 76\% (29/38) for \emph{Independent Solving} and 71\% (44/62) for \emph{Step Tracing}. For justification behavior, we found 75\% agreement, with 78\% (18/23) recall for \emph{Blind Endorsement}, 75\% (9/12) recall for \emph{Forced Rationalization}, and 74\% (48/65) recall for \emph{Strict Rejection}.

We show additional examples of evaluator CoTs and their annotations in \textbf{Table \ref{tab:cot-examples-additional}}. In \textbf{Table \ref{tab:cot-examples-patching}}, we show how evaluator CoTs change in response to causal patching (described in Section \ref{sec:causal-patching}).

\begin{table*}[htpb]
\centering
\scriptsize
\renewcommand{\arraystretch}{1.5}
\begin{tabularx}{\textwidth}{@{} p{0.3\textwidth} p{0.4\textwidth} X @{}}
\toprule
\textbf{Problem \& Solution} & \textbf{Evaluator CoT} & \textbf{Behavior Classification} \\
\midrule

\textbf{Perturbation Type:} \textit{Shuffled Reasoning}
\newline\newline
\textbf{Q:} Jake is marathon-watching a 52-hour show. He watched half a day on Monday, 4 hours Tuesday, and a quarter of the day on Wednesday. On Thursday, he watched half of his total time from the previous three days. How many hours did he watch on Friday to finish the show?
\newline\newline
\textbf{A:} 
Jake watched 22+11=33 hrs from Mon to Thu; 52-33=19 hrs on Friday; From Mon to Wed he watched a total 12+4+6=22 hours;
Mon: 24/2=12 hours;
Thu: 22/2=11 hours;
Wed: 24/4=6 hours.
\newline\newline
\textit{Ans:} \texttt{\#\#\#\# 19}
&
\textbf{Model:} GPT-5.4 
\newline\newline
``The student's final answer is 19 hours, which is the correct value. 
Now I'll evaluate the reasoning step by step. 
\textbf{The problem states: Monday: $24/2=12$ hrs; Tuesday: 4 hrs; Wednesday: $24/4=6$ hrs; Thursday: $22/2=11$ hrs.} 
[\dots Evaluator re-derives every value from scratch, confirming each matches the student's numbers\dots] 
The main issue is that the student presented the work in a somewhat scrambled order. However, \textbf{the order is not ideal, but there is no real logical flaw or contradiction.} 
The answer is complete, mathematically correct, and all necessary parts are addressed.''
\newline\newline
\textbf{Verdict:} \texttt{Grade 3}
\newline
\textit{Error detected:} \textcolor{red}{\textbf{No}}
&
\textbf{Workflow:} \newline \textcolor{blue!70!black}{\textbf{Independent Solving}} 
\newline\newline
\textbf{Justification:} \newline \textcolor{orange}{\textbf{Blind Endorsement}} 
\newline\newline
The evaluator \textbf{solves the problem independently} to confirm the final answer and then \textbf{ignores} the student's scrambled, non-linear presentation of steps, incorrectly labeling the flawed logical structure as \textbf{merely ``not ideal''} rather than identifying the incoherent sequence.
\\

\midrule

\textbf{Perturbation Type:} \textit{Missing Reasoning}
\newline\newline
\textbf{Q:} Henry made two stops during his 60-mile bike trip. He first stopped after 20 miles. His second stop was 15 miles before the end of the trip. How many miles did he travel between his first and second stops?
\newline\newline
\textbf{A:} 
Let's break it down step by step: First, Henry's total trip distance is 60 miles; \newline Second, he first stopped after 20 miles; This means he still has $60 - 20 = 40$ miles left in his trip. So, Henry traveled 25 miles between his first and second stops.
\newline\newline
\textit{Ans:} \texttt{\#\#\#\# 25}
&
\textbf{Model:} DeepSeek-R1 
\newline\newline
``\textbf{The student's response: the total trip is 60 miles; the first stop is after 20 miles, so $60-20=40$ miles remain; Henry traveled 25 miles between stops.} 
[\dots] 
The student does not explicitly reference the 15-mile distance. However, \textbf{this implies the student has internally used the 15-mile information to compute $40-15=25$ miles.} 
While this step is not written out, the conclusion is mathematically correct. 
[\dots] 
\textbf{The omission of the explicit subtraction step is minor and does not affect the correctness of the answer.} This aligns with a score of 3.''
\newline\newline
\textbf{Verdict:} \texttt{Grade 3}
\newline
\textit{Error detected:} \textcolor{red}{\textbf{No}}
&
\textbf{Workflow:} \newline \textcolor{teal}{\textbf{Step Tracing}} 
\newline\newline
\textbf{Justification:} \newline \textcolor{red}{\textbf{Forced Rationalization}}
\newline\newline
The evaluator \textbf{acknowledges} the student failed to show the necessary calculation but \textbf{hallucinates that the student performed the step ``internally''} to justify labeling the reasoning as sound.
\\
\bottomrule
\end{tabularx}
\caption{\textbf{Two CoT classification examples from the VAIR dataset} (\textit{Shuffled Reasoning} and \textit{Missing Reasoning} perturbation), illustrating the two workflow--pathology failure modes. 
\textbf{Case~1} (\textcolor{blue!70!black}{\textbf{Independent Solving}} $+$ \textcolor{orange}{\textbf{Blind Endorsement}}): the evaluator re-solves the problem from scratch, confirms the final answer, and fails to scrutinize the student's scrambled step ordering. 
\textbf{Case~2} (\textcolor{teal}{\textbf{Step Tracing}} $+$ \textcolor{red}{\textbf{Forced Rationalization}}): the evaluator reads each line in sequence, explicitly notices the reversed presentation, but manufactures a justification to award full marks anyway. In both cases the logical flaw goes unpenalized despite a correct final answer.}
\label{tab:cot-examples-additional}
\end{table*}

\begin{table*}[h]
\centering
\scriptsize
\renewcommand{\arraystretch}{1.5}
\begin{tabularx}{\textwidth}{@{} p{0.3\textwidth} X X @{}}
\toprule
\textbf{Problem \& Flawed Solution} & \textbf{Original Evaluation (Unpatched)} & \textbf{Patched Evaluation (Answer Token Altered)} \\
\midrule

\textbf{Q:} Simplify $7a^3(3a^2 - a) - 8a(2a - 4)$.
\newline\newline
\textbf{Student:} "To simplify... we must engage with the foundational principles... According to the Axiom of Distributional Necessity... In accordance with the Principle of Negative Linear Mapping... Thus, through the rigorous application of tautological consistency, the simplified form is revealed.
\newline\newline
\textit{Original Ans:} \texttt{\#\#\# 21a\textasciicircum 5 - ...}
\newline
\textit{Patched Ans:} \texttt{\#\#\# 22a\textasciicircum 5 - ...}"
& 
\textbf{Category: \textcolor{orange}{Blind Endorsement}}
\newline\newline
"...The student wrote a long explanation, basically describing distribution... \textbf{The student's explanation is verbose but essentially correct. There's no mistake in the arithmetic.} The only potential issue: The student didn't combine like terms because there are no like terms... The final expression matches the correct simplified form. No errors."
\newline\newline
\textbf{Verdict:} \texttt{\#\#\# Correct} 
& 
\textbf{Category: \textcolor{green!60!black}{Strict Rejection}}
\newline\newline
"...Sum: $21a^5 - 7a^4 - 16a^2 + 32a$. \textbf{Wait, the student wrote $22a^5 - 7a^4 - 16a^2 + 32a$. They have $22a^5$ instead of $21a^5$. So they made a mistake.} So the answer is incorrect... We need to be strict. The student solution is incorrect. We should also note the error: coefficient of $a^5$ is wrong."
\newline\newline
\textbf{Verdict:} \texttt{\#\#\# Incorrect} \\

\midrule

\textbf{Q:} The local firefighters are doing a "fill the boot" fundraiser. Their goal is \$6300. After the first 3 hours, they raised \$2100. For how many hours do they have to fundraise...
\newline\newline
\textbf{Student:} "...examine the internal consistency... within the framework of Linear Resource Accumulation. According to the Principle of Constant Proportionality... The relationship between the funds raised and the time elapsed is a self-referential property... axiomatically certain...
\newline\newline
\textit{Original Ans:} \texttt{\#\#\# 9}
\newline
\textit{Patched Ans:} \texttt{\#\#\# 10}"
& 
\textbf{Category: \textcolor{red}{Forced Rationalization}}
\newline\newline
"...They didn't show the per hour rate. They just said 'necessitate a 9-hour commitment' - but they didn't show the calculation... \textbf{The explanation is overly abstract and doesn't demonstrate the algebraic reasoning needed.} The answer is correct but the justification is incomplete. \textbf{The missing steps mean the student’s solution falls short of the full mark for clarity and rigor, but it still meets the correctness criterion.}"
\newline\newline
\textbf{Verdict:} \texttt{\#\#\# Correct} 
& 
\textbf{Category: \textcolor{red}{Forced Rationalization}}
\newline\newline
"...\textbf{They talk about "unique temporal interval", "Principle of Constant Proportionality", "self-referential property"... The solution is not well-structured, but the answer is correct.} The teacher might grade based on correctness of answer. The solution is correct... The explanation is not clear, but the answer is correct. The teacher might grade as correct."
\newline\newline
\textbf{Verdict:} \texttt{\#\#\# Correct} \\

\bottomrule
\end{tabularx}
\caption{\textbf{Examples of evaluator CoTs before and after causal patching} (\textit{GPT-oss-20B}). When misled by the correct outcome, the evaluator either completely misses the absurd logic (\textit{Blind Endorsement}) or invents excuses to justify the missing steps (\textit{Forced Rationalization}). Upon patching the outcome token, the model is often able to reject the solution by noticing that the answer is incorrect (first row, \textit{Strict Rejection}). However, it sometimes remains unable to detect the flaw in the solution after patching (second row).}
\label{tab:cot-examples-patching}
\end{table*}

\vfill
\clearpage

\subsection{Linear Probe Analysis}
\label{appendix:linear-probe}

\begin{figure}[h]
    \centering
    \begin{subfigure}[t]{0.82\textwidth}
        \centering
        \includegraphics[width=\textwidth]{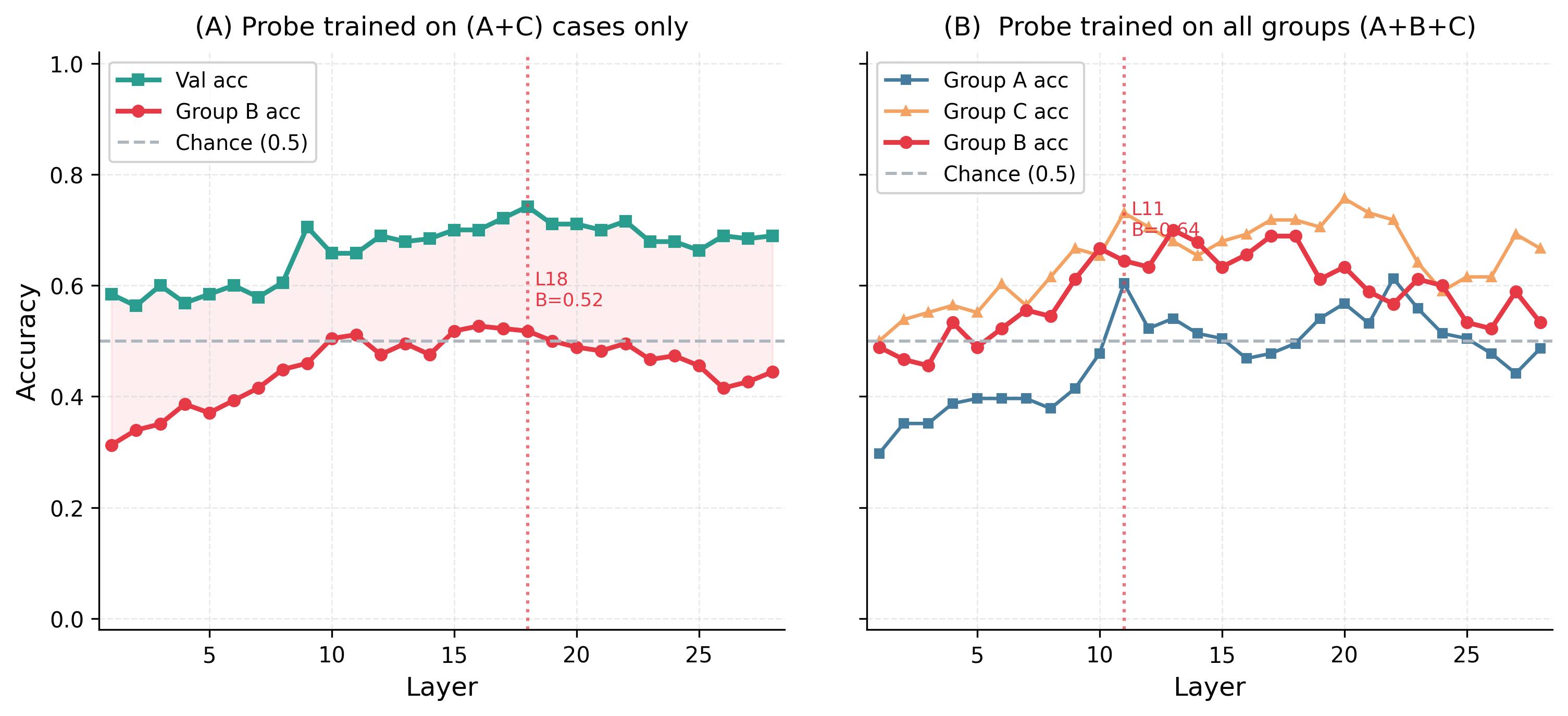}
        \caption{Static probe results on \textit{Qwen3-0.6B}}
        \label{fig:static_1}
    \end{subfigure}
    \begin{subfigure}[t]{0.82\textwidth}
        \centering
        \includegraphics[width=\textwidth]{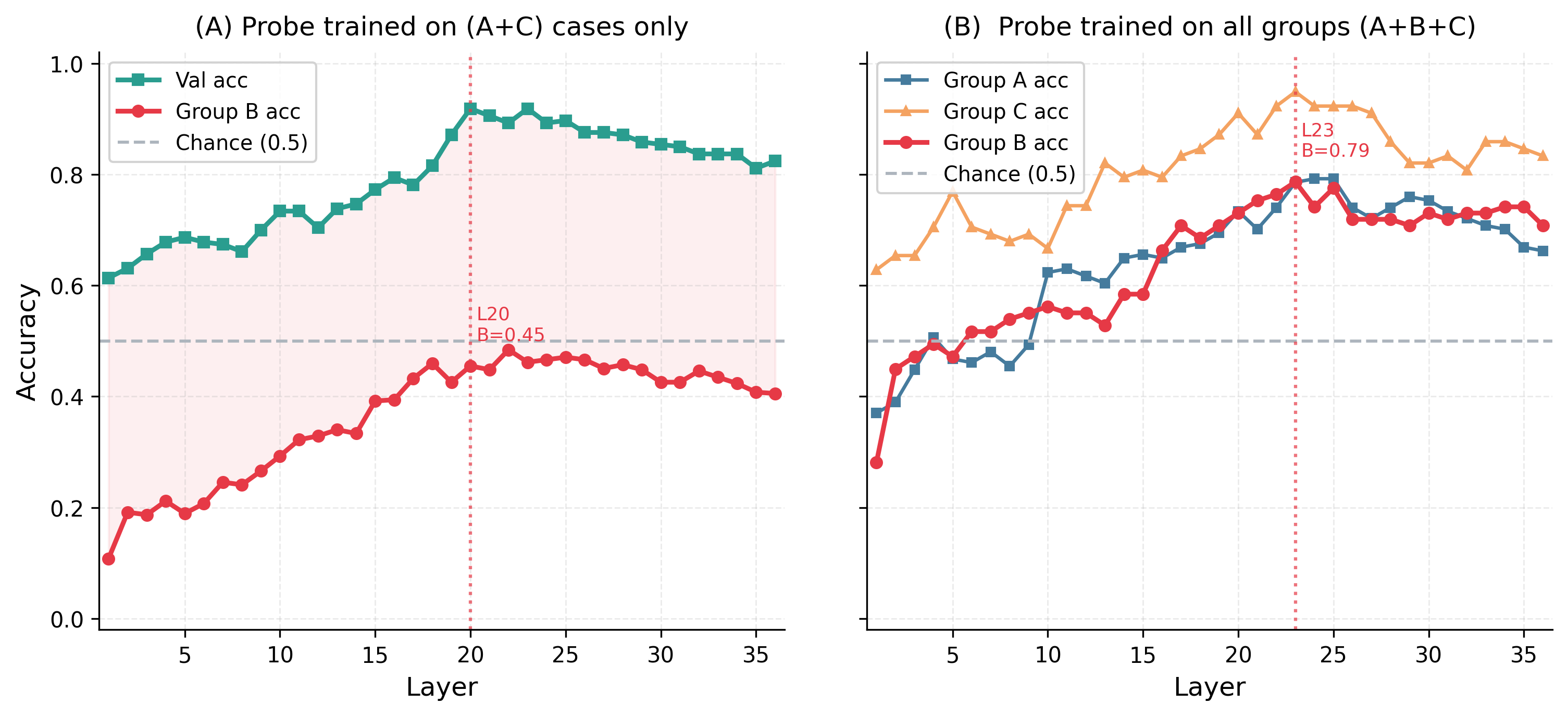}
        \caption{Static probe results on \textit{Qwen3-4B}}
        \label{fig:static_2}
    \end{subfigure}
    \vspace{0.5em}
    \begin{subfigure}[t]{0.82\textwidth}
        \centering
        \includegraphics[width=\textwidth]{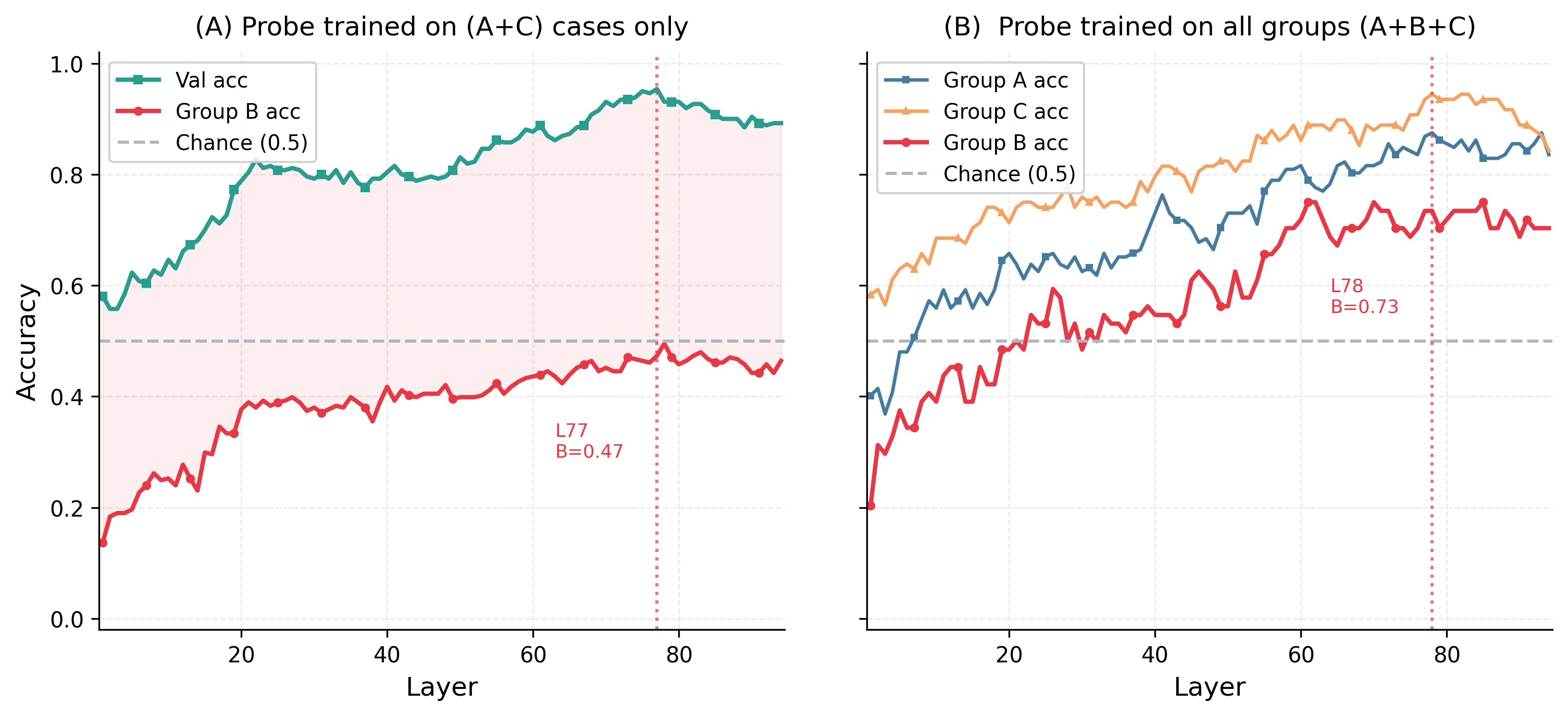}
        \caption{Static probe results on \textit{Qwen3-235B}}
        \label{fig:static_3}
    \end{subfigure}
    \caption{\textbf{Static probe results for \textit{Qwen} models.} \textit{Left:} When a static probe is trained to predict reasoning validity on the activations of concordant cases (Groups A and C), high accuracy is achieved on the validation set, with mid-to-late layers achieving the highest accuracy. However, accuracy falls to chance or below on Group B (fooled) cases. \textit{Right:} Oracle probes trained on all groups show different patterns for each model, but group B accuracy is generally lower than group C accuracy.}
    \label{fig:static-results}
\end{figure}

\begin{figure}[htbp]
    \centering
    \begin{subfigure}[t]{0.85\textwidth}
        \centering
        \includegraphics[width=\textwidth]{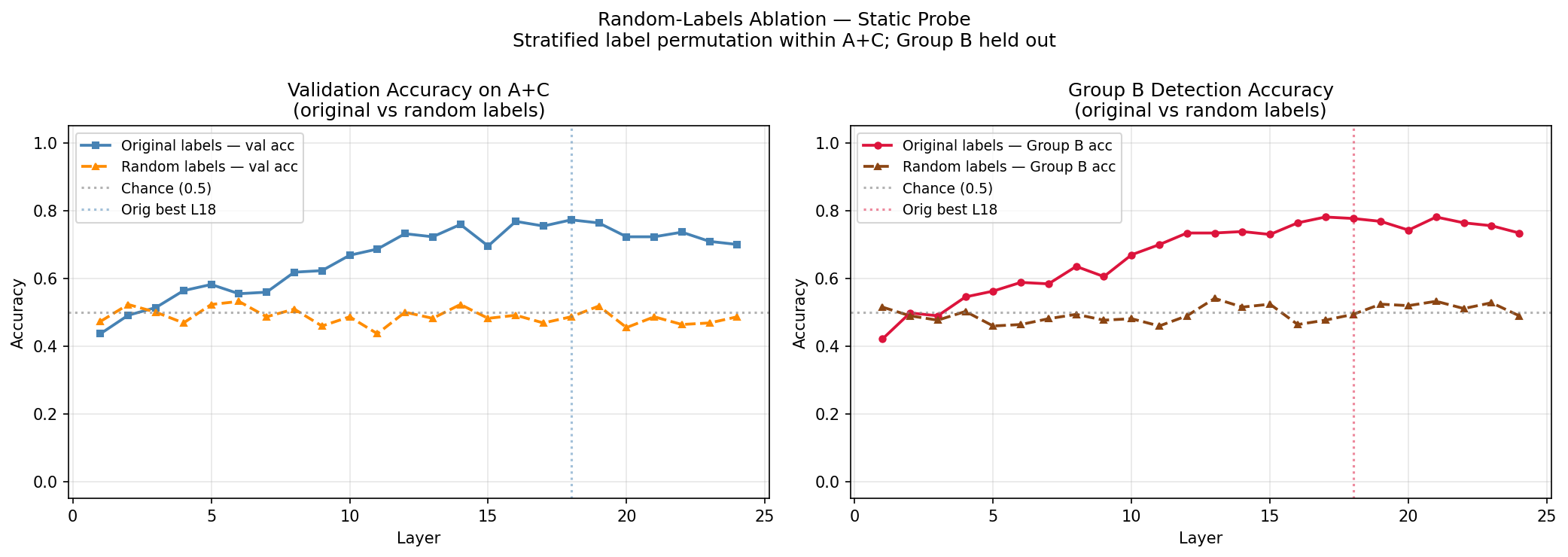}
        \caption{Static probe ablation study for \textit{GPT-oss-20B}}
        \label{fig:static_control_oss}
    \end{subfigure}
    \begin{subfigure}[t]{0.85\textwidth}
        \centering
        \includegraphics[width=\textwidth]{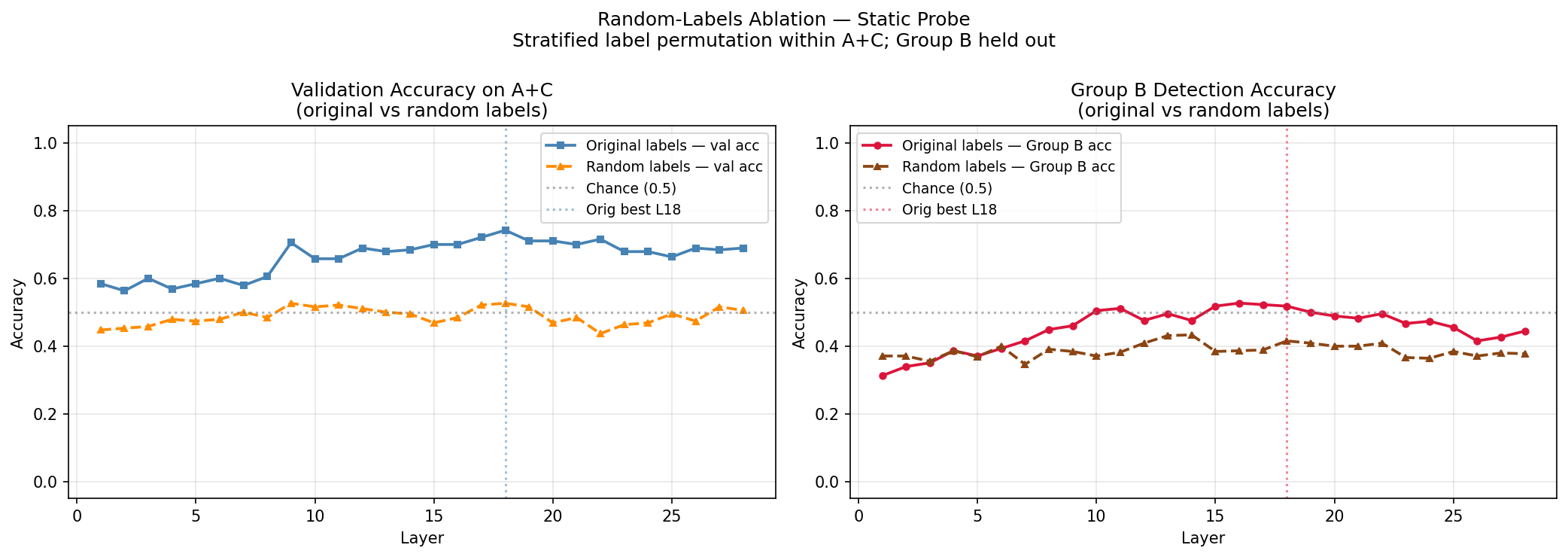}
        \caption{Static probe ablation study for \textit{Qwen3-0.6B}}
        \label{fig:static_control_0.6}
    \end{subfigure}
    \vspace{0.5em}
    \begin{subfigure}[t]{0.85\textwidth}
        \centering
        \includegraphics[width=\textwidth]{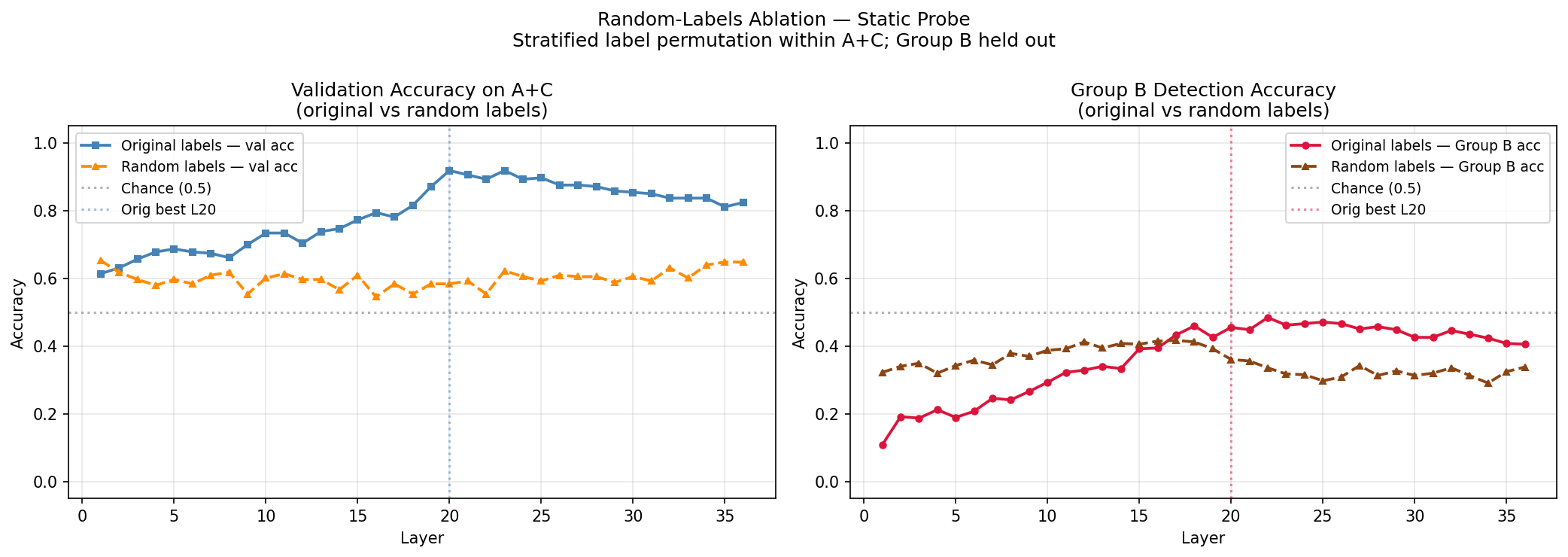}
        \caption{Static probe ablation study for \textit{Qwen3-4B}}
        \label{fig:static_control_4b}
    \end{subfigure}
    \vspace{0.5em}
    \begin{subfigure}[t]{0.85\textwidth}
        \centering
        \includegraphics[width=\textwidth]{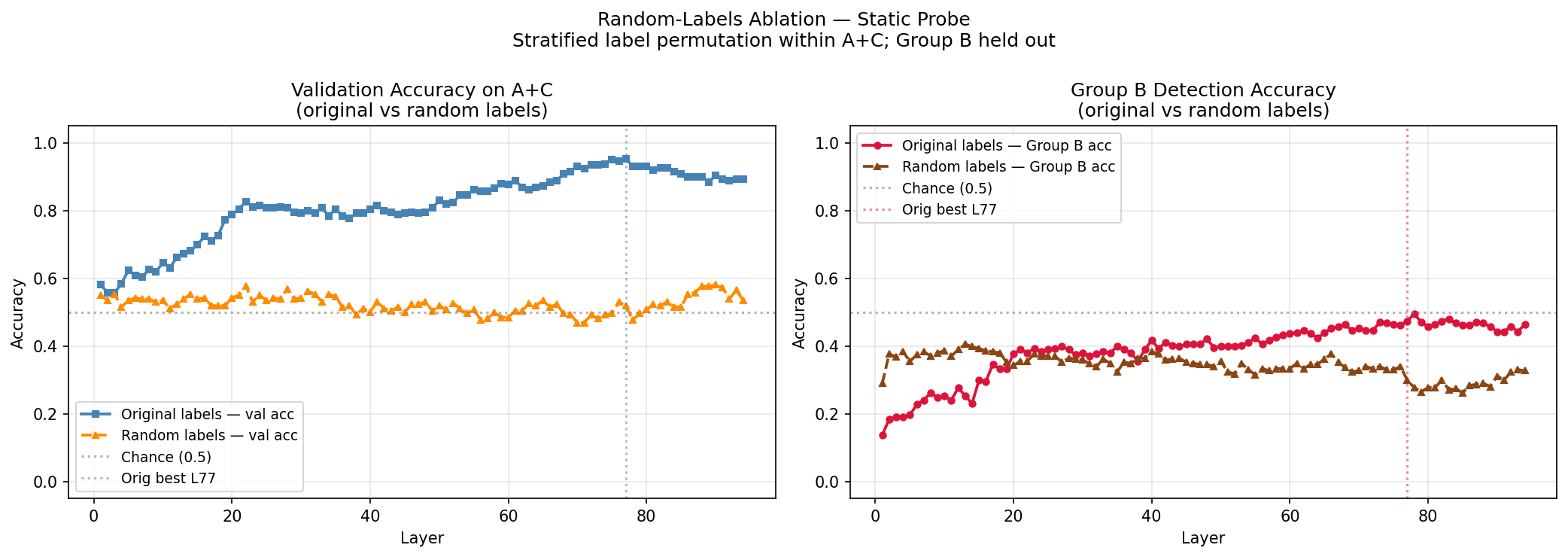}
        \caption{Static probe ablation study for \textit{Qwen3-235B}}
        \label{fig:static_control_235b}
    \end{subfigure}
    \caption{\textbf{Static probe ablation studies.} \textit{Left:} We validate that our probes are not learning spurious correlations by retraining the probes on random labels, finding that validation accuracy drops to chance as expected. \textit{Right:} For the held-out Group B, accuracy is around chance or worse when training on both random and original labels.}
    \label{fig:static-ablations}
\end{figure}

\vfill
\clearpage

\subsection{Dynamic Probe Analysis}
\label{appendix:dynamic-probe}

\begin{figure}[h]
    \centering
    \begin{subfigure}[t]{0.7\textwidth}
        \centering
        \includegraphics[width=\textwidth]{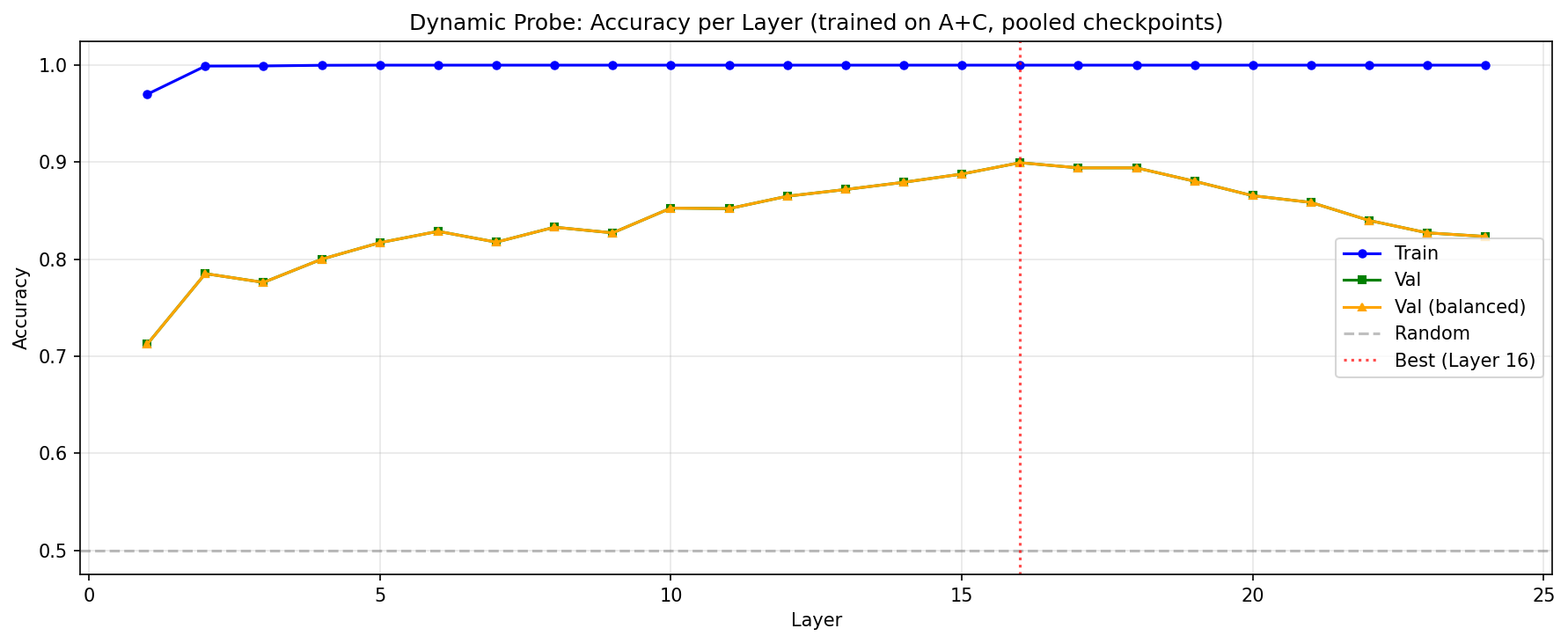}
        \caption{Dynamic probe accuracy on \textit{GPT-oss-20B}}
        \label{fig:causal-layers-oss}
    \end{subfigure}
    \vspace{0.5em}
    \begin{subfigure}[t]{0.7\textwidth}
        \centering
        \includegraphics[width=\textwidth]{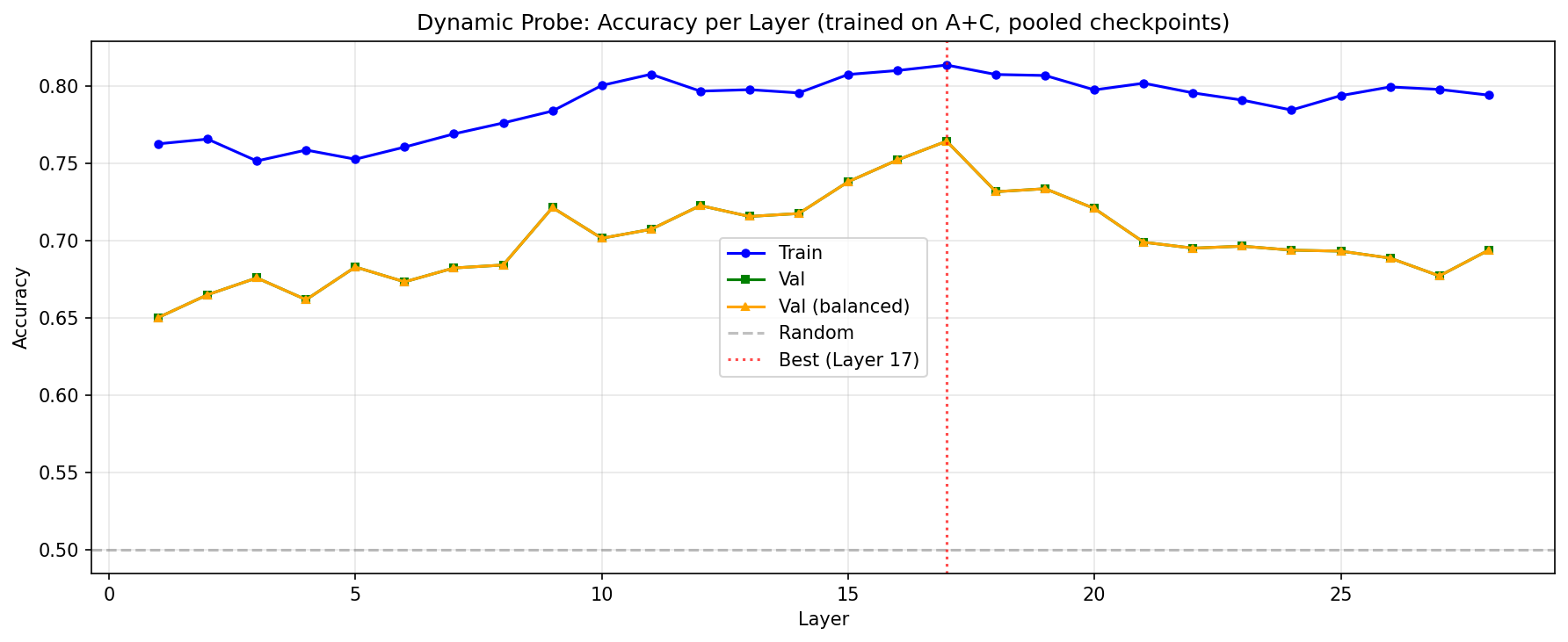}
        \caption{Dynamic probe accuracy on \textit{Qwen3-0.6B}}
        \label{fig:causal-layers-qwen06}
    \end{subfigure}
    \vspace{0.5em}
    \begin{subfigure}[t]{0.7\textwidth}
        \centering
        \includegraphics[width=\textwidth]{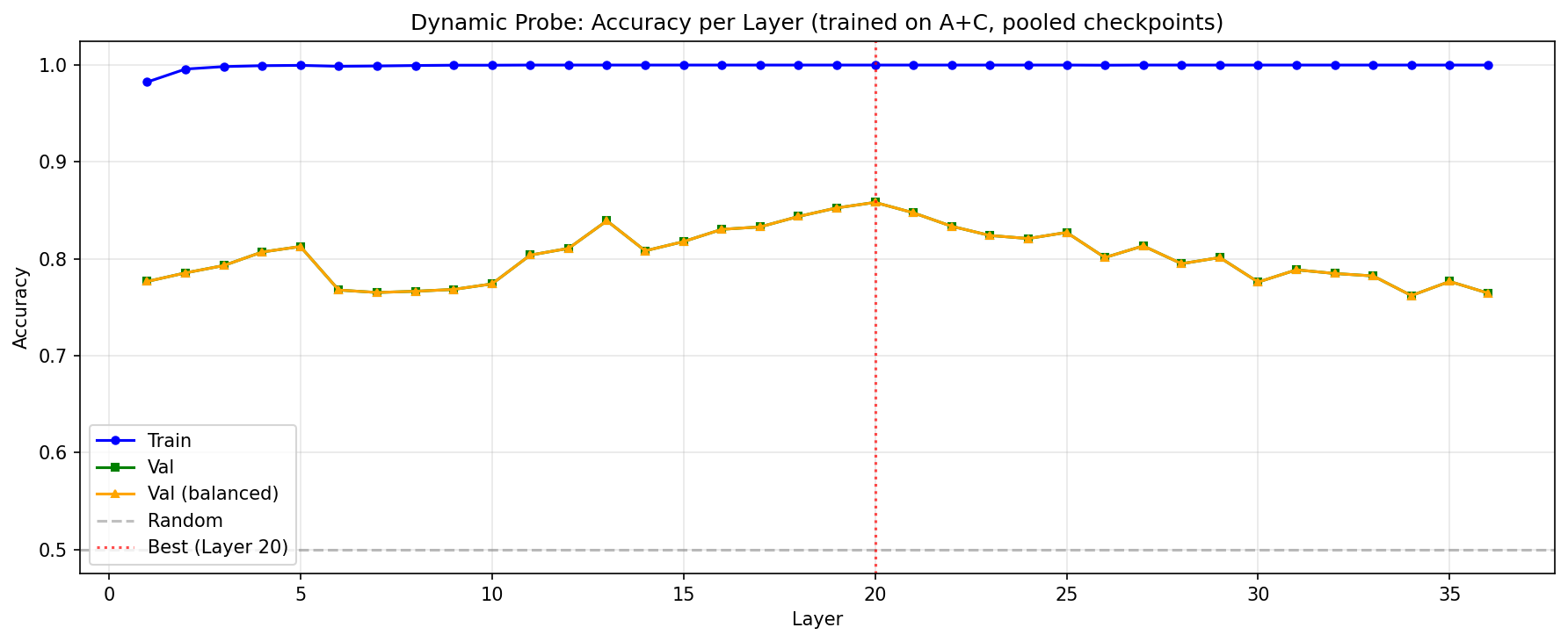}
        \caption{Dynamic probe accuracy on \textit{Qwen3-4B}}
        \label{fig:causal-layers-4b}
    \end{subfigure}
    \begin{subfigure}[t]{0.7\textwidth}
        \centering
        \includegraphics[width=\textwidth]{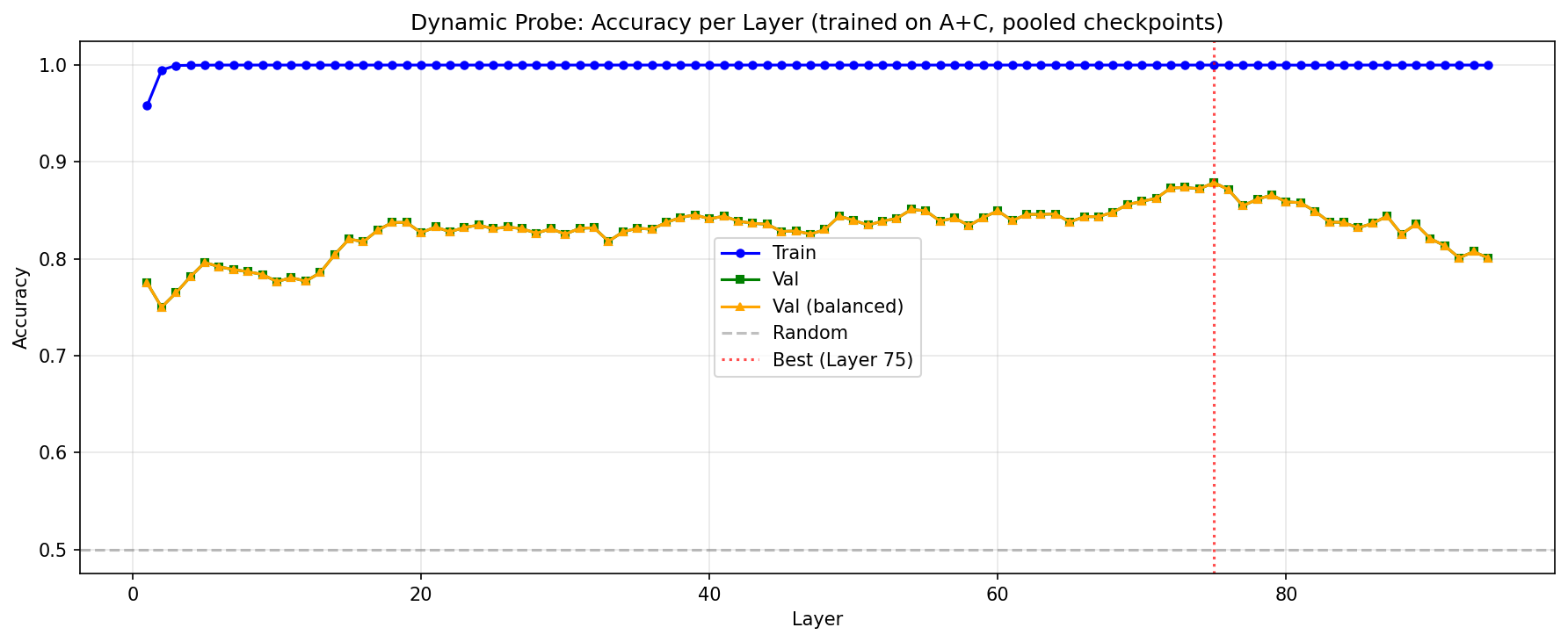}
        \caption{Dynamic probe accuracy on \textit{Qwen3-235B}}
        \label{fig:causal-layers-235b}
    \end{subfigure}
    \caption{\textbf{Dynamic probe accuracy across all models}, aggregated across all 10 checkpoints, and plotted by layer. We confirm that probe accuracy remains high on a held-out validation set (75--90\%), and identify the best performing layer for each model.}
    \label{fig:causal-layers}
\end{figure}

\vfill
\clearpage

\subsection{Causal Patching Analysis}
\label{appendix:causal-patching}

\begin{figure}[htbp]
    \centering
    \begin{subfigure}[b]{\textwidth}
        \centering
        \includegraphics[width=\textwidth]{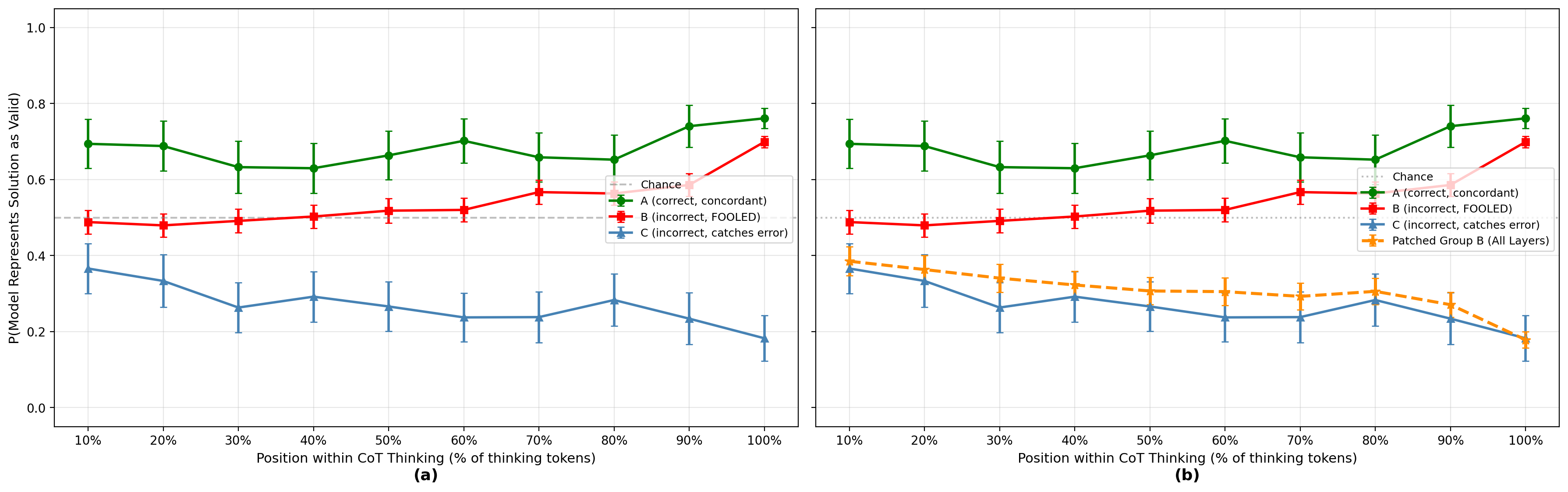}
        \caption{Dynamic probe and causal patching results on \textit{Qwen3-0.6B}}
        \label{fig:cot-causal-qwen06}
    \end{subfigure}
    
    \vspace{0.4cm}
    
    \begin{subfigure}[b]{\textwidth}
        \centering
        \includegraphics[width=\textwidth]{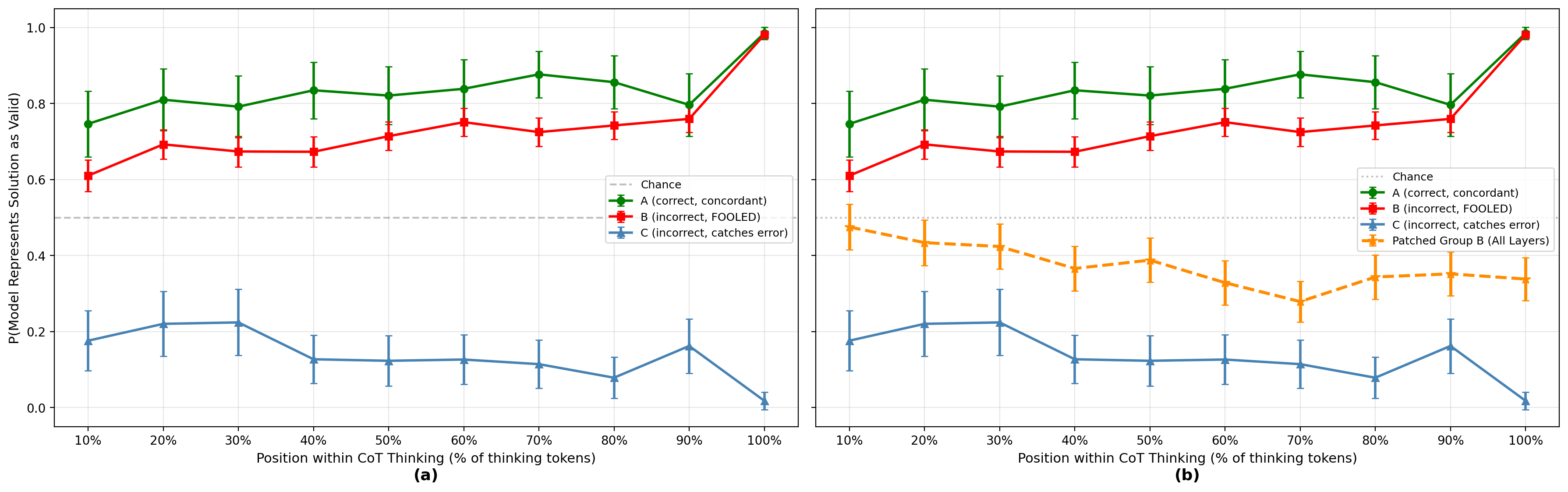}
        \caption{Dynamic probe and causal patching results on \textit{Qwen3-4B}}
        \label{fig:cot-causal-4b}
    \end{subfigure}
    
    \vspace{0.4cm}
    
    \begin{subfigure}[b]{\textwidth}
        \centering
        \includegraphics[width=\textwidth]{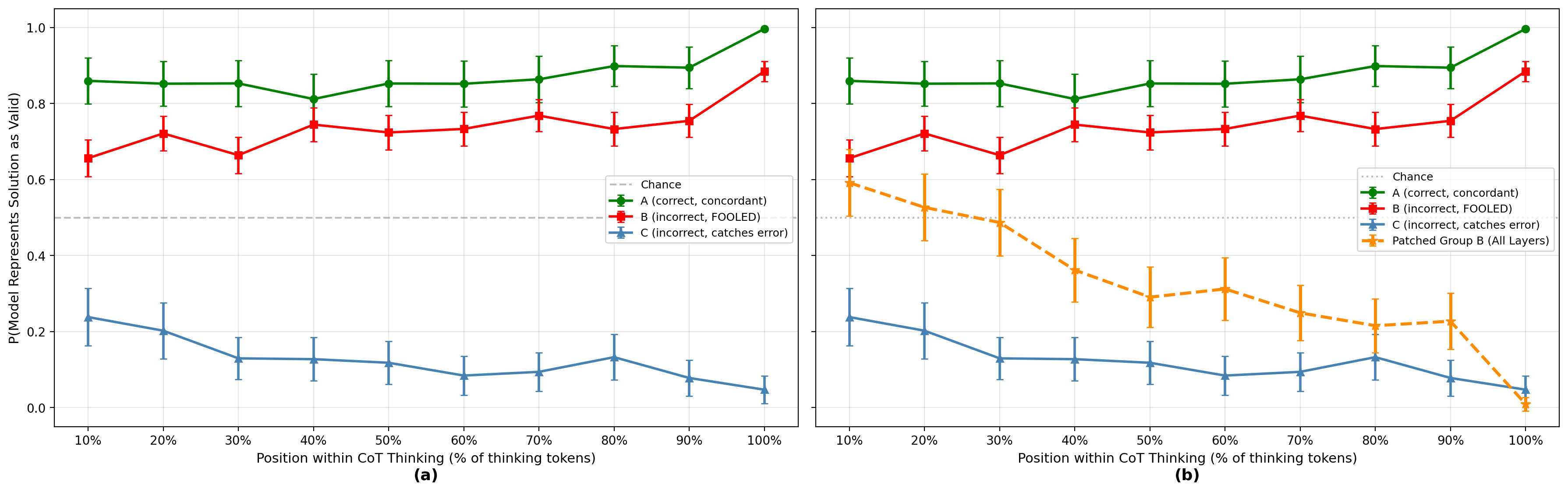}
        \caption{Dynamic probe and causal patching results on \textit{Qwen3-235B}}
        \label{fig:cot-causal-235b}
    \end{subfigure}
    
    \caption{\textbf{Dynamic probe and causal patching results across \textit{Qwen} models.} \textit{Left:} Similar to \textit{GPT-oss-20B}, dynamic probes generally classify Group B solutions along with Group A solutions as valid (and Group C as invalid), with increasingly strong verdicts as reasoning progresses from 10\% to 100\%. \textit{Right:} After patching Group B with the activations of an invalid answer across all layers, the probe classifies the solutions as invalid, with a similar decreasing trajectory as Group C.}
    \label{fig:cot-causal-combined}
\end{figure}

\begin{figure}[h]
    \centering
    \includegraphics[width=0.9\textwidth]{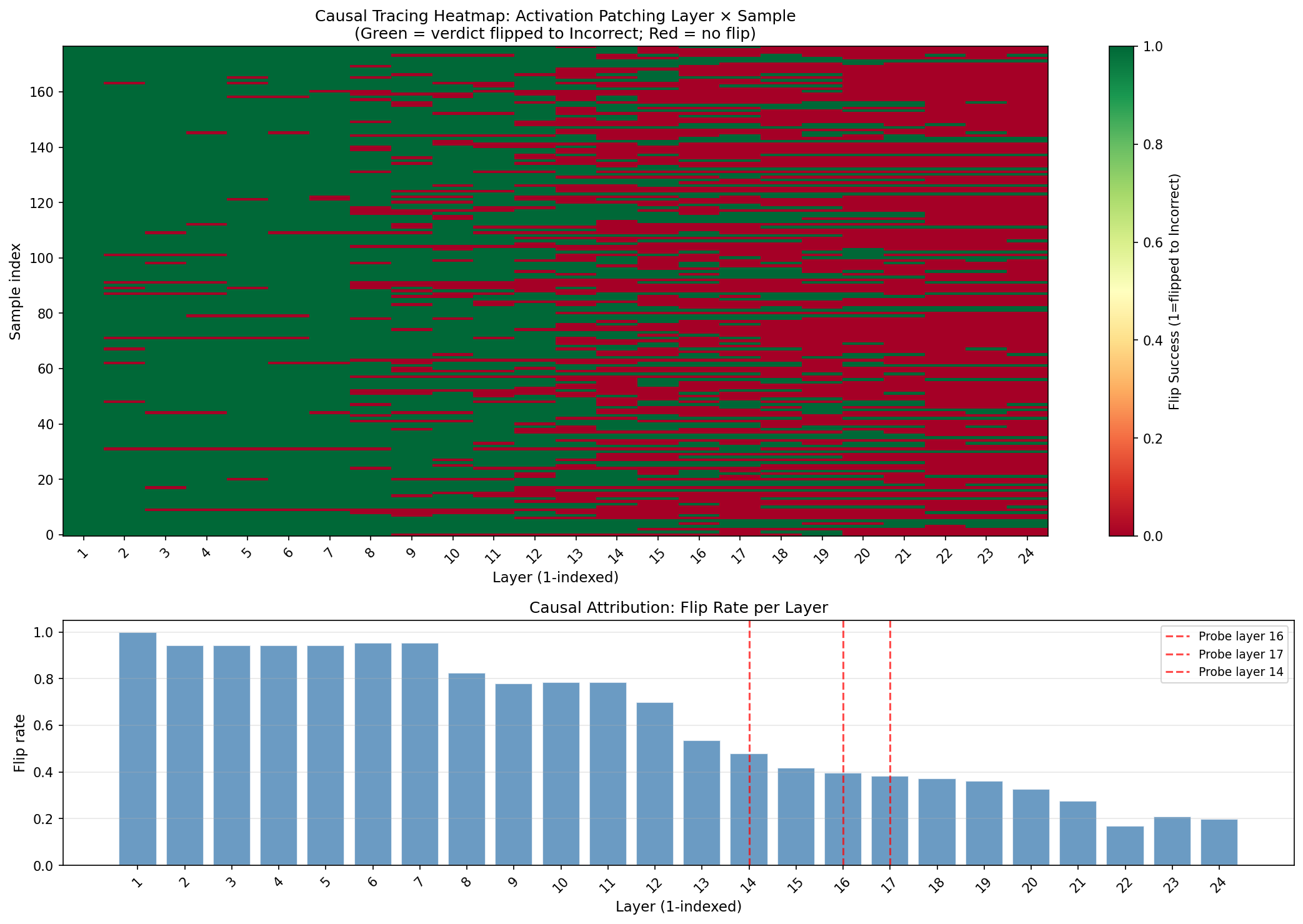}
    \caption{\textbf{Causal patching on each layer \& flip rate results} (\textit{GPT-oss-20B}). We plot the flip results (\textit{Top}) and overall flip rates (\textit{Bottom}) achieved by patching each layer of \textit{GPT-oss-20B} with invalid answer activation. The heatmap shows when the evaluation verdict flipped for each sample and layer in our patched dataset. Unsurprisingly, earlier layers achieve high flip rates, since patching those activations is similar to direct token replacement. However, reasonably high flip rates are also achieved for the layers that display high probe accuracy, indicating that the associated representations are not just informative of reasoning validity, but causally relevant to evaluator verdicts.}
    \label{fig:cot-causal-oss}
\end{figure}

\vfill
\clearpage

\subsection{Stratified Per-Perturbation Mechanistic Analysis}
\label{appendix:per-perturbation}

\vspace{1.5em}

As noted in the main text, LRMs perform worse at evaluating VAIR solutions with certain perturbation types, especially \emph{Shuffled Reasoning} and \emph{Missing Reasoning}. This raises the question of whether evaluation failures are simply due to structural tracking deficits (i.e., a model's inability to recognize that solution steps are disorganized or missing).

To investigate this, we compare the behavioral accuracy on each perturbation types against the accuracy of a linear probe trained to predict reasoning validity from the LRM's internal representations (\textbf{Table~\ref{tab:probe-accuracy-stratified}}). Linear probe accuracy significantly exceeds behavioral accuracy on \emph{Shuffled Reasoning} cases (increasing by 43\% to 67\%) and \emph{Missing Reasoning} cases (increasing by 16\% to 52\%). This gap indicates that internal representations retain sufficient signal to detect flawed reasoning, providing evidence against an inherent inability to detect missing or shuffled steps.

Instead, evaluation errors on these perturbations emerge primarily at the behavioral level despite internal detection, aligning with the forced rationalization behaviors observed in chain-of-thought traces. This pattern suggests a form of answer confirmation bias that selectively activates on invalid reasoning perceived as moderately flawed. \textit{Circular Reasoning} cases appear sufficiently obvious for models to bypass this bias in most instances, whereas \textit{Missing Premise} cases display minimal gaps between probe and behavioral accuracy, indicating a genuine failure to detect missing premises.

\begin{table}[h]
\centering
\caption{\textbf{Behavioral Accuracy vs.\ Probe Accuracy Across Perturbations.} Comparison between behavioral detection rates (LRM verdict accuracy) and the accuracy of a linear probe trained to predict reasoning validity from LRM internal representations (probe accuracy), stratified by perturbation types. The linear probe is trained to predict solution validity on Groups A, B, and C described in the main text, and evaluated on a held-out validation set.}
\label{tab:probe-accuracy-stratified}
\vspace{6pt}
\begin{tabular}{llccc}
\toprule
\textbf{Perturbation Type} & \textbf{Model} & \textbf{Behavioral Acc.} & \textbf{Probe Acc.} & \textbf{Behavior-Rep. Gap} \\
\midrule
\multirow{3}{*}{\textit{Circular Reasoning}} 
 & GPT-oss-20B & 93.7\% & 97.3\% & +3.6 pt  \\
 & Qwen3-4B    & 88.3\% & 95.8\% & +7.5 pt  \\
 & Qwen3-235B  & 100.0\% & 98.1\% & -1.9 pt  \\
\midrule
\multirow{3}{*}{\textit{Missing Premise}} 
 & GPT-oss-20B & 68.2\% & 66.4\% & -1.8 pt  \\
 & Qwen3-4B    & 63.0\% & 68.0\% & +5.0 pt  \\
 & Qwen3-235B  & 73.2\% & 78.0\% & +4.9 pt  \\
\midrule
\multirow{3}{*}{\textit{Shuffled Reasoning} }
 & GPT-oss-20B & 22.8\% & 84.0\% & +61.2 pt \\
 & Qwen3-4B    & 17.2\% & 84.1\% & +66.8 pt \\
 & Qwen3-235B  & 50.0\% & 92.5\% & +42.5 pt \\
\midrule
\multirow{3}{*}{\textit{Missing Reasoning} }
 & GPT-oss-20B & 41.1\% & 56.9\% & +15.7 pt \\
 & Qwen3-4B    & 13.7\% & 65.5\% & +51.8 pt \\
 & Qwen3-235B  & 37.5\% & 70.0\% & +32.5 pt \\
\bottomrule
\end{tabular}
\end{table}

\vfill
\clearpage

The above analysis suggests that certain types of reasoning flaws cause evaluation failure to occur more frequently when paired with valid answers --- i.e., that answer validity is \emph{jointly} causes failure along with perturbation type. This raises the question of whether answer validity is nonetheless causally necessary, regardless of perturbation type. 

To assess this, we stratify our causal patching analysis across perturbation types (\textbf{Table \ref{tab:causal-patching-stratified}}). We find that across all tested open-weight models and perturbation types, patching invalid answer activations produces substantial flip rates, causing models to alter their verdicts from 'correct' to 'Incorrect' 40\% to 66.7\% of the time.

A uniformity test indicates no statistically significant differences in flip rates across perturbation categories, demonstrating that the causal necessity of answer validity remains consistent regardless of perturbation type. In the absence of valid answer tokens, a large proportion of misjudged solutions are correctly classified as invalid. While answer validity alone is not causally sufficient (since this would require near-universal failure across all perturbation types and contexts) these findings demonstrate that answer validity serves as a key causal factor that induce evaluation failure.

\begin{table}[h]
\centering
\caption{\textbf{Causal Patching Across Perturbation Types.} Verdict flip rate (from Correct to Incorrect) on the patchable ``fooled'' subset (Group B) when substituting a valid answer token with the activations of an invalid shadow-answer token at the intervention layer. Results are stratified by perturbation type. Causal patching of the answer token flips LRM verdicts at uniformly high rates across perturbation types; the uniformity test ($\chi^2$ test) shows no statistically significant differences ($p > 0.49$).}
\label{tab:causal-patching-stratified}
\vspace{6pt}
\footnotesize
\begin{tabular}{lccc}
\toprule
\textbf{Perturbation Type} & \textbf{Qwen3-4B Flip Rate} & \textbf{GPT-oss-20B Flip Rate} & \textbf{Qwen3-235B Flip Rate} \\
\midrule
\textit{Circular Reasoning} & 44.4\% & 66.7\% & \textit{N/A}\textsuperscript{*} \\
\textit{Missing Premise}    & 60.0\% & 47.8\% & 40.0\% \\
\textit{Shuffled Reasoning}  & 48.3\% & 52.5\% & 44.9\% \\
\textit{Missing Reasoning}  & 55.0\% & 61.4\% & 46.2\% \\
\midrule
\textbf{Uniformity Test} & $\chi^2(3) = 2.38, p = 0.497$ & $\chi^2(3) = 2.10, p = 0.552$ & $\chi^2(2) = 0.42, p = 0.810$ \\
\bottomrule
\multicolumn{4}{l}{\footnotesize \textsuperscript{*}99.1\% detected as invalid; fooled subset too small.}
\end{tabular}
\end{table}

\clearpage
\end{document}